%% file: main.tex
\documentclass[10pt]{article} 
\usepackage[accepted]{tmlr}

\input{math_commands.tex}

\usepackage{hyperref}
\usepackage{url}
\usepackage{esvect}

\usepackage{algorithm}
\usepackage{algpseudocode}
\usepackage{graphicx}
\usepackage[font=small,labelfont=bf]{caption} 
\usepackage[T1]{fontenc} 

\usepackage{xcolor} 
\usepackage{subcaption} 
\usepackage{booktabs} 

\usepackage{graphicx}
\usepackage{wrapfig}
\usepackage{array}
\newcolumntype{H}{>{\setbox0=\hbox\bgroup}c<{\egroup}@{}}
\usepackage[small,compact]{titlesec} 
\usepackage{xcolor} 
\usepackage{float}
\usepackage[table]{colortbl}

\title{Lifelong Reinforcement Learning \\with Modulating Masks}

\author{\name Eseoghene Ben-Iwhiwhu \email e.ben-iwhiwhu@lboro.ac.uk \\
\addr Department of Computer Science, Loughborough University, UK
\AND
\name Saptarshi Nath \email s.nath@lboro.ac.uk \\
\addr Department of Computer Science, Loughborough University, UK
\AND
\name Praveen K. Pilly \email pkpilly@hrl.com \\
\addr HRL Laboratories, LLC, Malibu, CA, 90265, USA
\AND
\name Soheil Kolouri \email soheil.kolouri@vanderbilt.edu \\
\addr Department of Computer Science, Vanderbilt University, Nashville, TN, USA
\AND
\name Andrea Soltoggio \email a.soltoggio@lboro.ac.uk \\
\addr Department of Computer Science, Loughborough University, UK
}

\def\month{07}  
\def\year{2023} 
\def\openreview{\url{https://openreview.net/forum?id=V7tahqGrOq}} 

\begin{document}

\maketitle

\begin{abstract}
Lifelong learning aims to create AI systems that continuously and incrementally learn during a lifetime, similar to biological learning. Attempts so far have met problems, including catastrophic forgetting, interference among tasks, and the inability to exploit previous knowledge. While considerable research has focused on learning multiple supervised classification tasks that involve changes in the input distribution, lifelong reinforcement learning (LRL) must deal with variations in the state and transition distributions, and in the reward functions. Modulating masks with a fixed backbone network, recently developed for classification, are particularly suitable to deal with such a large spectrum of task variations. In this paper, we adapted modulating masks to work with deep LRL, specifically PPO and IMPALA agents. The comparison with LRL baselines in both discrete and continuous RL tasks shows superior performance. We further investigated the use of a linear combination of previously learned masks to exploit previous knowledge when learning new tasks: not only is learning  faster, the algorithm solves tasks that we could not otherwise solve from scratch due to extremely sparse rewards. The results suggest that RL with modulating masks is a promising approach to lifelong learning, to the composition of knowledge to learn increasingly complex tasks, and to knowledge reuse for efficient and faster learning.
\end{abstract}

\section{Introduction}
\label{sec:introduction}
Lifelong reinforcement learning (LRL) is a recent and active area of research that seeks to enhance current RL algorithms with the following capabilities: learning multiple tasks sequentially without forgetting previous tasks, exploiting previous tasks knowledge to accelerate learning of new tasks, or building solutions to increasingly complex tasks. The remarkable progress of RL in recent years, stemming in particular from the introduction of deep RL \citep{mnih2013playing}, has demonstrated the potential of algorithms that can learn to act in an environment with complex inputs and rewards. However, the limitation of learning only one task means that such algorithms have a narrow focus, and may not scale to complex tasks that first require learning of sub-tasks. It is arguable that acquiring knowledge of multiple tasks over a lifetime, generalizing across such tasks, exploiting acquired knowledge to master new tasks more quickly, or composing simple tasks to solve more complex tasks are necessary to develop more powerful AI systems that better mimic biological intelligence.

LRL shares some of the objectives of lifelong supervised learning (LSL) \citep{parisi2019continual,hadsell2020embracing,de2021continual}, but due to the unique objective and domains of RL, it has developed into a separate field \citep{khetarpal2020towards}. For example, tasks in LRL can vary across reward function, input distribution or transition function, while tasks in LSL largely vary in the input distribution. A variety of recently developed approaches in LSL can be placed under three categories: memory methods, synaptic consolidation methods, and parameter isolation or modular methods. In LRL, \citet{khetarpal2020towards} propose a taxonomy in which approaches can be classified as explicit knowledge retention, leveraging shared structures, and learning to learn. It can be appreciated that LRL exploits most advances in LSL while also developing RL-specific algorithms.  

Among the LSL categories mentioned above, parameter isolation methods have shown state-of-the-art results in classification with approaches such as Progressive Networks \citep{rusu2016progressive, yoon2018den}, PackNet \citep{mallya2018packnet}, Piggyback \citep{mallya2018piggyback} and Supermasks \citep{wortsman2020supermasks}. The key concept in such methodologies is to find smart and efficient ways to isolate parameters for each task. In \citet{mallya2018packnet}, redundancies in large networks are exploited to generate masks via iterative pruning, and thus free parameters to be used only for specific tasks, thus ``packing'' multiple tasks in one single network. Other approaches instead use the concept of directly learned masks \citep{zhou2019deconstructing,ramanujan2020s,wortsman2020supermasks} to select parts of a backbone network for each task. Instead of learning the parameters of a backbone network, which are set randomly, masking approaches focus on learning a multiplicative (or modulating) mask for each task. Masks can be effective in binary formats, enabling or disabling parts of the backbone network while requiring low memory. Interestingly, masks can be interpreted as \emph{modulatory} mechanisms with biological inspiration \citep{kudithipudi2022biological} because they exploit a gating mechanism on parameters or activations. 

Surprisingly, while the application of masks has been tested extensively in LSL for classification, very little is known on their effectiveness in LRL. Also, the current mask methods lack the ability of exploit knowledge from previously learned task (forward transfer), since each mask is derived independently. This study introduces the use of directly learned masks with policy optimization algorithms, specifically PPO \citep{schulman2017ppo} and IMPALA \citep{espeholt2018impala}. We explore (1) the effectiveness of masks to maximize lifelong learning evaluation metrics, and (2) the suitability of masks to reuse knowledge and accelerate learning (forward transfer). To demonstrate the first point, we test the approach on RL curricula with the Minigrid environment \citep{gym_minigrid}, the CT-graph \citep{soltoggio2019ctgraph, soltoggio2023ctgraph}, Metaworld \citep{yu2020meta}, and ProcGen \citep{cobbe2020procgen}, and assess the lifelong learning metrics \citep{new2022lifelong, baker2023} when learning multiple tasks in sequence. To demonstrate the second point, we exploit linear combinations of masks and investigate learning with curricula in which tasks have similarities and measure the forward transfer. The investigation of the second point gives rise to the understanding of the effectiveness of mask knowledge reuse in RL, the initial parameter configuration ideal for mask reuse, and the benefits obtained when task curriculum grow in complexity or contains interfering tasks. The proposed method assumes a task oracle, but the assumption can be eliminated through the introduction of task detection methods into the system as discussed in Section \ref{sec:discussion}. To ensure reproducibility, the hyper-parameters for the experiments are reported in Appendix \ref{apndx:hyperparameters}. The code is published at \url{https://github.com/dlpbc/mask-lrl}.

{\bf Contributions.} To the best of our knowledge, this is the first study to (1) introduce learned modulating masks in LRL and show competitive advantage with respect to well established LRL methods (2) show the exploitation and the composition of previously learned knowledge in a LRL setup with learned modulating masks. In addition to a baseline masking approach with independent mask search, we introduced two new incremental mask composition approaches (LC and BLC) with different initialization of linear combination parameters. The analysis reveals different properties in the exploitation of previous knowledge while learning new tasks, either favoring knowledge re-use or exploration.

\section{Related work}
\label{sec:related-work}
\subsection{Deep Reinforcement Learning (DeepRL)}
\label{subsec:deep-rl}
The use of deep networks as function approximators in reinforcement learning (RL) has garnered widespread adoption, showcasing results such as learning to play video games \citep{mnih2013playing, van2016doubledqn, shao2018learning, kempka2016vizdoom, mnih2016a3c, babaeizadeh2017ga3c, espeholt2018impala}, and learning to control actual and simulated robots \citep{lillicrap2016ddpg, schulman2015trpo, schulman2017ppo, fujimoto2018td3, haarnoja2018sac}. These algorithms enable the agent to learn how to solve a single task in a given evaluation domain. In a lifelong learning scenario with multiple tasks, they suffer from lifelong learning challenges such as catastrophic forgetting. Nevertheless, they serve as the basis for lifelong learning algorithms. For example, DQN \citep{mnih2013playing} was combined with the elastic weight consolidation (EWC) algorithm to produce a lifelong learning DQN agent \citep{kirkpatrick2017ewc, kessler2022owl}.

\subsection{Lifelong (Continual) Learning}
\label{subsec:lifelong-learning}
Several advances have recently been introduced in lifelong (continual) learning \citep{van2019three, parisi2019continual, hadsell2020embracing, de2021continual, khetarpal2020towards}, addressing challenges such as maintaining performance on previously learned tasks while learning a new task (overcoming forgetting), reusing past knowledge to rapidly learn new tasks (forward transfer), improving performance on previously learned tasks from newly acquired knowledge (backward transfer), the efficient use of model capacity to reduce intransigence, and reducing or avoiding interference across tasks \citep{kessler2022owl}. A large body of work focused on lifelong learning in the supervised learning domain and overcoming the challenge of forgetting \citep{mendez2022neuralcomposition}.

Lifelong learning algorithms can be clustered into key categories such as synaptic consolidation approaches \citep{kirkpatrick2017ewc, zenke2017si, aljundi2018mas, kolouri2019scp}, memory approaches \citep{lopez2017gem, zeng2019owm, chaudhry2018agem, rolnick2019clear, lin2022trgp}, modular approaches \citep{rusu2016progressive, mallya2018packnet, mallya2018piggyback, wortsman2020supermasks, mendez2022neuralcomposition} or a combination of the above. Synaptic consolidation approaches tackle lifelong learning by discouraging the update of parameters useful for solving previously learned tasks through a regularization penalty. Memory approaches either store and replay samples of previous and current tasks (from a buffer or a generative model) during training \citep{lopez2017gem, chaudhry2018agem, guo2020gem_unified, oswald2020continual_hypernet} or project the gradients of the current task being learned in an orthogonal direction to the gradients of previous tasks \citep{zeng2019owm, farajtabar2020ogd, saha2021gpm, lin2022trgp}. Memory methods aim to keep the input-output mapping for previous tasks unchanged while learning a new task. Modular approaches either expand the network as new tasks are learned \citep{rusu2016progressive, yoon2018den, martin2019ppn} or select sub-regions of a fixed-sized network (via masking) for each tasks \citep{wortsman2020supermasks}. The masks can be applied to (i) the neural representations \citep{sokar2021spacenet, serra2018hat}, (ii) or to synapses, where they are directly learned \citep{wortsman2020supermasks} or derived through iterative pruning \citep{mallya2018packnet}. The masks can be viewed as a form induced sparsity in the neural lifelong learner \citep{von2021learning, hoefler2021sparsity}.

In LRL, the learner is usually developed by combining a standard deep RL algorithm, either on-policy or off-policy (for example, DQN, PPO \citep{schulman2017ppo}, or SAC \citep{haarnoja2018sac}), with a lifelong learning algorithm. CLEAR \citep{rolnick2019clear} is a lifelong RL that combines IMPALA with a replay method and behavioral cloning. Progress \& Compress \citep{schwarz2018pnc} demonstrated the combination of IMPALA with EWC and policy distillation techniques. Other notable methods include the combination of a standard deep RL agent (SAC) with mask derived from pruning in PackNet \citep{mallya2018packnet, schwarz2021powerpropagation} as demonstrated in the Continual World robotics benchmark \citep{wolczyk2021continualworld}. In addition, \citet{mendez2022neuralcomposition} developed an algorithm combining neural composition, offline RL, and PPO to facilitate knowledge reuse and rapid task learning via functional composition of knowledge. To tackle a lifelong learning scenario with interference among tasks, current methods \citep{kessler2020unclear, kessler2022owl} employ a multi-head policy network (i.e., a network with shared feature extractor layers connected to different output layers/heads per task) that combine a standard RL algorithm (e.g. DQN or SAC) with synaptic consolidation methods. In another class of LRL approach, \citet{kaplanis2018continual} demonstrated a lifelong RL agent that combined DQN with multiple-time scale learning at the synaptic level \citep{benna2016computational}, and the model was adapted to multiple timescale learning at the policy level \citep{kaplanis2019policy} via the combination of knowledge distillation and PPO.

\subsection{Modulation}
\label{subsec:modulation}
Neuromodulatory processes \citep{avery2017neuromodulatory, bear2020neuroscience} enable the dynamic alteration of neuronal behavior by affecting synapses or the neurons connected to synapses. Modulation in artificial neural networks \citep{fellous1998computational,doya2002metalearn_neuromod} draws inspiration from modulatory dynamics in biological brains that have proven particularly effective in reward-based environments \citep{schultz1997neural,abbott2004synaptic}, in the evolution of reward-based learning  \citep{soltoggio2008evolutionary,soltoggio2018born}, and in meta RL \citep{ben2022context}. Modulatory methods in lifelong learning are set up as masks that alter either the neural activations \citep{serra2018hat} or weights of neural networks \citep{mallya2018packnet, mallya2018piggyback, wortsman2020supermasks, koster2022signing_sm}. The key insight is the use of a modulatory mask (containing binary or real values) to activate particular network sub-regions and deactivate others when solving a task. Therefore, each task is solved by different sub-regions of the network. For each task, PackNet \citep{mallya2018packnet} trains the model based on available network capacity and then prunes the network to select only parameters that are important to solving the task. A binary mask representing important and unimportant parameters is then generated and stored for that task, while ensuring that important parameters are never changed when learning future tasks. PiggyBack \citep{mallya2018piggyback} keeps a fixed untrained backbone network, while it trains and stores mask parameters per task. During learning or evaluation of a particular task, the mask parameters for the task are discretized and applied to the backbone network by modulating its weights. The Supermask \citep{wortsman2020supermasks} is a generalization of the PiggyBack method that uses a k-winner approach to create sparse masks.

\section{Background}
\label{sec:background}

\subsection{Problem Formulation}
\label{subsec:problem-formulation}
A reinforcement learning problem is formalized as a Markov decision process (MDP), with tuple $\mathcal{M} = \langle\mathcal{S}, \mathcal{A}, p, r, \gamma \rangle$, where $\mathcal{S}$ is a set of states, $\mathcal{A}$ is a set of actions, $p: \mathcal{S} \times \mathcal{A} \rightarrow \mathcal{S}$ is a transition probability distribution $p(s_{t+1} | s_t, a_t)$ of the next state given the current state and action taken at time $t$, $r: \mathcal{S} \times \mathcal{A} \rightarrow \mathbb{R}$ is a reward function that produces a scalar value based on the state and the action taken at the current time step, $\gamma \rightarrow [0, 1]$ is the discount factor that determines how importance future reward relative to the current time step $t$. An agent interacting in the MDP behaves based on a defined policy (either a stochastic $\pi$ or a deterministic $\mu$ policy). The objective of the agent is to maximize the total reward achieved, defined as an expectation over the cumulative discounted reward $\mathbb{E}[\sum_{t=0}^{\infty} \gamma^t r(s_t, a_t)]$. It achieves this by learning an optimal policy.

In a lifelong learning setup, a lifelong RL agent is exposed to a sequence of tasks $\mathcal{T}$. Given $N$ tasks, the agent is expected to maximize the RL objective for each task in $\mathcal{T} = \{\tau_1, \tau_2, \ldots, \tau_n\}$. As the agent learns one task after another in the sequence, it is required to maintain (avoid forgetting) or improve (backward transfer of knowledge) performance on previously learned tasks. A desired property of such an agent is the ability to reuse knowledge from previous tasks to rapidly learn the current or future tasks.

\subsection{Modulating masks}
\label{subsec:supermask}
The concept of modulating masks has recently emerged in the area of supervised classification learning \citep{mallya2018piggyback, serra2018hat, ramanujan2020s, wortsman2020supermasks, sokar2021spacenet, koster2022signing_sm}. Masks work by modulating the weights or the neural representations of a network, and they can be directly learned (optimized) or derived via iterative pruning \citep{hoefler2021sparsity}, with a goal of implementing sparsity in the network. The learned mask approach is employed in this work, and masks are used to modulate the weights of a network. By modulating the weights, sub-regions of the network is activated, while other regions are deactivated. Given a network with layers ${1, \ldots L}$, with each layer $l$ containing parameters $W^l \in \mathbb{R}^{m \times n}$, for each layer $l$, a score parameter $S^l \in \mathbb{R}^{m \times n}$ is defined. 

During a forward pass (in training or evaluation), a binary mask $M^l \in \{0, 1\}^{m \times n}$ is generated from $S^l$ based on an operation $g(S^l)$ according to one of the following: (i) a threshold $\epsilon$ (where values greater than $\epsilon$ are set to $1$, otherwise $0$) \citep{mallya2018piggyback}, or (ii) top-k values (the top $k\%$ values in $S^l$ yields $1$ in $M^l$, while the rest are set to $0$) \citep{ramanujan2020s}, or (iii) probabilistically sampled from a Bernoulli distribution, where the $p$ parameter for the distribution is derived from the sigmoid function $\sigma(S^l)$ \citep{zhou2019deconstructing}. An alternative approach is the generation of ternary masks (i.e., with values $\{-1, 0, 1\}$) from $S^l$, as introduced in \citet{koster2022signing_sm}. The binary or ternary mask modulates the layer's parameters (or weights), thus activating only a subset of the $W^l$. 

Given an input sample $\mathbf{x} \in \mathbb{R}^m$, a forward pass through the layer is given as $f(\mathbf{x}, W^l, S^l) = (W^l \odot M^l) \cdot \mathbf{x}$, where $\odot$ is the element wise multiplication operation. Only the score parameters $S^l$ are updated during training, while the weights $W^l$ are kept fixed at their initial values. For brevity, the weights and scores across layers are denoted as $W$ and $S$.

The training algorithm updating $S^l$ depends on the operation used to generate the binary mask. When the binary masks are generated from Bernoulli sampling, standard backpropagation is employed. However, in the case of thresholding or selecting top $k\%$, an algorithm called edge-popup is employed, which combines backpropagation with the straight-through gradient estimator (i.e., where the gradient of the binary mask operation is set to identity to avoid zero-value gradients) \citep{bengio2013straight_through, courbariaux2015binaryconnect}.

In a lifelong learning scenario with multiple tasks, for each task $\tau_k$, a score parameter $S_k$ is learned and used to generate and store the mask $M_k$. During evaluation, when the task is encountered, the mask learned for the task is retrieved and used to modulate the backbone network. Since the weights of the network are kept fixed, it means there is no forgetting. However, this comes at the cost of storage (i.e., storing a mask for each task).

\section{Methods}
\label{sec:methods}
We first introduce the adaptation of modulating masks to RL algorithms (Section \ref{subsec:supermask}). Following, we hypothesize that previously learned masks can accelerate learning of unseen tasks by means of a linear combination of masks (Section \ref{subsec:supermask-linear-comb}). Finally, we suggest that continuous masks might be necessary to solve RL tasks in continuous environments (Section \ref{subsec:continuous-masks}).

\subsection{Modulatory masks in Lifelong RL}
\label{subsec:supermaks-lifelong-rl}
We posit that a stochastic control policy $\pi_{\theta, \Phi}$ can be parameterized by $\theta$, the weights of a fixed backbone network, and $\Phi = \{\phi_1 \ldots \phi_k\}$, a set of mask score parameters for tasks $1 \ldots k$. $\phi_k$ are the  scores of all layers of the network for task $k$, comprising the layers $1 \ldots L$, i.e., $\phi_k = \{S_k^1 \ldots S_k^L\}$. The weights $\theta$ of the network are randomly initialized using the signed Kaiming constant method \citep{ramanujan2020s}, and are kept fixed during training across all tasks. The mask score parameters $\Phi$ are trainable, but only $\phi_k$ is trained when the agent is exposed to task $k$. To reduce memory requirements,  \citet{wortsman2020supermasks} discovered that masks can be reduced to binary without significant loss in performance. We test this assumption by binarizing masks  using an element-wise thresholding function
\begin{equation}
\label{eqn:score-to-binary-mask}
    g(\phi_k) =
      \begin{cases}
        1 & \phi_{k, \{i, j\}} > \epsilon \\
        0 & \text{otherwise}
      \end{cases}       
\end{equation}
where $\epsilon$ is set to 0. The resulting LRL algorithm is described in Algorithm \ref{algo1}. To assess this algorithm, we paired it with the on-policy PPO \citep{schulman2017ppo} algorithm and the off-policy IMPALA \citep{espeholt2018impala} algorithm.
\begin{algorithm}
    \caption{Lifelong RL Algorithm with modulating masks}
    \label{alg:lrl-train}
    \begin{algorithmic}[1]
        \Require Number of tasks $N$, maximum training steps per task P
        \Require Rollout length (steps) z

        \State Initialize policy $\pi_{\theta, \Phi}$
        \For{k $\leftarrow$ 1 \ldots N}
            \State Set task $k$
            \State Set $steps \leftarrow 0$
            \While{step $<$ P}
                \State Rollout experiences $\{(s_1, a_1, r_1, s_1^\prime) \ldots (s_z, a_z, r_z, s_z^\prime)\}$ using $\pi_{\theta, \phi_k}$.
                \State Update $steps \leftarrow steps + z$
                \State Compute loss $\mathcal{L}_{k}(\pi_{\theta, \phi_k})$ based on an RL algorithm objective.
                \State Compute gradients $\nabla_{\phi_{k}}\mathcal{L}_{k}(\pi_{\theta, \phi_k})$ with respect to $\phi_k$.
                \State Update mask score parameter for task $k$, $\phi_k \leftarrow \phi_k - \alpha \nabla_{\phi_{k}}\mathcal{L}_{k}(\pi_{\theta, \phi_k})$
            \EndWhile
            \State Store mask score $\phi_k$ or the binary mask $g(\phi_k)$
        \EndFor
    \end{algorithmic}
    \label{algo1}
\end{algorithm}

\subsection{Exploiting previous knowledge to learn new tasks}
\label{subsec:supermask-linear-comb}
One assumption in LRL, often measured with metrics such as forward and backward transfer \citep{new2022lifelong}, is that some tasks in the curriculum have similarities. Thus, previously acquired knowledge, stored in masks, may be exploited to learn new tasks. To test this hypothesis, we propose an incremental approach to using previously learned masks when facing a new task. Rather than learn a large number of masks and infer which mask is useful for solving a task at test time via gradient optimized linear combination, as in \citet{wortsman2020supermasks}, we instead start to exploit previous knowledge from any small number of masks combined linearly, plus a trainable random mask $\phi_{k+1} = \{S_{k+1}^1 \ldots S_{k+1}^L\}$. The intuition is that strong linear combination parameters can be discovered quickly if similarities are strong, otherwise more weight is placed on the newly trained mask.

A new mask at layer $l$ is given by
\begin{equation}
    \label{eqn:linear_comb}
    S^{l, lc} = \beta_{k+1}^{l}S_{k+1}^{l} + \sum_{i=1}^{k}\beta_i^{l} S_{i}^{l, *}
\end{equation}
where $S_{k+1}^{l} \in \phi_{k+1}$ are the scores  of layer $l$ in task $k+1$, $S^{l, lc}$ denotes the transformed scores for task $k+1$ after the linear combination (lc) operation, $S_i^{l, *}$ denotes the optimal scores for previously learned task $i$, and $\beta_1^{l}, \ldots \beta_{k+1}^{l}$ are the weights of the linear combination (at layer $l$). To maintain a normalized weighted sum, a $\mathrm{Softmax}$ function is applied to the linear combination parameters before they are applied in Equation \ref{eqn:linear_comb}. When no knowledge is present in the system, the first task is learned starting with a random mask. Task two is learned using $\bar{\beta_1} = 0.5$, weighting task one's mask, and $\bar{\beta_2} = 0.5$, weighting the new random mask. The third task will have $\bar{\beta_1} = \bar{\beta_2} = \bar{\beta_3} = 0.33$, and so on. Note that $\bar{\beta_k}$ denotes a vector of size $L$ which contains the co-efficient parameters for task $k$ across all $L$ layers of the network (i.e., $\bar{\beta_k} = \{\beta_k^1 \ldots \beta_k^L\}$).

In short, two approaches can be devised. The first one in which each mask is learned independently of the others. Experimentation of this approach will determine the baseline capabilities of modulating masks in LRL. We name this \emph{Mask Random Initialization} ($\mathrm{Mask_{RI}}$).
\begin{figure}
    \centering
    \includegraphics[width=\textwidth]{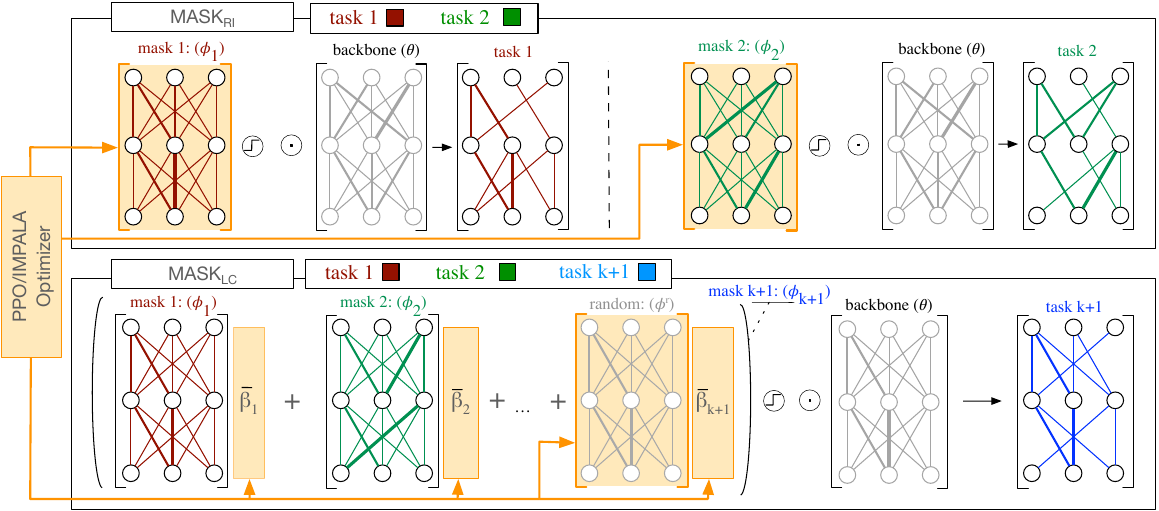}
    \caption{Graphical representations of the methods $\mathrm{Mask_{RI}}$ (top) and $\mathrm{Mask_{LC}}$ (bottom). $\mathrm{Mask_{RI}}$ searches for a new mask for each new task independently. $\mathrm{Mask_{LC}}$ attempts to exploit previous knowledge to learn a new task: known masks are linearly combined with a new randomly initialized mask while learning a new task. Gradient updates search for the parameters $\ \bar{\beta_1},...,\bar{\beta_{n}}$ and the new random mask.}
    \label{fig:method}
\end{figure} 
The second approach attempts to exploit the knowledge acquired so far during the curriculum learning by using a linear combination of masks to learn a new one. Experimentation of this second approach will determine the capabilities of modulating masks to exploit previously learned knowledge. We name this second approach \emph{Mask Linear Combination} ($\mathrm{Mask_{LC}}$). The idea is graphically summarized in Figure \ref{fig:method}.

It can be noted that, as the number of known tasks increases, the relative weight of each mask decreases in $\mathrm{Mask_{LC}}$. This could be a problem, particularly as the weight of the new random mask is reduced, possibly biasing the search excessively towards an average of previous policies. Therefore, we introduce a third approach that attempts to combine the benefits of both $\mathrm{Mask_{RI}}$ and $\mathrm{Mask_{LC}}$: we set the initial weight of the new random mask to 0.5, while the remaining 0.5 weight is shared by the masks of all known tasks. We name this third approach  \emph{Balanced Linear Combination} ($\mathrm{Mask_{BLC}}$). It must be noted that the difference between $\mathrm{Mask_{LC}}$ and $\mathrm{Mask_{BLC}}$ is only in the initialization of weights when a new task is encountered, where $\bar{\beta}_{k+1} = 1/(k+1)$ in $\mathrm{Mask_{LC}}$ and $\bar{\beta}_{k+1} = 0.5$ in $\mathrm{Mask_{BLC}}$. However, the parameters $\bar{\beta}$ can be modified arbitrarily by backpropagation during training.

For both $\mathrm{Mask_{LC}}$ and $\mathrm{Mask_{BLC}}$, updates are made only to $S_{k+1}^{l}$ and $\bar{\beta}_i^{l} \ldots \bar{\beta}_{k+1}^{l}$ across each layer $l$. After the training on task $k+1$ is completed, the knowledge from the linear combination is consolidated into the scores for the current task $S_{k+1}^{l}$ by applying Equation \ref{eqn:linear_comb}. Therefore, the other masks and the linear combination parameters are no longer required. Algorithm \ref{alg:forward_pass} reports  the forward pass operations for $\mathrm{Mask_{LC}}$.
\begin{algorithm}
    \caption{Forward pass in network layer $l$ in $\mathrm{Mask_{LC}}$}
    \label{alg:forward_pass}
    \begin{algorithmic}[1]
        \Require Number of task learned so far $k$
        \Require $S_1^* \ldots S_{k}^*$, $S_{k+1}$, $\bar{\beta}_1 \ldots \bar{\beta}_{k+1}$, $W$
        \Procedure{ForwardPass}{$\mathbf{x}$}
        \If {$k \leftarrow 0$}
            \State Set $P \leftarrow S_1$
        \Else
            \State Compute the task score from linear combination: $P \leftarrow \bar{\beta}_{k+1}S_{k+1} + \sum_{i=1}^{k}\bar{\beta}_i S_{i}^*$
        \EndIf
        \State Compute mask from score $M_{k+1} \leftarrow g(P)$
        \State Modulate weight $W^{mod} \leftarrow (W \odot M_{k+1})$
        \State Compute output $\mathbf{y} \leftarrow W^{mod} \cdot \mathbf{x}$
        \State \textbf{return} $\mathbf{y}$
        \EndProcedure
    \end{algorithmic}
\end{algorithm}

\subsection{Continuous Modulatory Masks for Continuous RL problems}
\label{subsec:continuous-masks}
In previous studies, binarization of masks was discovered to be effective in classification to reduce the memory requirement and improve scalability. As shown in our simulations, this approach was effective also in discrete RL problems. However, in continuous action space RL environments, we discovered that binary masks did not lead to successful learning. It is possible that the quantization operation hurts the loss landscape in such environments since the input-output mapping is continuous to continuous values. Therefore, a modification of the supermask method can be devised to support the use of continuous value masks. In the thresholding mask generation operation (given a threshold $\epsilon$), the modified version becomes
\begin{equation}
    \label{eqn:score-to-continuous-mask}
    g(\phi_k) =
      \begin{cases}
        \phi_{k, \{i, j\}} & \phi_{k, \{i, j\}} > \epsilon \\
        0 & \text{otherwise}\quad.
      \end{cases}       
\end{equation}

Such a modification still maintains the ability to learn sparse masks, but replaces the unitary positive values with continuous values. The results of the empirical investigation of the binary and continuous masks in a continuous action space environment is reported in Section \ref{subsec:exp_cw10}.

\section{Experiments }
\label{sec:experiments}
The three novel approaches, $\mathrm{Mask_{RI}}$, $\mathrm{Mask_{LC}}$ and $\mathrm{Mask_{BLC}}$, are tested on a set of LRL benchmarks across discrete and continuous action space environments. A complete set of hyper-parameters for each experiment is reported in Appendix \ref{apndx:hyperparameters}. 

First, we employed the ProcGen benchmark \citep{cobbe2020procgen} that consists of a set of video games with high diversity, fast computation, procedurally generated scenarios, visual recognition and motor control. The masking methods were implemented with IMPALA: $\mathrm{Mask_{RI/LC/BLC}+IMPALA}$ \citep{espeholt2018impala} and tested following the ProcGen lifelong RL curriculum presented in \citet{powers2022cora} (a subset of games in ProcGen). The properties of the benchmark make the curriculum challenging (e.g., high dimensional RGB observations and procedural generations of levels). The curriculum was designed to test generalization abilities: for each task, the lifelong evaluations are carried out using procedurally generated levels that are unseen by the agents during training. The masking methods were compared with baselines such as online EWC, P\&C \citep{schwarz2018pnc}, CLEAR \citep{rolnick2019clear}, and IMPALA.

With the aim of understanding the learning mechanism and dynamics of the masking approaches, other benchmarks were chosen to assess the robustness of the method against the following aspects: discrete and continuous environments; variations across tasks in input, transition and reward distributions. The CT-graph \citep{soltoggio2019ctgraph, soltoggio2023ctgraph} (sparse reward, fast, scalable to large search spaces, variation of reward), Minigrid \citep{gym_minigrid, chevalier2023minigrid} (variation of input distributions and reward functions) and Continual World \citep{wolczyk2021continualworld} (continuous robot-control) were used to assess the approaches, with PPO serving as the base RL algorithm. In these benchmarks, the masking approaches ($\mathrm{Mask_{RI/LC/BLC}+PPO}$) are compared with a lifelong learning baseline, online EWC multi-head ($\mathrm{EWC_{MH}+PPO}$), and with the non-lifelong learning algorithm PPO. Experiments with a PPO single task expert (STE) were also conducted to enable the computation of the forward transfer metric for each method, following the setup in \citet{wolczyk2021continualworld}. Control experiments with EWC single head, denoted as $\mathrm{EWC_{SH}}$, performed poorly as a confirmation that our benchmarks contain interfering tasks \citep{kessler2022owl}: we report those results in the Appendix \ref{apndx:ewc-sh-mh}.

The metrics report a lifelong evaluation across all tasks at different points during the lifelong training, computed as the average sum of reward obtained across all tasks in the curriculum. The area under the curve (AUC) is reported in corresponding tables. A forward transfer metric, following the formulation employed in \citet{wolczyk2021continualworld}, is computed for the CT-graph, Minigrid and Continual World. For each task, the forward transfer is computed as the normalized difference between the AUC of the training plot for the lifelong learning agent and the AUC for the reference single task expert. For the ProcGen experiments, the lifelong training and evaluation plot format reported in \citet{powers2022cora} was followed to enable an easier comparison with the results in the original paper. As tasks are learned independently of other tasks in $\mathrm{Mask_{RI}}$, there is no notion of forward transfer in the method. Therefore, the method is omitted when forward the transfer metrics are reported. 

The results presented in the evaluation plots and the total evaluation metric reported in the tables below were computed as the mean of the seed runs per benchmark, with the error bars denoting the 95\% confidence interval. The \emph{CT8}, \emph{CT12}, \emph{CT8 multi depth}, and \emph{MG10} results contained 5 seed runs per method, while the \emph{CW10} results contained 3 seed runs due to its high computational requirement. While the sample size for the evaluation metric is the number of seed runs, the sample size used for computing the mean and 95\% confidence intervals for the forward transfer metric and the training plots is the number of seeds multiplied by the number of tasks per curriculum (i.e., 40, 60, 40, 50, 30 for \emph{CT8}, \emph{CT12}, \emph{CT8 multi depth}, \emph{MG10}, and \emph{CW10} respectively). A significance test was conducted on the results obtained to validate the performance difference between algorithms. The result of the test is reported in Appendix \ref{apndx:significance-testing}.

\subsection{ProcGen}
\label{subsec:exp_procgen}
The experimental protocol employed follows the setup of \citet{powers2022cora}, with a sequence of six tasks ($\mathrm{0-Climber}$, $\mathrm{1-Dodgeball}$, $\mathrm{2-Ninja}$,
$\mathrm{3-Starpilot}$, $\mathrm{4-Bigfish}$, $\mathrm{5-Fruitbot}$) with five learning cycles. Screenshots of the games are reported in the Appendix (\ref{apndx:env-procgen}). Several environment instances, game levels, and variations in objects, texture maps, layout, enemies, etc., can be generated within a single task. For each task, agents are trained on 200 levels. However, the evaluation is carried out on the distribution of all levels, which is a combination of levels seen and unseen during training.

\begin{figure}[ht]
    \centering
    \includegraphics[width=0.95\textwidth]{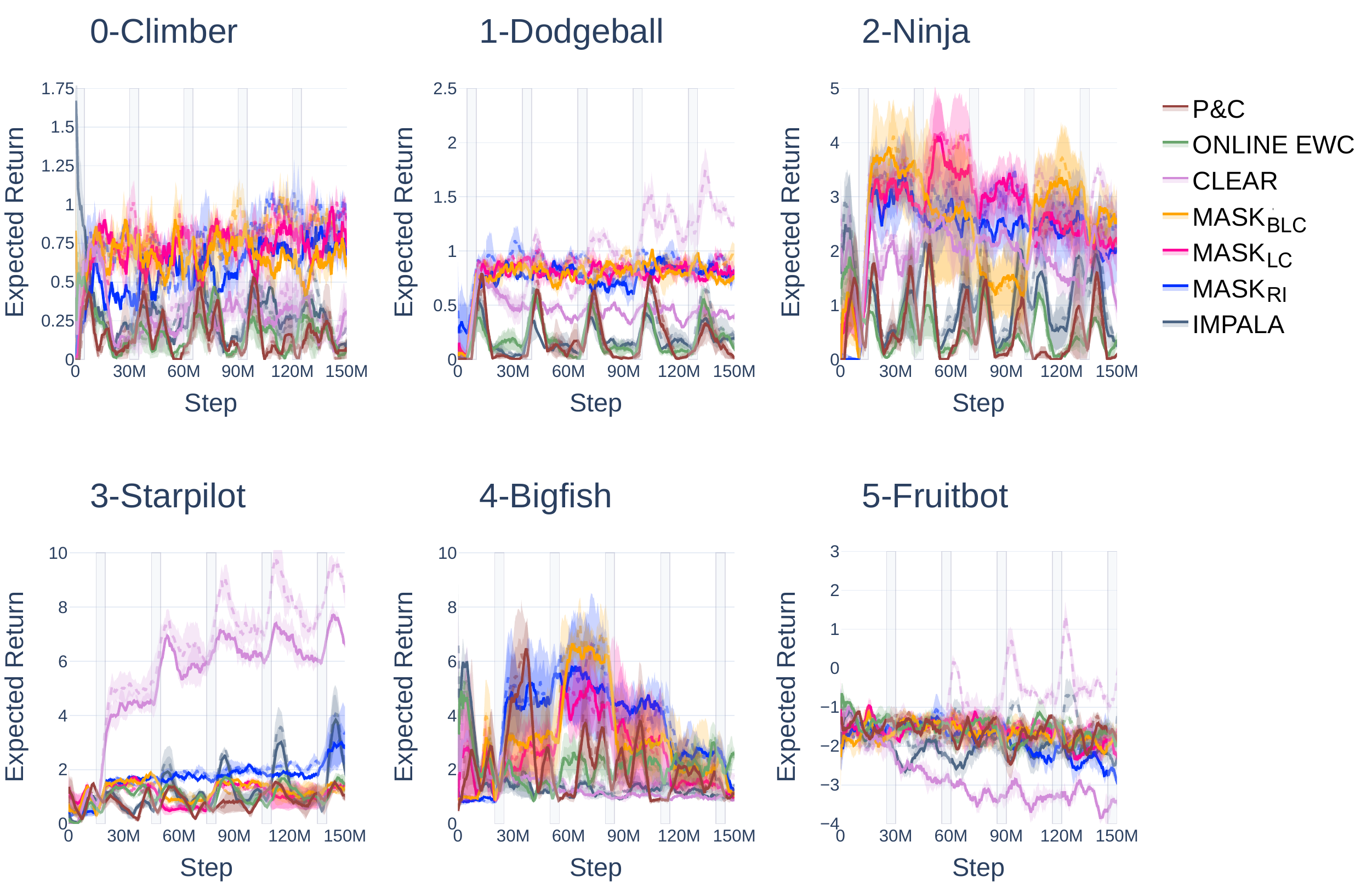}
    \caption{Evaluation results in the ProcGen environment ($6$ tasks, $5$ cycles), measured as average across 3 runs. The solid line represents evaluation on unseen environments, while the dotted line represents evaluation on the training environments. The gray shaded rectangles show at what point in time an agent is been trained on each task.}
    \label{fig:procgen}
\end{figure}

IMPALA was used as the base RL optimizer on which we deployed the novel masking methods. The results are reported in Figure \ref{fig:procgen}, following the presentation style used in \citet{powers2022cora}.
The masking methods ($\mathrm{Mask_{RI}}$, $\mathrm{Mask_{LC}}$, and  $\mathrm{Mask_{BLC}}$) show better performance with respect to other baselines across most tasks, while maintaining generalization capabilities across training and evaluation environments. As the tasks are visually diverse, possibly resulting in less similarity across tasks, reusing previous knowledge may not offer much advantage. Nevertheless, the evaluation performance for each method reported in Table \ref{tab:procgen_tbl_eval_metrics} illustrates a significant advantage of the masking methods with respect to the baselines, particularly in the test tasks, where $\mathrm{Mask_{LC}}$ is 44\% better than the closest runner up (CLEAR) and over 300\% better than P\&C.

\begin{table}
    \centering
    \begin{tabular}{lcc}
        \hline
        {} & \multicolumn{2}{c}{Evaluation Performance} \\
        Method  & Train Tasks & Test Tasks \\
        \hline
        IMPALA & 3687.57 $\pm$ 1194.23 & 2818.44 $\pm$ 1214.34 \\
        Online EWC & 4645.87 $\pm$ 1963.42 & 3831.94 $\pm$ 1258.38 \\
        P\&C & 3403.59 $\pm$ 3428.37 & 3390.97 $\pm$ 1795.58 \\
        CLEAR & 11706.34 $\pm$ 5271.45 & 8447.64 $\pm$ 2286.21 \\
        $\mathrm{Mask_{RI}}$ & 12462.82 $\pm$ 1057.68 & 12222.10 $\pm$ 6064.18 \\
        $\mathrm{Mask_{LC}}$ & 12488.31 $\pm$ 3416.90 & 12377.77 $\pm$ 1440.80 \\
        $\mathrm{Mask_{BLC}}$ & 11683.21 $\pm$ 3172.61 & 11913.48 $\pm$ 3869.95 \\
       \hline
    \end{tabular}
    \caption{Total evaluation return (AUC of the lifelong evaluation plot in Figure \ref{fig:procgen}) on the train and test tasks in ProcGen curriculum. Mean $\pm$ 95\% confidence interval reported.}
    \label{tab:procgen_tbl_eval_metrics}
\end{table}

\subsection{CT-graph}
\label{subsec:exp_CT-graph}
\begin{figure}
    \centering
    \begin{tabular}{cc}
    \includegraphics[width=0.49\textwidth]{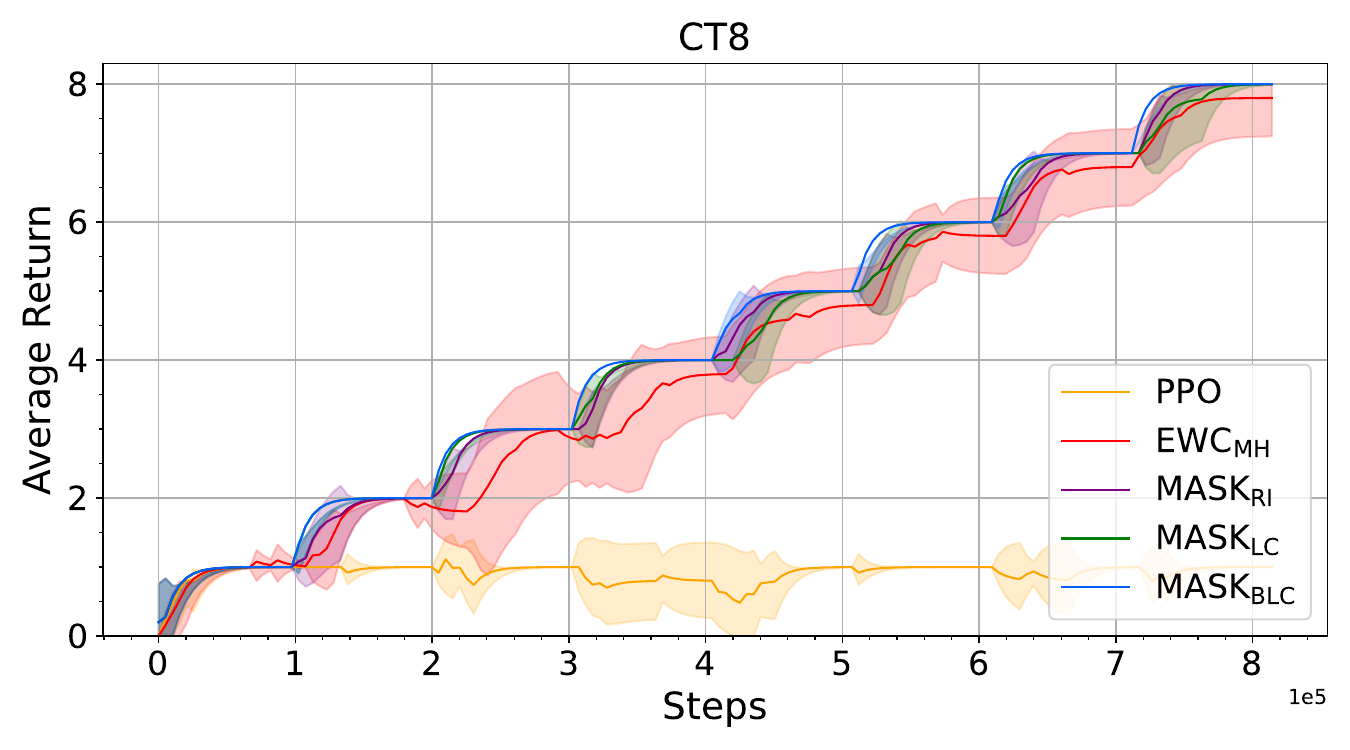}&
       \includegraphics[width=0.49\textwidth]{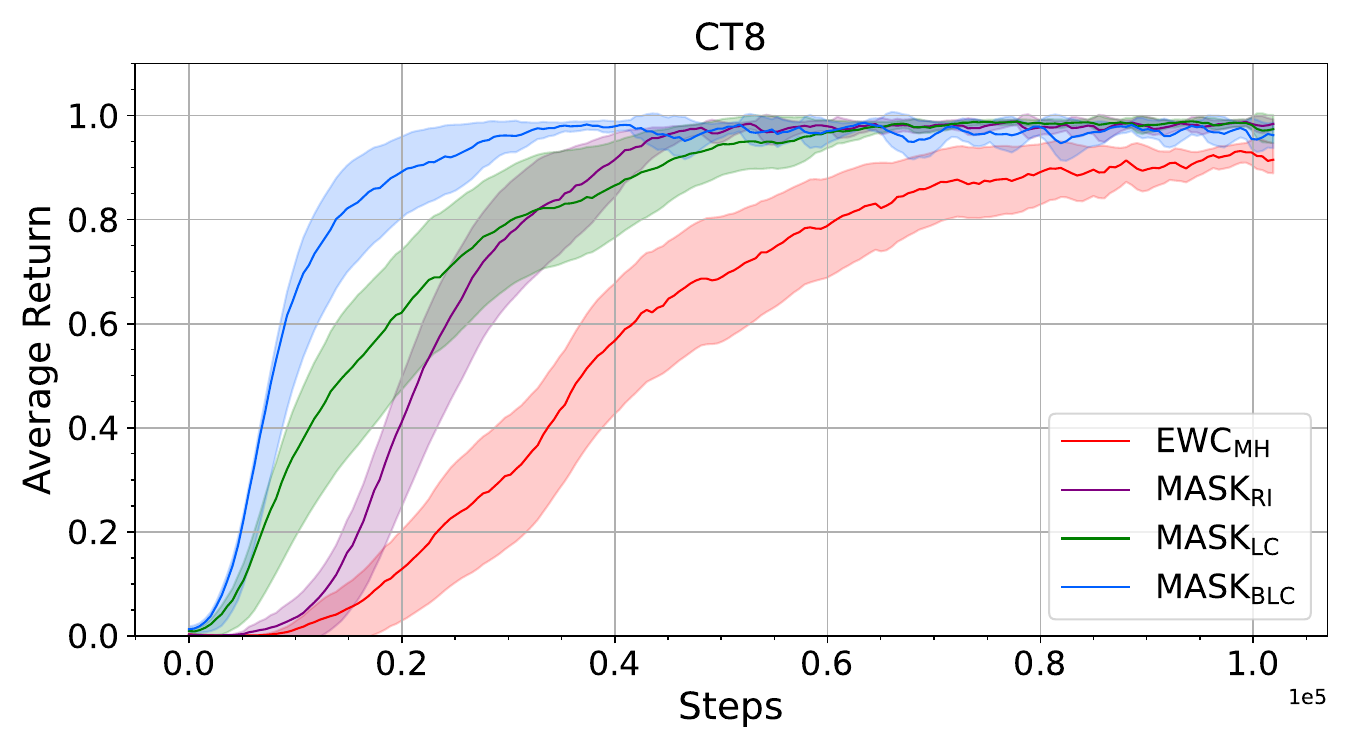}\vspace{-10pt}
    \end{tabular}
      \caption{
      Performance in the \emph{CT8} curriculum. (Left) Lifelong evaluation performance on all tasks (mean and 95\% confidence interval on 5 seeds/samples). (Right) Training performance on each task, measured as the average return across all tasks and seeds runs (mean and 95\% confidence interval on 8 tasks and 5 seeds, i.e., 40 samples). In the lifelong evaluation, a clear difference can be noted between the lifelong learning algorithms and the non-lifelong learning algorithm PPO, which is therefore not reported in the training plot (Right). The numerical values for the AUC are reported in Tables \ref{res:ct8_tbl_metrics}. The training curves show how $\mathrm{Mask_{BLC}}$ is on average faster at learning new tasks, followed by $\mathrm{Mask_{LC}}$ and $\mathrm{Mask_{RI}}$. The performance on each individual task is reported in the Appendix in Figure \ref{res:ct_mg_cw_plot_train_all_methods}
      }
    \label{res:ct8_plot_eval_train}
\end{figure}

Each instance of the CT-graph environment contains a start state followed by a number of states (2D patterned images) that lead to leaf states and rewards only with optimal policies. A task is defined by setting one of the leaf states as a desired goal state that can be reached only via one trajectory. Two experimental setups were employed: the first with $8$ leaf states (reward locations) that serves for $8$ tasks, denoted as \emph{CT8} curriculum (depth-3, breadth-2 graph with $2^3=8$ leaves). In the second setup, two graph instances with 4 and 8 different reward locations, with depth 2 and 3 respectively, result in combined curriculum of 12 tasks, denoted as \emph{CT12}. Such tasks have levels of similarities due to similar input distributions, but also interfere due to opposing reward functions and policies for the same inputs. Additionally, the $8$-task graph has a longer path to the reward that introduces variations in both the transition and reward functions. Graphical illustrations of the CT-graph instances are provided in Appendix \ref{apndx:environments}. 

\begin{figure}
    \centering
       \begin{tabular}{cc}
       \includegraphics[width=0.49\textwidth]{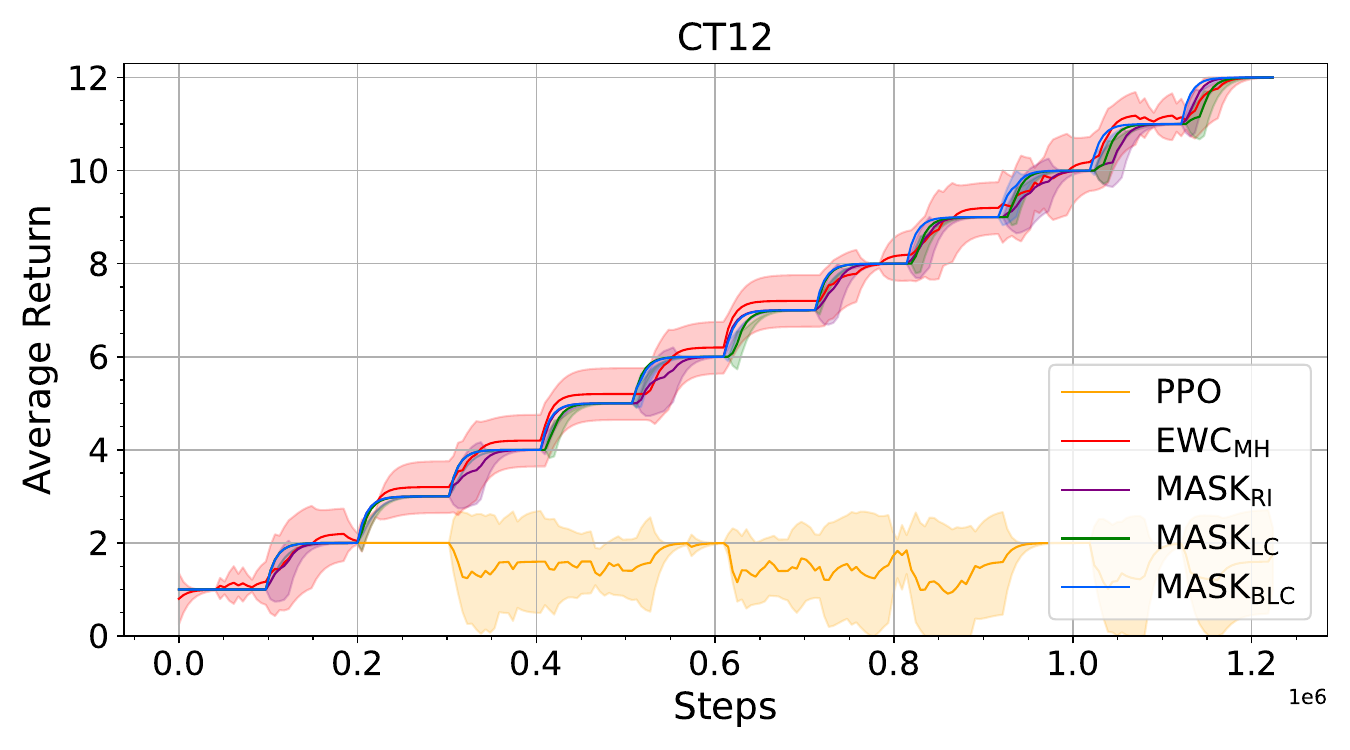}
  &       \includegraphics[width=0.49\textwidth]{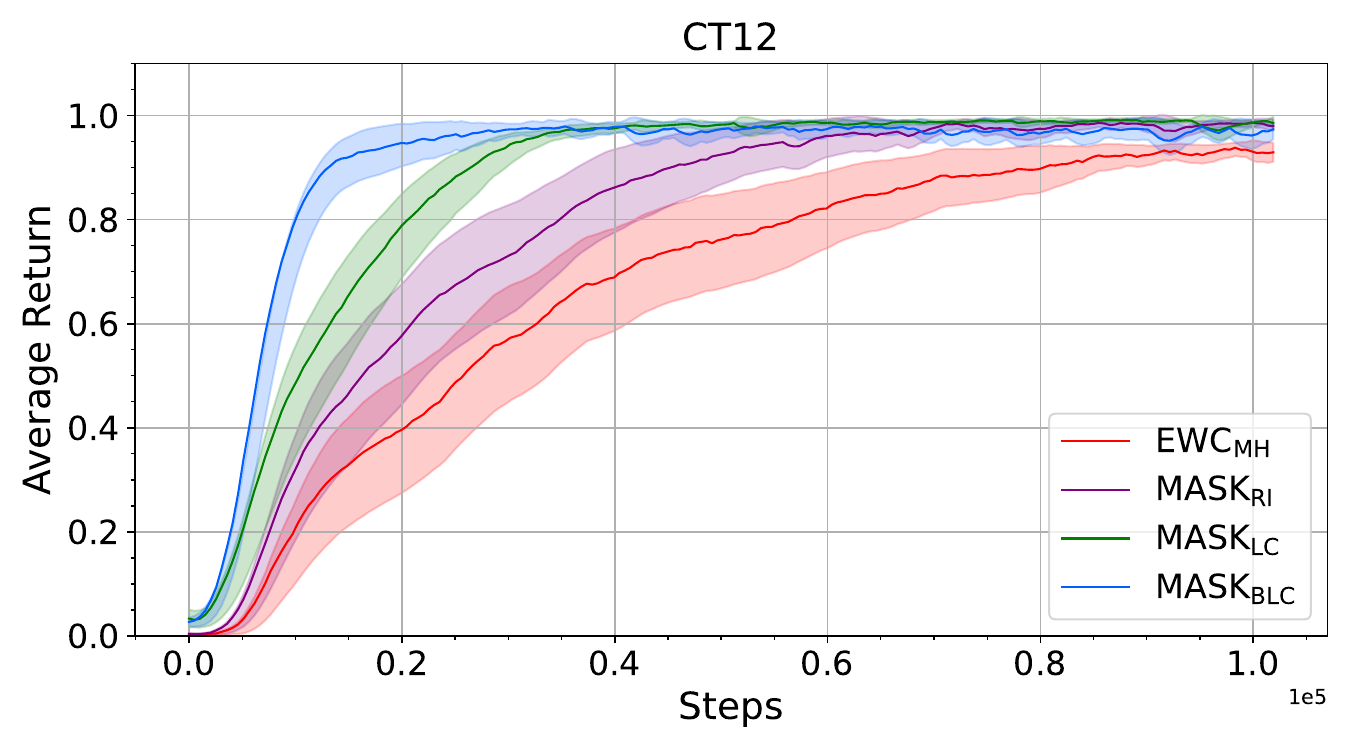}\vspace{-10pt}
    \end{tabular}
    \caption{
      Performance in the \emph{CT12} curriculum. (Left) Lifelong evaluation performance on all tasks (mean and 95\% confidence interval on 5 seeds/samples). (Right) Training performance on each task, measured as the average return across all tasks and seeds runs (mean and 95\% confidence interval on 12 tasks and 5 seeds, i.e., 60 samples).
      }     
    \label{res:ct12_plot_eval_train}
\end{figure}

Each task is trained for 102.4K time steps. Figures \ref{res:ct8_plot_eval_train} and \ref{res:ct12_plot_eval_train} report evaluations in the \emph{CT8} and \emph{CT12} curricula as agents are sequentially trained across tasks. The forward transfer and the total evaluation performance metrics are presented in Tables \ref{res:ct8_tbl_metrics} and \ref{res:ct12_tbl_metrics} respectively.

\begin{minipage}{\textwidth}
  \centering
  \begin{minipage}{0.45\textwidth}
    \centering
    \begin{tabular}{lcc}\hline
      Method & Total Eval & Fwd Trnsf. \\ \hline
      PPO & 147.00 $\pm$ 13.63 & 0.15 $\pm$ 0.20 \\
      $\mathrm{EWC_{MH}}$ & 661.20 $\pm$ 64.14 & -0.23 $\pm$ 0.19 \\
      $\mathrm{Mask_{RI}}$ & 701.20 $\pm$ 6.11 & -- \\
      $\mathrm{Mask_{LC}}$ & 702.00 $\pm$ 5.89 & 0.46 $\pm$ 0.13 \\
      $\mathrm{Mask_{BLC}}$ & 716.80 $\pm$ 2.22 & 0.67 $\pm$ 0.06 \\ \hline
    \end{tabular}
    \captionsetup{type=table}
    \captionof{table}{Total evaluation return (AUC of the lifelong evaluation plot in Figure \ref{res:ct8_plot_eval_train}(Left)) and forward transfer during lifelong training in the \emph{CT8}. Mean $\pm$ 95\% confidence interval reported.}
    \label{res:ct8_tbl_metrics}
  \end{minipage}
  \hspace{1.5em}
  \begin{minipage}{0.45\textwidth}
    \centering
    \begin{tabular}{lcc}\hline
      Method & Total Eval & Fwd Trnsf. \\ 
      \hline
      PPO & 378.40 $\pm$ 22.27 & -0.15 $\pm$ 0.18 \\
      $\mathrm{EWC_{MH}}$ & 1565.00 $\pm$ 92.36 & -0.11 $\pm$ 0.14 \\
      $\mathrm{Mask_{RI}}$ & 1535.00 $\pm$ 10.20 & -- \\
      $\mathrm{Mask_{LC}}$ & 1544.60 $\pm$ 4.17 & 0.50 $\pm$ 0.08 \\ 
      $\mathrm{Mask_{BLC}}$ & 1558.20 $\pm$ 3.09 & 0.65 $\pm$ 0.04 \\ 
      \hline
    \end{tabular}
    \captionsetup{type=table}
    \captionof{table}{Total evaluation return (AUC of the lifelong evaluation plot in Figure \ref{res:ct12_plot_eval_train}(Left)) and forward transfer during lifelong training in the \emph{CT12}. Mean $\pm$ 95\% confidence interval reported.}
    \label{res:ct12_tbl_metrics}
  \end{minipage}
\end{minipage}

The plots show that the masking methods ($\mathrm{Mask_{RI}}$, $\mathrm{Mask_{LC}}$, and $\mathrm{Mask_{BLC}}$) are capable of avoiding forgetting and obtain high evaluation performance, with significantly better forward transfer in comparison to $\mathrm{EWC_{MH}}$ and PPO. On average, the $\mathrm{Mask_{LC}}$ approach recovers performance faster than $\mathrm{Mask_{RI}}$, and $\mathrm{Mask_{BLC}}$ performs best. An expanded version of the training plots showing the learning curves per tasks and averaged across seed runs is reported in the Appendix \ref{apndx:train-plots-all-methods}.

\subsection{Minigrid}
\label{subsec:minigrid}
\begin{figure}
    \centering
    \begin{tabular}{cc}
        \includegraphics[width=0.49\textwidth]{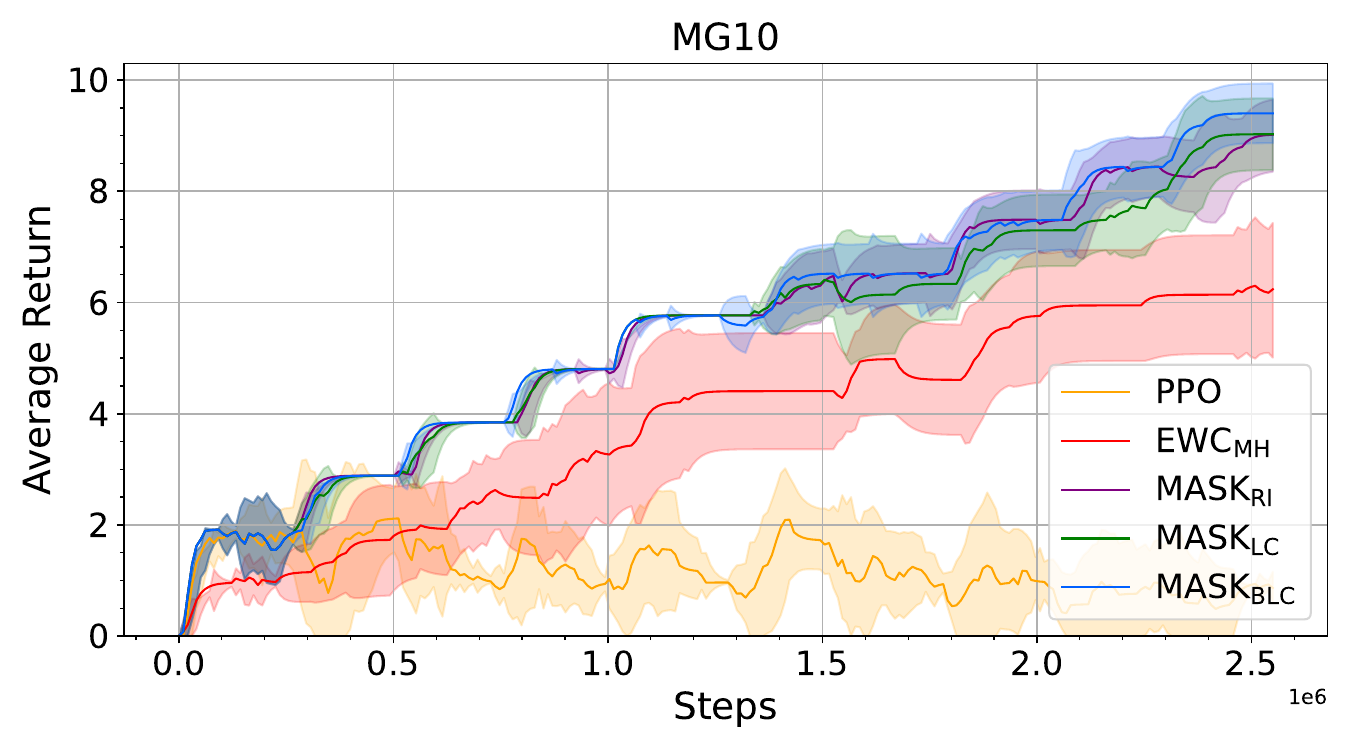}&
     \includegraphics[width=0.49\textwidth]{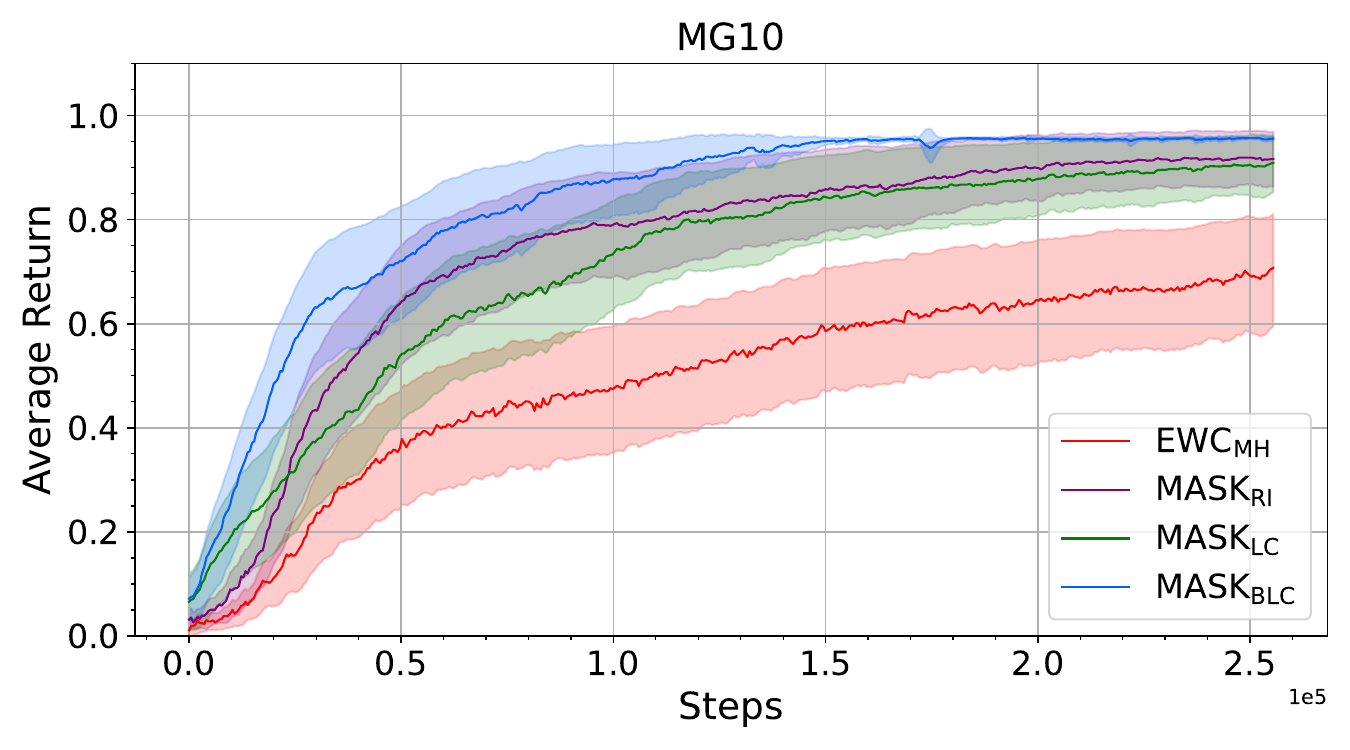}\\
    \end{tabular}
    \caption{Performance in the \emph{MG10} curriculum. (Left) Lifelong evaluation performance on all tasks (mean and 95\% confidence interval on 5 seeds/samples). (Right) Training performance on each task, measured as the average return across all tasks and seeds runs (mean and 95\% confidence interval on 10 tasks and 5 seeds, i.e., 50 samples).}
    \label{res:mg10_plot_eval_train}
\end{figure}

The experiment protocol employs a curriculum of ten tasks (referred to as \emph{MG10}), which consist of two variants of each of the following: $\mathrm{SimpleCrossingS9N1}$, $\mathrm{SimpleCrossingS9N2}$, $\mathrm{SimpleCrossingS9N3}$, $\mathrm{LavaCrossingS9N1}$, $\mathrm{LavaCrossingS9N2}$. Screenshots of all tasks are reported in the Appendix (\ref{apndx:env-minigrid}). The variations across tasks include change in the state distribution and reward function. The results for the \emph{MG10} experiments are presented in Figure \ref{res:mg10_plot_eval_train} and Table \ref{res:mg10_tbl}. The masking methods obtained better performance in comparison to the baselines, with $\mathrm{Mask_{BLC}}$ obtaining the best performance. Appendix \ref{apndx:train-plots-all-methods} provides a full experimental details.

\begin{table}
    \centering
    \begin{tabular}{lcc}\hline
      Method & Total Eval & Fwd Trnsf. \\ 
      \hline
      PPO & 298.69 $\pm$ 24.12 & -0.40 $\pm$ 0.33 \\
      $\mathrm{EWC_{MH}}$ & 959.38 $\pm$ 141.32 & -1.04 $\pm$ 0.32 \\
      $\mathrm{Mask_{RI}}$ & 1379.71 $\pm$ 51.86 & -- \\
      $\mathrm{Mask_{LC}}$ & 1358.08 $\pm$ 75.29 & -0.25 $\pm$ 0.30 \\ 
      $\mathrm{Mask_{BLC}}$ & 1405.30 $\pm$ 62.49 & 0.27 $\pm$ 0.14 \\ 
      \hline
    \end{tabular}
    \caption{Total evaluation return (AUC of the lifelong evaluation plot in Figure \ref{res:mg10_plot_eval_train}(Left)) and forward transfer in the \emph{MG10}. Mean $\pm$ 95\% confidence interval reported.}
    \label{res:mg10_tbl}
\end{table}

\subsection{Continual World}
\label{subsec:exp_cw10}
\begin{figure}
    \centering
    \begin{tabular}{cc}
    \includegraphics[width=0.49\textwidth]{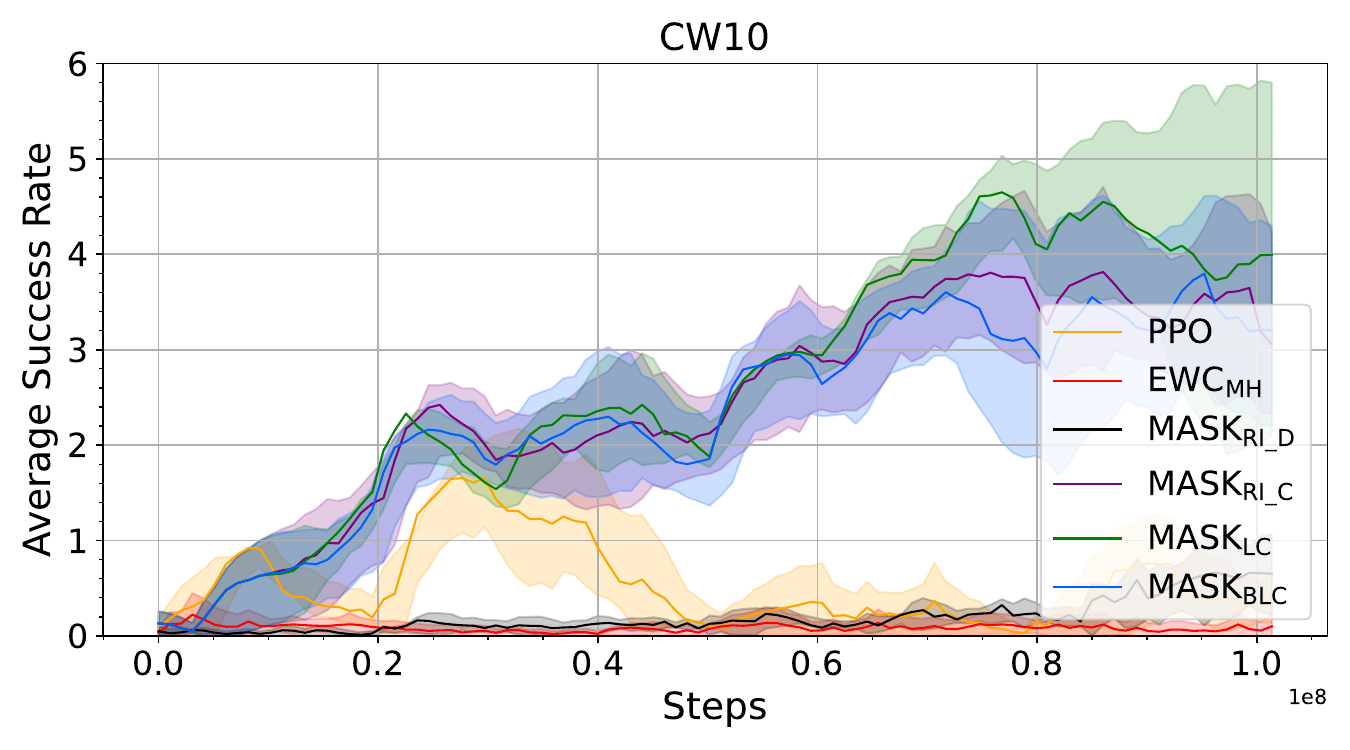}&
    \includegraphics[width=0.49\textwidth]{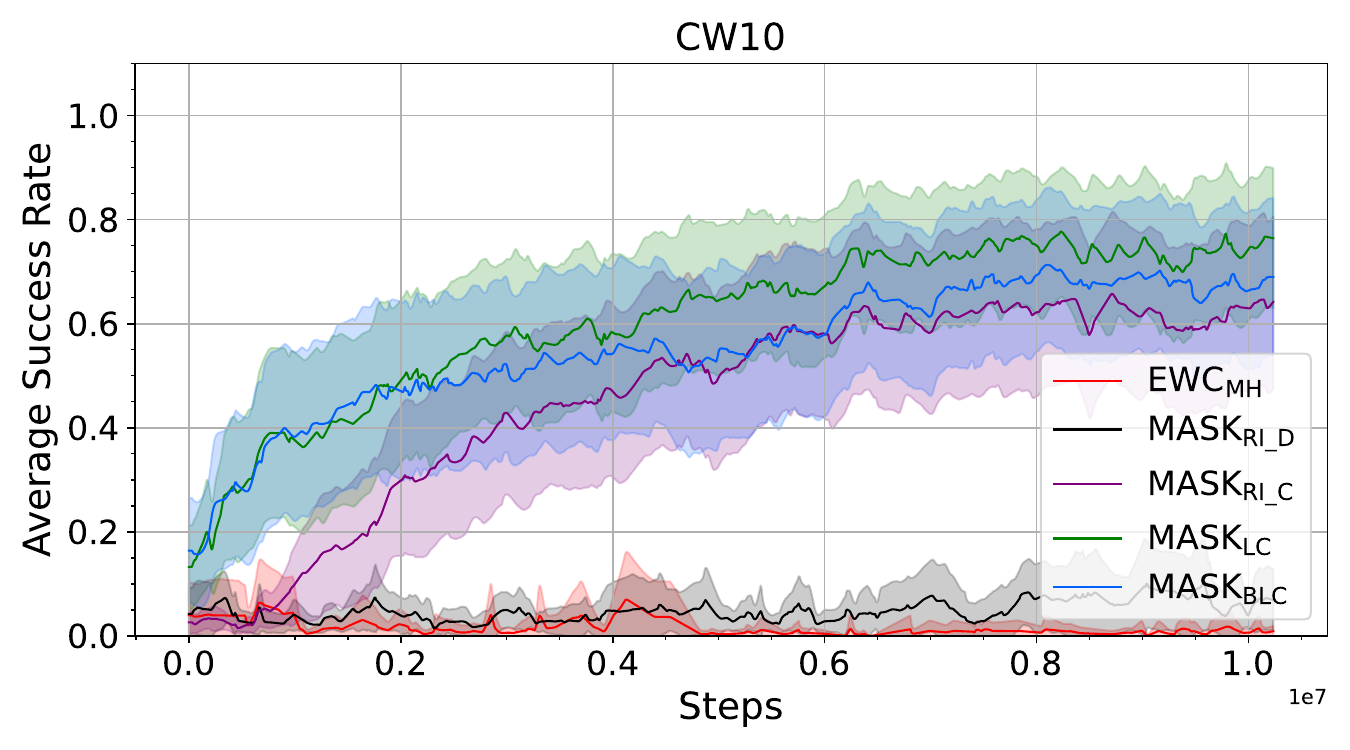}
    \end{tabular}
    \caption{Performance in the \emph{CW10} curriculum measured using the success rate metric. (Left) Lifelong evaluation performance on all tasks (mean and 95\% confidence interval on 3 seeds/samples). (Right) Training performance on each task, measured as the average success rate across all tasks and seeds runs (mean and 95\% confidence interval on 10 tasks and 3 seeds, i.e., 30 samples).}
    \label{res:cw10_plot_eval_train}
\end{figure}

We evaluated the novel methods in a robotics environment with continuous action space, the Continual World \citep{wolczyk2021continualworld}. The environment was adapted from the MetaWorld environment \citep{yu2020meta} that contains $50$ classes of robotics manipulation tasks. The $\mathrm{CW10}$ curriculum consists of $10$ robotics tasks: visual screenshots are provided in the Appendix (\ref{apndx:env-continual-world}). The results for all methods were measured using the success rate metric introduced in \citet{yu2020meta}, which awards a 1 if an agent solves a task or 0 otherwise. For the  masking methods in this curriculum, the standard quantization of masks into binary performs poorly. To demonstrate this, two variants are run: the standard setting, where a binary mask is derived from the scores, denoted as $\mathrm{Mask_{RI\_D}}$, and another where a continuous mask is derived from the scores (discussed in Section \ref{subsec:continuous-masks}), denoted as $\mathrm{Mask_{RI\_C}}$. The results from Figure \ref{res:cw10_plot_eval_train} and Table \ref{res:cw10_tbl} 
show that $\mathrm{Mask_{RI\_C}}$ performs significantly better than $\mathrm{Mask_{RI\_D}}$. Motivated by the results, the linear combination of masks method  $\mathrm{Mask_{LC}}$ presented for this curriculum also employed the use of continuous masks. $\mathrm{Mask_{LC}}$ and $\mathrm{Mask_{BLC}}$ performed markedly better than the baseline $\mathrm{EWC_{MH}}$ that appears to struggle on this benchmark. Appendix \ref{apndx:train-plots-all-methods} reports the details for all methods.

\begin{table}
    \centering
    \begin{tabular}{lcc}\hline
      Method & Total Eval & Fwd Trnsf. \\ 
      \hline
      PPO & 53.83 $\pm$ 72.32 & -4.06 $\pm$ 2.58 \\
      $\mathrm{EWC_{MH}}$ & 8.60 $\pm$ 12.91 & -7.39 $\pm$ 3.76 \\
      $\mathrm{Mask_{RI\_D}}$ & 20.45 $\pm$ 51.46 & -- \\
      $\mathrm{Mask_{RI\_C}}$ & 246.83 $\pm$ 157.39 & -- \\
      $\mathrm{Mask_{LC}}$ & 272.43 $\pm$ 124.92 & -0.33 $\pm$ 0.36 \\ 
      $\mathrm{Mask_{BLC}}$ & 237.00 $\pm$ 167.25 & -0.61 $\pm$ 0.52 \\ 
      \hline
    \end{tabular}
    \caption{Total evaluation success metric (AUC of the lifelong evaluation plot in Figure \ref{res:cw10_plot_eval_train}(Left)) and forward transfer in the \emph{CW10}. Mean $\pm$ 95\% confidence interval reported.}
    \label{res:cw10_tbl}
\end{table}

\section{Analysis}
\label{sec:analysis-and-discussion}
The results of the previous section prompt the following questions: what linear coefficients emerge after learning? How is rapid learning in $\mathrm{Mask_{LC}}$ achieved? How is knowledge reused?

Another interesting question that arise is the training efficiency derived from knowledge reuse in mask methods. The training plots in Figures \ref{res:ct8_plot_eval_train}(Right), \ref{res:ct12_plot_eval_train}(Right), \ref{res:mg10_plot_eval_train}(Right), and \ref{res:cw10_plot_eval_train}(Right) showed that the mask methods required fewer training steps on average to learn tasks in comparison to the baselines. For interested readers, Appendix \ref{apndx:target-performance} reports the analysis conducted on the time (i.e., training steps) taken to reach certain level of performance per task.

\subsection{Coefficients for the linear combination of masks}
\label{subsec:analyis-linear-comb-coeffs}
To validate whether the proposed linear combination process can autonomously discover important masks that are useful for the current task, a visualization of the co-efficients after learning each task is presented.

\begin{figure}
    \centering
    \includegraphics[width=0.75\textwidth]{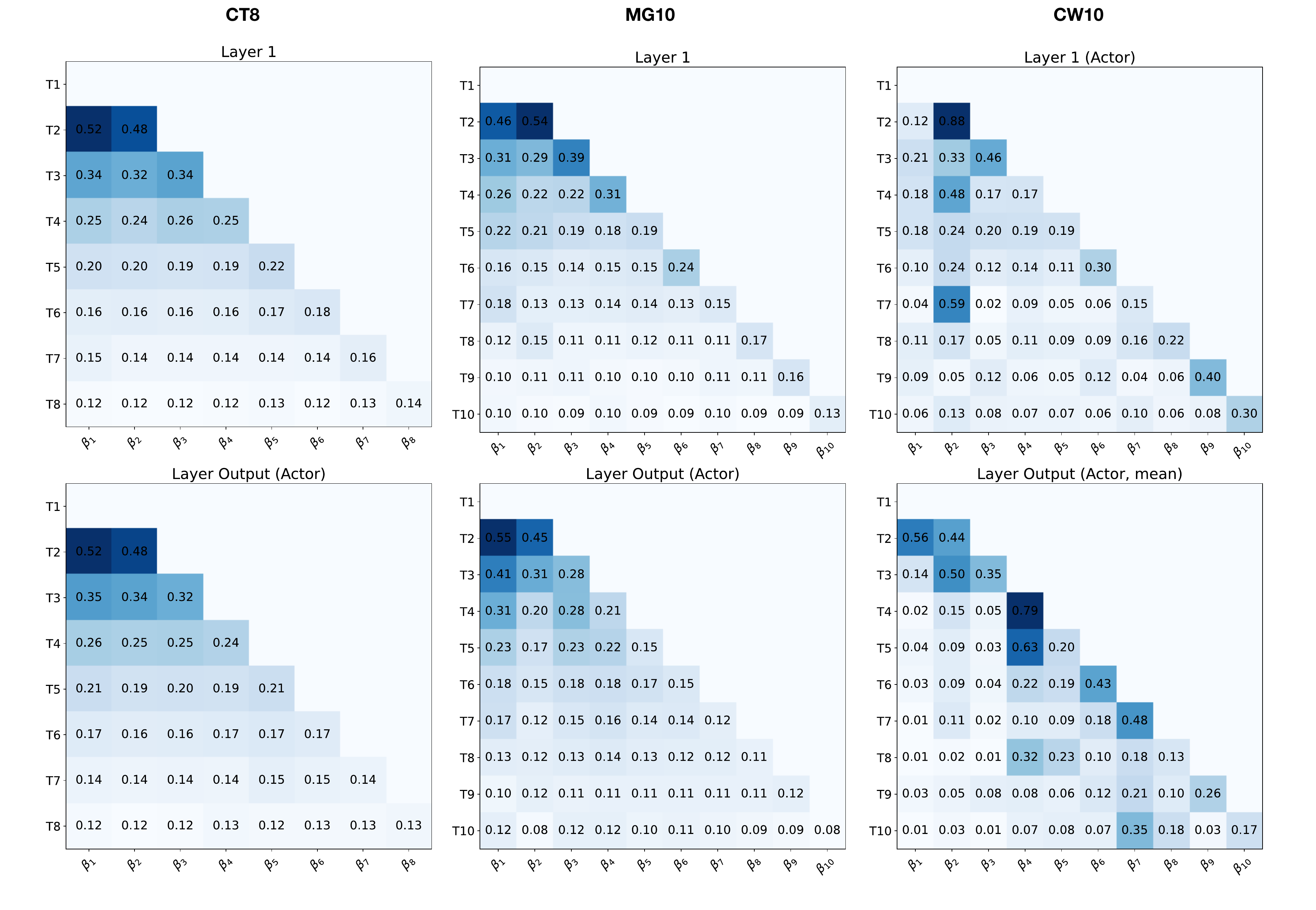}
    \caption{Coefficients $\mathbf{\bar{\beta}}$ in $\mathrm{Mask_{LC}}$ after training on the \emph{CT8}, \emph{MG10}, and \emph{CW10} curricula. Each row represents a task, and the values (which sums to 1) in it are the final set of coefficients after training on the task: the higher the value of a cell, the higher the level of importance of the corresponding mask in the linear combination operation. The figure only shows coefficients for the first layer and the output (actor) layer. See Appendix \ref{apndx:lcomb-coeffs} for plots of all other layers.}
    \label{fig:summary_mask_lc_lcomb_coeff}
\end{figure}

Figure \ref{fig:summary_mask_lc_lcomb_coeff} presents the visualization of the coefficients (for the input and output layers) for a $\mathrm{Mask_{LC}}$ agent trained in the \emph{CT8}, \emph{MG10}, and \emph{CW10} curricula respectively. In each plot, each row reports the final set of coefficients after training on a task. For example, the third row in each plot represents the third task in the curriculum and reports three coefficients used for the linear combination of the masks for two tasks and the new mask. For the first task (first row), there are no previous masks to combine. 

For the \emph{CT8} and \emph{MG10}, the plots show that the coefficients have similar weights for each task (i.e., row wise in Figure \ref{fig:summary_mask_lc_lcomb_coeff}). This observation is consistent across the layers of the network (see Appendix \ref{apndx:lcomb-coeffs} for plots across all layers). The uniform distribution across co-efficients means that the knowledge from all previous tasks is equally important and reused when learning new tasks, possibly indicating that tasks are equally diverse or similar to each other. The knowledge reuse of previous tasks thus accelerates learning and helps the agent quickly achieve optimal performance for the new task. For example, in the CT-graph curricula where navigation abilities are essential to reach the goal, the knowledge on how to navigate/traverse to different parts of the graph is encoded in each previously learned task. Rather than re-learn how to navigate the graph in each task, and subsequently the solve the task, the agent can leverage on the existing navigational knowledge. The performance improvement therefore comes from the fact the agent with knowledge reuse can leverage existing knowledge to learn new tasks quickly. Note that we will not expect any performance improvement if the new task bears no similarity with the previously learned task.

Comparing the values across layers (i.e., column wise  in Figure \ref{fig:summary_mask_lc_lcomb_coeff}), we note that there is little variation. In other words, the standard deviation of each vector $\bar{\beta}$ is low, as all values are similar. From such an observation, it follows that the vector $\bar{\beta}$ could be replaced by a scalar for these two benchmarks.

A different picture emerges from the analysis of the coefficients in the CW10 curriculum (Figure \ref{fig:summary_mask_lc_lcomb_coeff} right-most column). Here the coefficients appear to have a larger standard deviation both across masks and across layers. Particular values may suggest relationships between tasks. E.g., the input layer coefficient $\beta^1_2$ (from layer 1, mask 2) is high in task 4 and 7. Similar diverse patterns can be seen in the output layer. We speculate that tasks in CW10 are more diverse and the optimization process is enhancing specific coefficients to reuse specific skills. The non-uniformity of the co-efficients in the analysis could be a consequence of different levels of task similarities as reported in the forward transfer matrix in \citet{wolczyk2021continualworld} for the CW10.

\subsection{Exploitation of previous knowledge}
\label{subsec:analyis-targeted-exploration}
The linear combination of masks appears to accelerate learning significantly as indicated in Figures \ref{res:ct8_plot_eval_train}(Right), \ref{res:ct12_plot_eval_train}(Right), \ref{res:mg10_plot_eval_train}(Right), and \ref{res:cw10_plot_eval_train}(Right). To investigate the causes of such learning dynamics, we plot the probabilities of actions during an episode in a new task. The idea is to observe what are the softmax probabilities that are provided by the linear combination of masks at the start of learning for a new task. A full analysis would require the unfeasible task of observing all trajectories: we therefore focus on the optimal trajectory by measuring probabilities when traversing such an optimal trajectory.

\begin{figure}
    \centering
    \includegraphics[width=\textwidth]{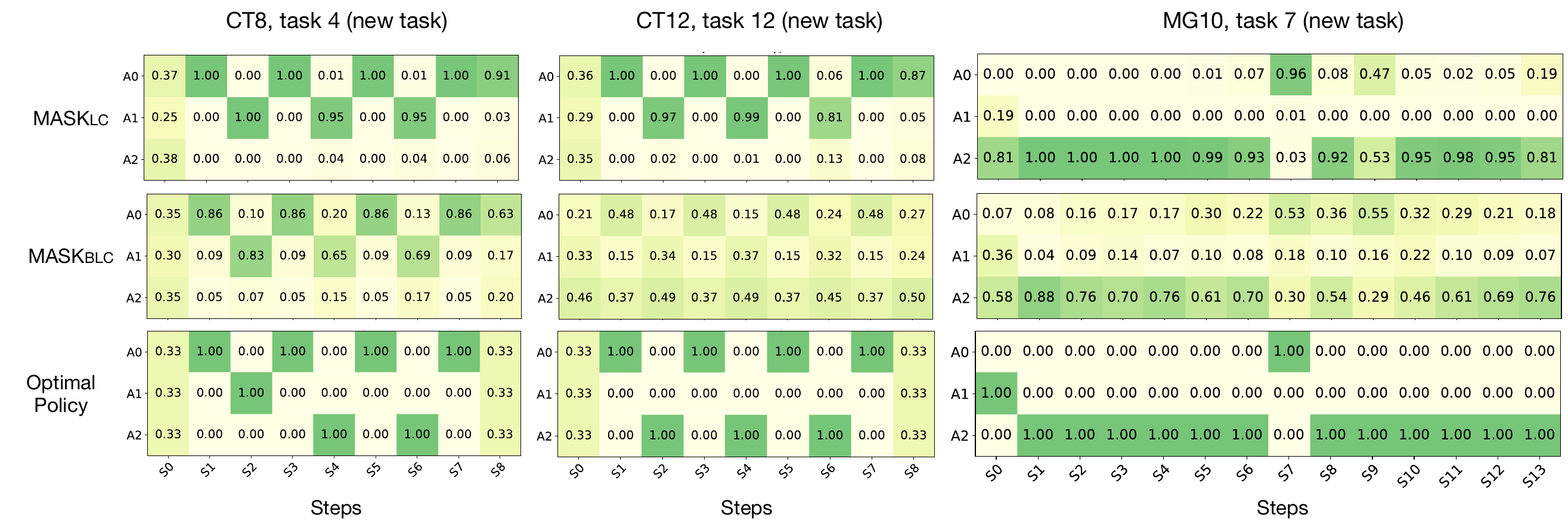}
    \caption{Knowledge reuse when learning a new tasks. The softmax output probabilities are shown during an episode with an unforeseen task (task 4 in \emph{CT8}, task 12 in \emph{CT12}, and task 7 in \emph{MG10}) as the agent is guided through an optimal policy. From the top row down, $\mathrm{Mask_{LC}}$ displays biased probabilities representing an average behavior across previous tasks. $\mathrm{Mask_{BLC}}$ shows biased probabilities but contains more randomness in comparison to $\mathrm{Mask_{LC}}$. A visual comparison with the optimal policy (bottom row) suggests that both $\mathrm{Mask_{LC}}$ and $\mathrm{Mask_{BLC}}$ start learning unforeseen tasks with useful knowledge. Note, $\mathrm{EWC_{MH}}$ produced balanced probabilities (uniform distribution) for all actions at each time step, and is thus omitted from the figure.}
    \label{fig:target_explore}
\end{figure}

Figure \ref{fig:target_explore} shows the analysis for the following cases: facing a 4th task after learning 3 tasks in the \emph{CT8} benchmark; facing the 12th task after learning 11 tasks in the \emph{CT12} benchmark; facing the 7th task after learning 6 tasks in the \emph{MG10} benchmark. The chosen task for each curriculum is set to test different instances of knowledge reuse. For \emph{CT8}, the 4th task tests the agent's ability to reuse knowledge after learning a few similar tasks, while the 12th task in \emph{CT12} investigates the agent's behavior after learning many tasks. In \emph{MG10}, the input distribution of the 7th task differs from the previous ones (i.e., from tasks with no lava to tasks with lava), thus testing the agent's ability to reuse previously learned navigation skills while dealing with new input scenarios. The result of the analysis is generalizable to other task changes in the curricula.

In the analysis, $\mathrm{EWC_{MH}}$ produced a purely random policy, with a uniform distribution over actions for each time step. Despite learning previous tasks, the new random head results in equal probabilities for all actions. On the contrary, both $\mathrm{Mask_{LC}}$ and $\mathrm{Mask_{BLC}}$ use previous masks to express preferences. In particular, in the \emph{CT8} and \emph{CT12}, the policy at steps 1, 3, 5 and 7 coincides with the optimal policy for the new task, likely providing an advantage. However, at steps 4 and 6 for the \emph{CT8}, and 2, 4, and 6 for the \emph{CT12}, $\mathrm{Mask_{LC}}$ has a markedly wrong policy: this is due to the fact that the new task has a different reward function and is therefore interfering. Due to the balanced combination of previous knowledge with a new mask, $\mathrm{Mask_{BLC}}$ seems to strike the right balance between trying known policies and exploring new ones. Such a balanced approach is also visible in the \emph{MG10} task. Here, task 7 (see the Appendix \ref{apndx:env-minigrid}) consists of avoiding the walls and the lava while proceeding ahead, then turning left and reaching the goal: most such skills are similar to those acquired in previously seen tasks, and therefore task 7 is learned starting from a policy that is close to being optimal.

If $\mathrm{Mask_{LC}}$ and $\mathrm{Mask_{BLC}}$ are capable of exploiting previous knowledge, it is natural to ask whether such knowledge can be exploited to learn increasingly more difficult tasks. The CT-graph \citep{soltoggio2019ctgraph} environment allows for increasing the complexity by increasing the depth of the graph. In particular, with a depth of 5, the benchmark results in a highly sparse reward environment with a probability of getting a reward with a random policy of only one in $3^{11} = 177,147$ episodes \citep{soltoggio2023ctgraph}. We therefore designed a curriculum, the \emph{CT8 multi depth}, composed of a set of related but increasingly complex tasks with depth 2, 3, 4 and 5 (two tasks each depth with a different reward function). 

\begin{figure}[ht]
    \centering
    \begin{tabular}{cc}
    \includegraphics[width=0.49\textwidth]
    {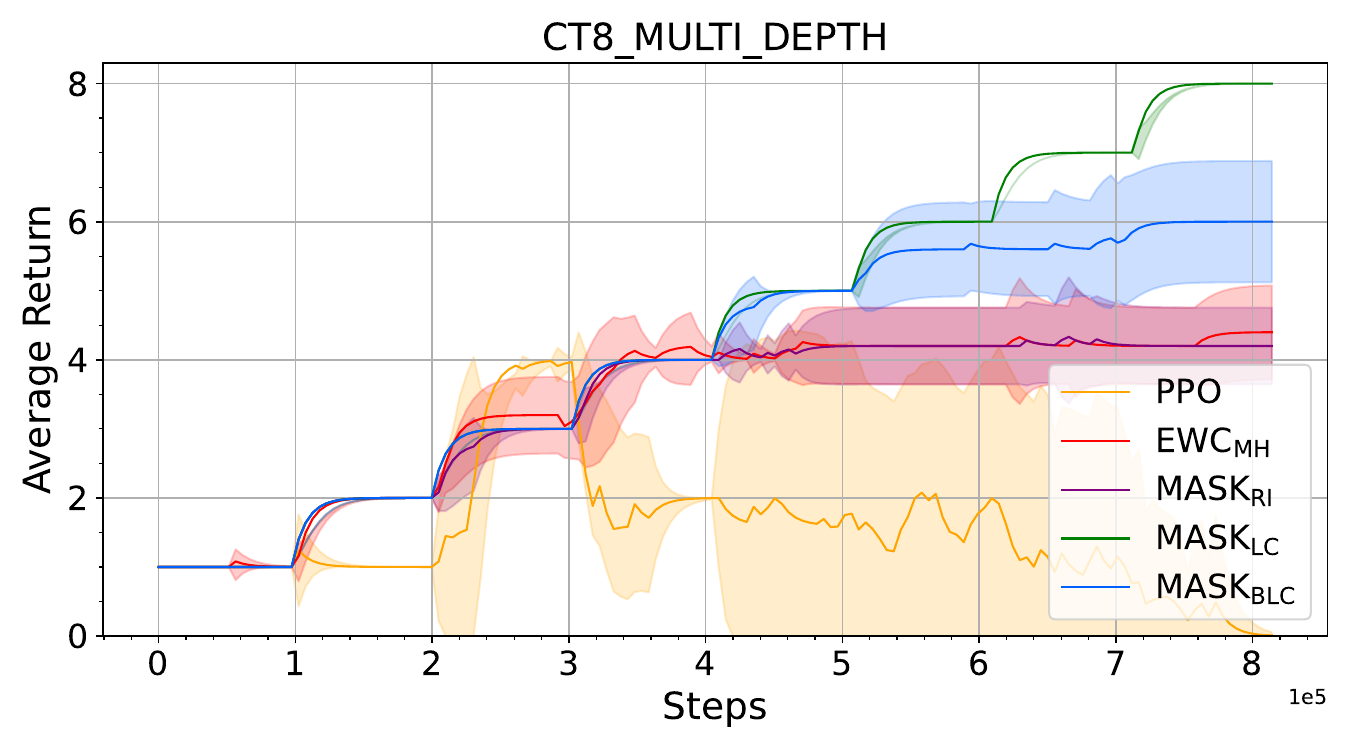}&
    \includegraphics[width=0.49\textwidth]{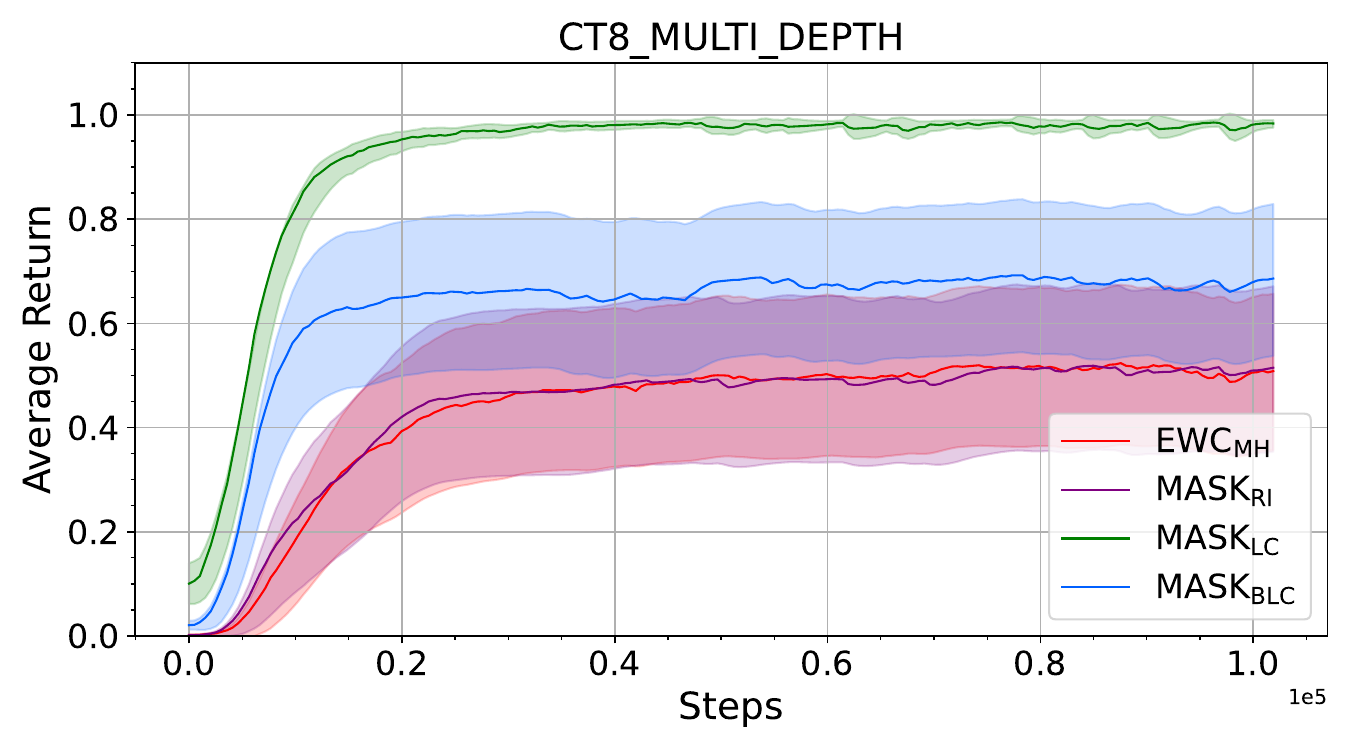}\\
     \end{tabular}
     \caption{
      Performance in the \emph{CT8 multi depth} curriculum. (Left) Lifelong evaluation performance on all tasks (mean and 95\% confidence interval on 5 seeds/samples). (Right) Training performance on each task, measured as the average return across all tasks and seeds runs (mean and 95\% confidence interval on 8 tasks and 5 seeds, i.e., 40 samples). Excluding $\mathrm{MASK_{LC}}$, the training performance for other methods are sub-optimal, due to the failure to solve later tasks in the curriculum.}
     \label{res:ct8_multi_depth_plot_eval_train}
\end{figure}

\begin{figure}
    \centering
     \includegraphics[width=0.5\textwidth]{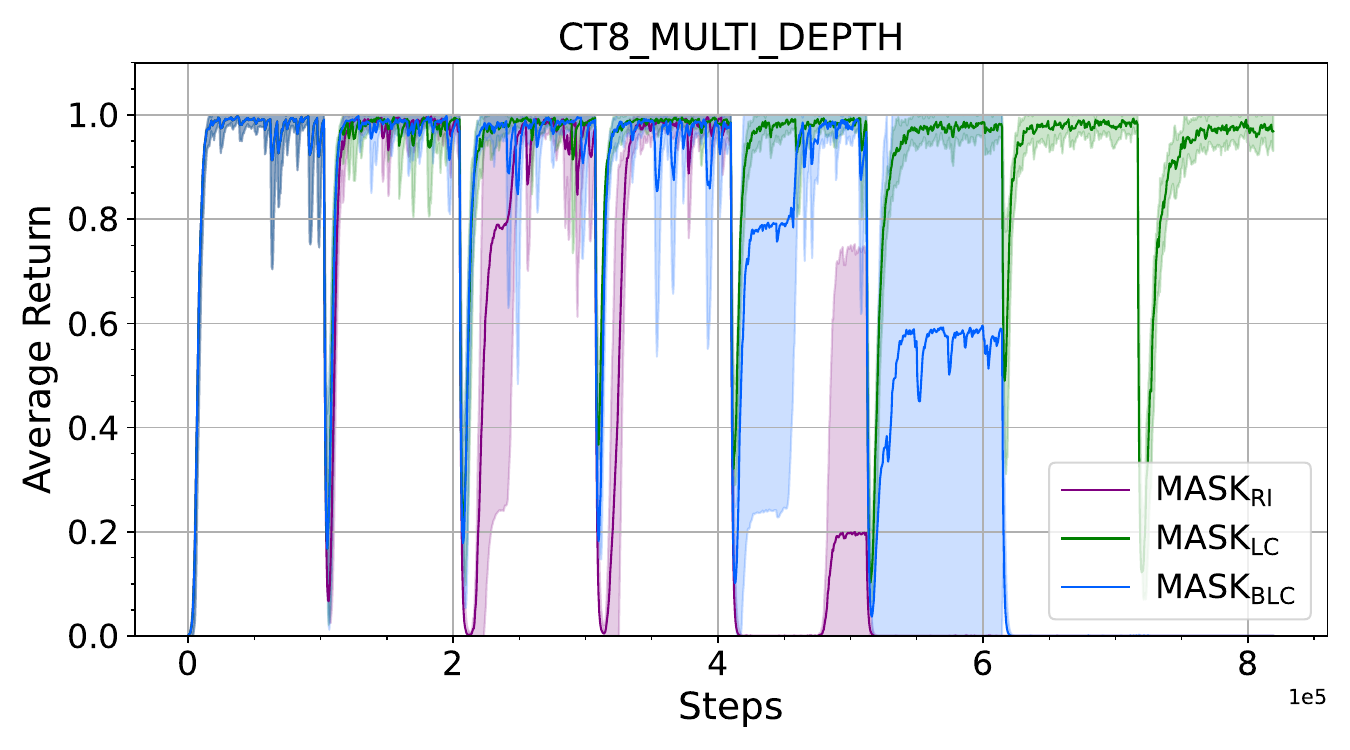}
    \caption{Training performance on each task, measured as the average return across all seeds runs (mean and 95\% confidence interval on 5 seeds/samples), expanded from Figure \ref{res:ct8_multi_depth_plot_eval_train}(Right).}
    \label{res:ct8_multi_depth_plot_train_ind_tasks}
\end{figure}

Figure \ref{res:ct8_multi_depth_plot_eval_train} shows the performance in the \emph{CT8 multi depth} curriculum. $\mathrm{EWC_{MH}}$ was able to learn the first 4 tasks with depth 2 and 3, but failed to learn the last 4 more complex tasks. Interestingly, $\mathrm{Mask_{BLC}}$ managed to learn task 5 and 6 only partially. $\mathrm{Mask_{LC}}$ was able to learn all tasks, demonstrating that it could reuse previous knowledge to solve increasingly difficult tasks. Figure \ref{res:ct8_multi_depth_plot_train_ind_tasks} presents the training performance for each individual task, highlighting were most methods fail in the curriculum.
\begin{table}
    \centering
    \begin{tabular}{lcc}\hline
      Method & Total Eval & Fwd Trnsf. \\ \hline
      PPO & 237.80 $\pm$ 115.65 & 0.05 $\pm$ 0.11 \\
      $\mathrm{EWC_{MH}}$ & 540.00 $\pm$ 39.88 & 0.10 $\pm$ 0.08 \\
      $\mathrm{Mask_{RI}}$ & 533.00 $\pm$ 41.13 & -- \\
      $\mathrm{Mask_{LC}}$ & 719.60 $\pm$ 0.68 & 0.79 $\pm$ 0.06 \\
      $\mathrm{Mask_{BLC}}$ & 644.00 $\pm$ 41.47 & 0.48 $\pm$ 0.11 \\
      \hline
    \end{tabular}
    \caption{Total evaluation return (AUC of the lifelong evaluation plot in Figure \ref{res:ct8_multi_depth_plot_eval_train}(Left)) and forward transfer during lifelong training in the \emph{CT8 multi depth}. Mean $\pm$ 95\% confidence interval reported.}
    \label{res:ct8_multi_depth_tbl_metrics}
\end{table}

\section{Discussion}
\label{sec:discussion}
The proposed approach employs learned modulatory masking methods to deliver lifelong learning performance in a range of reinforcement learning environments. The evaluation plots (the left panels of Figures \ref{res:ct8_plot_eval_train}, \ref{res:ct12_plot_eval_train}, \ref{res:mg10_plot_eval_train} and \ref{res:cw10_plot_eval_train}) show how the performance increases as the agent learns through a given curriculum. The monotonic increase, indicating minimal forgetting, is most clear for the CT-graph and Minigrid, while the Continual World appears to have a more noisy performance. The $\mathrm{EWC_{MH}}$ baseline algorithm shows to be a highly performing baseline in the \emph{CT8} and \emph{CT12} curricula, it performs less well in the MG10 curriculum, and poorly in the \emph{CW10} curriculum. The masking methods, instead, perform consistently in all benchmarks, with the linear combination methods, $\mathrm{Mask_{LC}}$ and $\mathrm{Mask_{BLC}}$, showing some advantage over the random initialization $\mathrm{Mask_{RI}}$. Evaluations on the ProcGen environments, while noisy to interpret from Figure \ref{fig:procgen}, reveal that the masking methods outperform IMPALA, Online EWC, P\&C and CLEAR by significant margins (Table \ref{tab:procgen_tbl_eval_metrics}).

While the learning dynamics of the core algorithm $\mathrm{Mask_{RI}}$ indicate superior performance to the baselines, we focused in particular on two extensions of the algorithm, $\mathrm{Mask_{LC}}$ and $\mathrm{Mask_{BLC}}$. These   use a linear combination of previously learned masks to search for the optimal policy in a new unforeseen task. These two variations combine previously learned masks with a new random mask to search for the optimal policy. One catch with this approach is that the performance on a new task will depend on which and how many tasks were previously learned. However, this is a property of all lifelong learning algorithms that leverage on previous knowledge. The balanced approach $\mathrm{Mask_{BLC}}$ starts with a 0.5 weight on the new random mask and can be seen as a blend between the random initialization $\mathrm{Mask_{RI}}$ and the linear combination $\mathrm{Mask_{LC}}$. On average, it appears to achieve slightly better performance than either $\mathrm{Mask_{RI}}$ or $\mathrm{Mask_{LC}}$. The standard linear combination $\mathrm{Mask_{LC}}$ was the only algorithm able to learn the most difficult task on the CT-graph. This suggests that the algorithm is capable of exploiting previous knowledge to solve challenging RL problems. The fact that both $\mathrm{Mask_{LC}}$ and $\mathrm{Mask_{BLC}}$ have superior performance to the core random initialization $\mathrm{Mask_{RI}}$ validates the hypothesis that previous knowledge stored in masks can be reused.

The analysis of the coefficients of the linear combination (Section \ref{subsec:analyis-linear-comb-coeffs}) reveals that the optimization can tune them to adapt to the nature of the curriculum. In the CT-graph and Minigrid curricula, previous masks are used in balanced proportions. On the Continual World environment, instead, particular masks, and layers within those masks, had significantly larger weights than others. From this observation, we conclude that the new proposed approach may be flexible enough to adapt to a variety of different curricula with different degrees of task similarity.

One concern with modulating masks is that memory requirements increase linearly with the number of tasks. This makes the approach not particularly scalable to large numbers of tasks. However, the promising performance of the linear combination approaches suggests that an upper limit could be imposed on the number of masks to be stored. After such a limit has been reached, new tasks can be learned solely as linear combinations of known masks, significantly reducing memory requirements. While this paper tested vector parameters with a scalar for each network layer, the analysis of the tuned parameters suggests that in some cases a single scalar for each mask could be used, further reducing memory requirements. Other suggestions to combat memory requirements in the mask approach are discussed in Appendix \ref{appendx:learned-mask-and-memory}.

In the modulating masking setup, it is assumed that a task oracle informs the lifelong RL agent of task boundaries and the task presented at any given time (during training and evaluation/testing). This is made possible by providing a task identifier to the agent  to select the correct mask. While the explicit specification of task boundaries is a limitation of all LRL methods that make this assumption, the nature of the proposed mask method, where a mask is associated to a task, implies that existing task detection methods could be combined with the mask setup to address this limitation. One approach could involve the use of \emph{forget-me-not} Bayesian approach to task detection \citep{milan2016forgetmenot}, as was employed in \citet{kirkpatrick2017ewc}. Another approach that could be explored is the use of optimal transport methods \citep{alvarez2020otdd, liu2022wasserstein} to measure distance of states/input across tasks. Also, the few-shots optimization of the linear superposition of masks via gradient descent that was employed in \citet{wortsman2020supermasks} for LSL could be employed to infer task mask in the LRL setup.

Given the fixed nature of the backbone network and the use of binary masks to sparsify the network, the representational capacity of the network could be affected. Masking approaches have been extensively studied in fixed neural networks \citep{zhou2019deconstructing, ramanujan2020s}, including lifelong supervised learning setup \citep{mallya2018piggyback, wortsman2020supermasks}, with discussions about representational capacity and generalization. While representation capacity could be affected, there are gains in generalization \citet{zhou2018non, frankle2018lottery} and robustness to noise \citep{arora2018stronger} in sparse networks. In the future investigations, the gains in generalization could be useful to model-based RL approaches in lifelong learning.

\section{Conclusion}
\label{sec:conclusion}
This work introduces the use of modulating masks for lifelong reinforcement learning problems. Variations of the algorithm are devised to ignore or exploit previous knowledge. All versions demonstrate the ability to learn on sequential curricula and show no catastrophic forgetting thanks to separate mask training. The analysis revealed that using previous masks to learn new tasks is beneficial as the linear combination of masks introduces knowledge in new policies. This finding promises potential new developments for compositional knowledge RL algorithms. Exploiting previous knowledge, one version of the algorithm was able to solve extremely difficult problems with reward probabilities as low as $5.6 \cdot 10^{-6}$ per episode simply using random exploration. These results suggest that modulating masks are a promising tool to expand the capabilities of lifelong reinforcement learning, in particular with the ability to exploit and compose previous knowledge to learn new tasks. 

\section*{Broader Impact Statement}
\label{sec:broader-impact-statement}
The advances introduced in this work contribute to the development of intelligent systems that can learn multiple tasks over a lifetime. Such a system has the potential for real-world deployment, especially in robotics and automation of manufacturing processes. As real-world automation increases, it may lead to reduce the demand for human labor in some industries, thereby impacting the economic security of people. Also, when such systems are deployed, careful considerations are necessary to ensure a smooth human machine collaboration in the workforce, ethical considerations and mitigation of human injuries.

\section*{Acknowledgments}
This material is based upon work supported by the United States Air Force Research Laboratory (AFRL) and Defense Advanced Research Projects Agency (DARPA) under Contract No.\, FA8750-18-C-0103 (Lifelong Learning Machines) and Contract No.\, HR00112190132 (Shared Experience Lifelong Learning). Any opinions, findings and conclusions or recommendations expressed in this material are those of the author(s) and do not necessarily reflect the views of the United States Air Force Research Laboratory (AFRL) and Defense Advanced Research Projects Agency (DARPA).

\bibliography{main}
\bibliographystyle{tmlr}

\newpage
\Large{\textbf{Appendix}}

\Large{\textbf{Lifelong Reinforcement Learning with Modulating Masks}}

\normalsize

\appendix

\section{Significance Testing}
\label{apndx:significance-testing}
Results obtained from experiments in RL usually produce high variance across seeds \citep{henderson2018deep}. This issue further leads to challenge of reproducible results. To address this concern, a difference test \citep{colas2018many}, using the Welch t-test and bootstrap confidence interval (BCI) were performed on the main results, evaluation performance and forward transfer. The tests were carried out at a significance level of 0.05. The BCI tests were run with 10,000 bootstrap iterations.

The outcome of the evaluation performance tests are reported in Tables \ref{tab:sig-test-tbl-eval-perf-ctgraph}, \ref{tab:sig-test-tbl-eval-perf-minigrid}, \ref{tab:sig-test-tbl-eval-perf-continual-world}, and \ref{tab:sig-test-tbl-eval-perf-procgen}, while the forward transfer tests are reported in Tables \ref{tab:sig-test-tbl-fwd-trnsf-ctgraph}, \ref{tab:sig-test-tbl-fwd-trnsf-minigrid}, and \ref{tab:sig-test-tbl-fwd-trnsf-continual-world}. The $\mathrm{MASK_{LC}}$ was chosen as the method to compare against for the difference testing. The table cells colored green signify that the test reported enough evidence to establish an order relationship between compared methods, and vice versa for cells colored red. Also, when a positive only interval is reported in the BCI test, it signifies that $\mathrm{MASK_{LC}}$ has a higher value than the method compared, and vice versa for negative only interval. The number of samples for the evaluation performance test is 3, the number of seed runs per method, while the number of samples for the forward transfer test is 3 multiplied by the number of tasks in each curriculum.

\begin{table}[h]
    \centering
    \begin{tabular}{l|cc|cc|cc}
        \toprule
        {} & \multicolumn{2}{c|}{CT8} & \multicolumn{2}{c|}{CT12} & \multicolumn{2}{c}{CT8 MD} \\
        Method & p-value & BCI & p-value & BCI & p-value & BCI \\
        \midrule
        PPO & \cellcolor{green!10}3.48e-10 & \cellcolor{green!10}[544.80, 563.00] & \cellcolor{green!10}4.87e-09 & \cellcolor{green!10}[1152.20, 1180.80] & \cellcolor{green!10}3.19e-04 & \cellcolor{green!10}[414.20, 559.20] \\
        
        $\mathrm{EWC_{MH}}$ & \cellcolor{red!10}1.52e-01 & \cellcolor{red!10}[-6.20, 67.20] & \cellcolor{red!10}5.73e-01 & \cellcolor{red!10}[-62.00, 47.00] & \cellcolor{green!10}2.35e-04 & \cellcolor{green!10}[157.80, 207.60] \\
        
        $\mathrm{MASK_{RI}}$ & \cellcolor{red!10}8.00e-01 & \cellcolor{red!10}[-4.60, 6.00] & \cellcolor{red!10}5.74e-02 & \cellcolor{green!10}[2.80, 16.20] & \cellcolor{green!10}2.28e-04 & \cellcolor{green!10}[169.60, 216.60] \\
        
        $\mathrm{MASK_{BLC}}$ & \cellcolor{green!10}1.16e-03 & \cellcolor{green!10}[-18.60, -10.80] & \cellcolor{green!10}1.30e-04 & \cellcolor{green!10}[-17.00, -10.60] & \cellcolor{green!10}7.17e-03 & \cellcolor{green!10}[50.40, 100.60] \\
        \bottomrule
    \end{tabular}
    \caption{Total evaluation performance significance testing for the CT-graph curricula. Welch t-test (p-value) and bootstrap confidence interval significance difference testing at 5\%, for $\mu_1 - \mu_2$, where $\mu_1$ is the average total evaluation performance achieved by $\mathrm{MASK_{LC}}$ and $\mu_2$ that of the comparisons (rows) in the table. Green colored cells are statistically significant result where there is enough evidence to establish an order (difference) between $\mu_1$ and $\mu_2$, and vice versa cells are colored red.}
    \label{tab:sig-test-tbl-eval-perf-ctgraph}
\end{table}

\begin{minipage}[h]{\textwidth}
  \centering
  \begin{minipage}{0.45\textwidth}
    \centering
    \begin{tabular}{l|cc}
        \toprule
        {} & \multicolumn{2}{c}{MG10} \\
        Method & p-value & BCI \\
        \midrule
        PPO & \cellcolor{green!10}4.16e-07 & \cellcolor{green!10}[1010.67, 1109.96] \\
        $\mathrm{EWC_{MH}}$ & \cellcolor{green!10}4.21e-04 & \cellcolor{green!10}[299.03, 499.05] \\
        $\mathrm{MASK_{RI}}$ & \cellcolor{red!10}5.32e-01 & \cellcolor{red!10}[-79.28, 35.90] \\
        $\mathrm{MASK_{BLC}}$ & \cellcolor{red!10}2.18e-01 & \cellcolor{red!10}[-109.42, 14.67] \\
        \bottomrule
    \end{tabular}
    \captionsetup{type=table}
    \captionof{table}{Total evaluation performance significance testing for the Minigrid curriculum. Welch t-test (p-value) and bootstrap confidence interval (BCI) significance difference testing at 5\%, for $\mu_1 - \mu_2$, where $\mu_1$ is the average total evaluation performance achieved by $\mathrm{MASK_{LC}}$ and $\mu_2$ that of the comparisons (rows) in the table. Green colored cells are statistically significant result where there is enough evidence to establish an order (difference) between $\mu_1$ and $\mu_2$, and vice versa cells are colored red.}
    \label{tab:sig-test-tbl-eval-perf-minigrid}
  \end{minipage}
  \hspace{0.5em}
  \begin{minipage}{0.5\textwidth}
    \centering
    \begin{tabular}{l|cc}
        \toprule
        {} & \multicolumn{2}{c}{CW10} \\
        Method & p-value & BCI \\
        \midrule
        PPO & \cellcolor{green!10}5.98e-03 & \cellcolor{green!10}[161.63, 268.27] \\
        $\mathrm{EWC_{MH}}$ & \cellcolor{green!10}1.12e-02 & \cellcolor{green!10}[206.87, 301.57] \\
        $\mathrm{MASK_{RI\_D}}$ & \cellcolor{green!10}1.18e-02 & \cellcolor{green!10}[195.02, 290.22] \\
        $\mathrm{MASK_{RI\_C}}$ & \cellcolor{red!10}6.14e-01 & \cellcolor{red!10}[-46.63, 96.80] \\
        $\mathrm{MASK_{BLC}}$ & \cellcolor{red!10}5.09e-01 & \cellcolor{red!10}[-41.97, 106.07] \\
        \bottomrule
    \end{tabular}
    \captionsetup{type=table}
    \captionof{table}{Total evaluation performance significance testing for the Continual World curriculum. Welch t-test (p-value) and bootstrap confidence interval (BCI) significance difference testing at 5\%, for $\mu_1 - \mu_2$, where $\mu_1$ is the average total evaluation performance achieved by $\mathrm{MASK_{LC}}$ and $\mu_2$ that of the comparisons (rows) in the table. Green colored cells are statistically significant result where there is enough evidence to establish an order (difference) between $\mu_1$ and $\mu_2$, and vice versa cells are colored red.}
    \label{tab:sig-test-tbl-eval-perf-continual-world}
  \end{minipage}
\end{minipage}

\begin{table}[h]
    \centering
    \begin{tabular}{l|cc|cc}
        \toprule
         & \multicolumn{2}{c|}{Welch test p-value for $\mu_1 - \mu_2$} & \multicolumn{2}{c}{Confidence Interval for $\mu_1 - \mu_2$} \\
        Method & Train tasks & Test tasks & Train tasks & Test tasks \\
        \midrule
        IMPALA & \cellcolor{green!10}4.09e-03 & \cellcolor{green!10}3.26e-05 & \cellcolor{green!10}[7421.60, 9977.69] & \cellcolor{green!10}[8912.31, 10226.61] \\
        Online EWC & \cellcolor{green!10}2.63e-03 & \cellcolor{green!10}4.94e-05 & \cellcolor{green!10}[6276.49, 9157.50] & \cellcolor{green!10}[7858.06, 9233.60] \\
        P\&C & \cellcolor{green!10}1.28e-03 & \cellcolor{green!10}1.01e-04 & \cellcolor{green!10}[7426.82, 10827.51] & \cellcolor{green!10}[8170.99, 9802.61] \\
        CLEAR & \cellcolor{red!10}6.25e-01 & \cellcolor{green!10}5.77e-03 & \cellcolor{red!10}[-1468.47, 3087.61] & \cellcolor{green!10}[2982.40, 4946.90] \\
        $\mathrm{MASK_{RI}}$ & \cellcolor{red!10}9.78e-01 & \cellcolor{red!10}9.23e-01 & \cellcolor{red!10}[-1474.67, 1156.01] & \cellcolor{red!10}[-1844.39, 2830.19] \\
        $\mathrm{MASK_{BLC}}$ & \cellcolor{red!10}4.99e-01 & \cellcolor{red!10}6.67e-01 & \cellcolor{red!10}[-871.74, 2481.94] & \cellcolor{red!10}[-846.61, 2214.05] \\
        \bottomrule
    \end{tabular}
    \caption{ProcGen Total evaluation performance: Welch t-test and bootstrap confidence interval significance difference testing at 5\%, for $\mu_1 - \mu_2$, where $\mu_1$ is the average total evaluation performance achieved by $\mathrm{MASK_{LC}}$ and $\mu_2$ that of the comparisons (rows) in the table. Green colored cells are statistically significant result where there is enough evidence to establish an order (difference) between $\mu_1$ and $\mu_2$, and vice versa cells are colored red.}
    \label{tab:sig-test-tbl-eval-perf-procgen}
\end{table}

\begin{table}[h]
    \centering
    \begin{tabular}{l|cc|cc|cc}
        \toprule
        {} & \multicolumn{2}{c|}{CT8} & \multicolumn{2}{c|}{CT12} & \multicolumn{2}{c}{CT8 MD} \\
        Method & p-value & BCI & p-value & BCI & p-value & BCI \\
        \midrule
        PPO & \cellcolor{green!10}9.73e-03 & \cellcolor{green!10}[0.07, 0.54] & \cellcolor{green!10}5.76e-09 & \cellcolor{green!10}[0.45, 0.84] & \cellcolor{green!10}1.18e-16 & \cellcolor{green!10}[0.62, 0.87] \\
        
        $\mathrm{EWC_{MH}}$ & \cellcolor{green!10}2.50e-08 & \cellcolor{green!10}[0.48, 0.91] & \cellcolor{green!10}3.36e-11 & \cellcolor{green!10}[0.45, 0.77] & \cellcolor{green!10}1.03e-21 & \cellcolor{green!10}[0.59, 0.78] \\
        
        $\mathrm{MASK_{BLC}}$ & \cellcolor{green!10}4.23e-03 & \cellcolor{green!10}[-0.33, -0.06] & \cellcolor{green!10}3.02e-03 & \cellcolor{green!10}[-0.23, -0.05] & \cellcolor{green!10}1.10e-05 & \cellcolor{green!10}[0.18, 0.42] \\
        \bottomrule
    \end{tabular}
    \caption{Forward transfer significance testing for the CT-graph curricula. Welch t-test (p-value) and bootstrap confidence interval significance difference testing at 5\%, for $\mu_1 - \mu_2$, where $\mu_1$ is the average total evaluation performance achieved by $\mathrm{MASK_{LC}}$ and $\mu_2$ that of the comparisons (rows) in the table. Green colored cells are statistically significant result where there is enough evidence to establish an order (difference) between $\mu_1$ and $\mu_2$, and vice versa cells are colored red.}
    \label{tab:sig-test-tbl-fwd-trnsf-ctgraph}
\end{table}

\begin{minipage}[h]{\textwidth}
  \centering
  \begin{minipage}{0.45\textwidth}
    \centering
    \begin{tabular}{l|cc}
        \toprule
        {} & \multicolumn{2}{c}{MG10} \\
        Method & p-value & BCI \\
        \midrule
        PPO & \cellcolor{red!10}5.11e-01 & \cellcolor{red!10}[-0.29, 0.58] \\
        $\mathrm{EWC_{MH}}$ & \cellcolor{green!10}5.39e-04 & \cellcolor{green!10}[0.36, 1.21] \\
        $\mathrm{MASK_{BLC}}$ & \cellcolor{green!10}2.03e-03 & \cellcolor{green!10}[-0.84, -0.19] \\
        \bottomrule
    \end{tabular}
    \captionsetup{type=table}
    \captionof{table}{Forward transfer significance testing for the Minigrid curriculum. Welch t-test (p-value) and bootstrap confidence interval (BCI) significance difference testing at 5\%, for $\mu_1 - \mu_2$, where $\mu_1$ is the average total evaluation performance achieved by $\mathrm{MASK_{LC}}$ and $\mu_2$ that of the comparisons (rows) in the table. Green colored cells are statistically significant result where there is enough evidence to establish an order (difference) between $\mu_1$ and $\mu_2$, and vice versa cells are colored red.}
    \label{tab:sig-test-tbl-fwd-trnsf-minigrid}
  \end{minipage}
  \hspace{0.5em}
  \begin{minipage}{0.5\textwidth}
    \centering
    \begin{tabular}{l|cc}
        \toprule
        {} & \multicolumn{2}{c}{CW10} \\
        Method & p-value & BCI \\
        \midrule
        PPO & \cellcolor{green!10}6.37e-03 & \cellcolor{green!10}[0.94, 5.93] \\
        $\mathrm{EWC_{MH}}$ & \cellcolor{green!10}6.34e-04 & \cellcolor{green!10}[3.30, 10.29] \\
        $\mathrm{MASK_{BLC}}$ & \cellcolor{red!10}3.66e-01 & \cellcolor{red!10}[-0.34, 0.83] \\
        \bottomrule
    \end{tabular}
    \captionsetup{type=table}
    \captionof{table}{Forward transfer significance testing for the Continual World curriculum. Welch t-test (p-value) and bootstrap confidence interval (BCI) significance difference testing at 5\%, for $\mu_1 - \mu_2$, where $\mu_1$ is the average total evaluation performance achieved by $\mathrm{MASK_{LC}}$ and $\mu_2$ that of the comparisons (rows) in the table. Green colored cells are statistically significant result where there is enough evidence to establish an order (difference) between $\mu_1$ and $\mu_2$, and vice versa cells are colored red.}
    \label{tab:sig-test-tbl-fwd-trnsf-continual-world}
  \end{minipage}
\end{minipage}

\section{Hyper-parameters}
\label{apndx:hyperparameters}
In the experiments across the CT-graph, Minigrid and Continual World, all lifelong RL agents were built on top of the PPO algorithm. The hyper-parameters for the experiments are presented in Table \ref{tab:hyperparams_ct_mg_cw}. The $EWC_{MH}$ and $EWC_{SH}$ lifelong RL methods contain additional hyper-parameters which defines the weight preservation (consolidation) loss coefficient $\lambda$ and the weight of the moving average $\alpha$, for the online estimation of the fisher information matrix parameters following \citet{chaudhry2018riemannian}. For Continual World, $\alpha = 0.75$ and $\lambda = 1 \times 10^4$, while for the CT-graph and Minigrid experiments, $\alpha = 0.5$ and $\lambda = 1 \times 10^2$. The hyper-parameters for each method were set based on well-established values and preliminary tests. In each aforementioned benchmark, the hyper-parameters for the PPO algorithm were kept the same across all methods to enable fair comparison. 

For the ProcGen experiments, the setup reported in \citet{powers2022cora} was followed, with each lifelong RL agent built on top of the IMPALA algorithm. The hyper-parameters for the baselines (IMPALA, Progress \& Compress (P\&C), ONLINE EWC, and CLEAR) were kept the same as in \citet{powers2022cora} for the experiments are presented in Table \ref{tab:hyperparams_procgen}. ONLINE EWC contains additional hyper-parameter such as $\lambda = 175$ and $\mathrm{replay\_buffer\_size} = 1 \times 10^{6}$. For P\&C, $\lambda = 3000$,  $\mathrm{replay\_buffer\_size} = 1 \times 10^{5}$, and $\mathrm{num\_train\_steps\_of\_progress} = 3906$. For CLEAR, $\mathrm{replay\_buffer\_size} = 5 \times 10^{6}$.

\begin{table}[ht]
    \centering
    \begin{tabular}{p{11em}ccc}
        \toprule
        Hyper-parameter &  CT8 / CT12 / CT8 MD & MG10 & CW10 \\
        \midrule
        Learning rate & 0.00015 & 0.00015 & 0.0005 \\
        Optimizer & RMSprop & RMSprop & Adam \\
        Discount factor & 0.99 & 0.99 & 0.99 \\
        Gradient clip & 5 & 5 & 5 \\
        Entropy & 0.1 & 0.1 & 0.005 \\
        GAE & 0.99 & 0.99 & 0.97 \\
        Rollout length & 128 & 128 & 5120 \\
        Num. of workers & 4 & 4 & 1 \\
        PPO ratio clip & 0.1 & 0.1 & 0.2 \\
        PPO optim. epochs & 8 & 8 & 16 \\
        PPO optim. mini batch & 64 & 64 & 160 \\
        Train steps per task & 102,400 & 256,000 & 10.24M \\
        Train iterations per task: $\frac{\mathrm{train steps}}{\mathrm{rollout} \times \mathrm{workers}}$ & 200 & 500 & 2000 \\
        \midrule
        Eval. interval & 10 & 20  & 200 \\
        Eval. episodes & 10 & 10 & 10 \\
        \bottomrule
    \end{tabular}
    \caption{Hyper-parameters for curricula in the CT-graph (CT8, CT12, CT8 Multi Depth), Minigrid (MG10) and Continual World (CW10) environments.}
    \label{tab:hyperparams_ct_mg_cw}
\end{table}

\begin{table}[htbp]
    \centering
    \begin{tabular}{lc}
        \toprule
        Hyper-parameter &  Value\\
        \midrule
        Num. of workers & 64 \\
        Batch size & 32 \\
        Rollout length & 20 \\
        Entropy & 0.01 \\
        Learning rate & $4 \times 10^{-4}$ \\
        Optimizer & RMSprop \\
        Gradient clip & 40 \\
        Discount factor & 0.99 \\
        Num. of cycles (repeat curriculum) & 5 \\
        Num. of train step per task per cycle & 5M \\
        Num. of eval episodes & 10 \\
        Eval. interval & 0.25M train steps \\
        \bottomrule
    \end{tabular}
    \caption{Hyper-parameters for the curriculum in the ProcGen environment.}
    \label{tab:hyperparams_procgen}
\end{table}

\section{Network Specifications}
\label{apndx:network-specification}

The policy network specification for the CT-graph (i.e., \emph{CT8, CT12, and CT8 multi depth}) and Minigrid (i.e., \emph{MG10}) curricula is presented in Table \ref{tab:network-spec-ct-mg}, with ReLU activation function employed. The output of the actor layer produces logits of a categorical distribution.

\begin{table}[h]
    \centering
    \begin{tabular}{lcc}
        \toprule
        Layer  & Input units & Output units \\
        \midrule
        Linear 1 (shared)  & - & 200 \\
        Linear 2 (shared)  & 200 & 200 \\
        Linear 3 (shared)  & 200 & 200 \\
        \midrule
        Linear (actor output)  & 200 & 3 \\
        Linear (value output)  & 200 & 1 \\
        \bottomrule
    \end{tabular}
    \caption{Network specification of policy network across all methods for CT-graph and Minigrid curricula. Note, for multi head EWC network, the are multiple $\mathrm{Linear\ (actor\ output)}$ corresponding to the number of tasks.}
    \label{tab:network-spec-ct-mg}
\end{table}

For the Continual World (i.e, \emph{CW10}) curriculum, the policy network specification is presented in Table \ref{tab:network-spec-cw}, with $\mathrm{Tanh}$ activation function employed. The output of the actor layer produces the mean and standard deviation of a gaussian distribution.The output of the standard deviation actor output layer is clipped within the range [-0.6931, 0.4055]. 

\begin{table}[ht]
    \centering
    \begin{tabular}{lcc}
        \toprule
        Layer  & Input units & Output units \\
        \midrule
        Linear (actor body 1)  & - & 128 \\
        Linear (actor body 2)  & 128 & 128 \\
        Linear (actor output, mean)  & 128 & 3 \\
        Linear (actor output, log std)  & 128 & 3 \\
        \midrule
        Linear (value body 1)  & - & 128 \\
        Linear (value body 2)  & 128 & 128 \\
        Linear (value output)  & 128 & 1 \\
        \bottomrule
    \end{tabular}
    \caption{Network specification of policy network across all methods for Continual World curriculum. Note, for multi head EWC network, the are multiple $\mathrm{Linear\ (actor\ output)}$ corresponding to the number of tasks.}
    \label{tab:network-spec-cw}
\end{table}

For the ProcGen environment, the input observation is an RGB image with shape $3 \times 64 \times 64$ and 15 discrete actions. ReLU activation was employed in the network. The policy specification for the network across all methods is presented in Table \ref{tab:network-spec-procgen}

\begin{table}[!htbp]
    \centering
    \begin{tabular}{lccccc}
        \toprule
        Layer & Input channels/units & Output channels/units & Kernel & Stride & Pad  \\
        \midrule
        Conv 1 (shared) & 3 & 32 & [8, 8] & 4 & 0 \\
        Conv 2 (shared)& 32 & 64 & [4, 4] & 2 & 0 \\
        Conv 3 (shared) & 64 & 64 & [3, 3] & 1 & 0 \\
        \hline
        Flatten & $4 \times 4 \times 64 $ & 1024 & - & - & - \\
        Linear 1 (shared) & 1024 & 512 & - & - & - \\
        \hline
        Linear (actor output) & 528 & 15 & - & - & - \\
        Linear (value output) & 528 & 1 & - & - & - \\
        \bottomrule
    \end{tabular}
    \caption{Network specification for the ProcGen experiments. Note, the number of input units for the actor and value output heads changes to 528 because the one-hot action vector (i.e., size 15) and reward scalar (i.e., size 1) from the previous time step is concatenated to the output of Linear 1.}
    \label{tab:network-spec-procgen}
\end{table}

Note that across all multi-head EWC experiments, the policy network contains multiple actor output layer corresponding to the number of tasks.

\subsection{Backbone Network Initialization for Modulatory Masking Methods}
\label{subsec:masking-backbone-net-init}
Across all experiments, the weights of the backbone network for the modulatory masking methods were initialized using the signed Kaiming constant method, introduced in \citet{ramanujan2020s}. The constant $\pm c$ is the standard deviation of the Kaiming normal (distribution) initialization method, and could vary from layer to layer in the network. Furthermore, the bias parameters were disabled for the backbone networks in the masking methods, following the setup in \citet{wortsman2020supermasks}.

\section{Environments}
\label{apndx:environments}
\subsection{CT-graph}
\label{apndx:env-ctgraph}
The configurable tree graph (CT-graph) \citep{soltoggio2019ctgraph, soltoggio2023ctgraph} is a sparse reward, discrete action space environment with configurable parameters that define the search space. The environment is represented as a graph, where each node is a state represented as a $12 \times 12$ gray scale image. There exist a number of state/node types in the environment, which are start (H), wait (W), decision (D), end/leaf (E), and fail (F) state. Each environment instance contains one home state, one fail states, and a number of wait, decision and end states. The goal of an agent is to navigate from the home state to one of the end states designated as the goal --- the agent receives a reward of 1 when it enters the goal state, but 0 at every other time step. If the agent takes an incorrect action in any state, the agent transitions to the fail state, after which an environment reset takes it back to the home state.

The size and complexity (search space) of each environment (graph) instance is determined by a set of configuration parameters --- hence the term "configurable in the name". Two majors parameters in the CT-graph are the branch $b$ and depth $d$ that defines the branch (i.e., the width or number of decision actions at a decision state) and depth (i.e., the length) of the instantiated graph. The combination of the $b$ and $d$ determine how many end states exist in a each environment instance. Also, $b$ determines the action space of an instance --- defined as $b + 1$. The search space grow exponentially as $b$ and $d$ increase. Thus, the benchmark can be set up to the appropriate complexity to test the limits of RL algorithms.

A task is defined by setting one of the leaf states as a desired goal state that can be reached only via one trajectory. 

\begin{figure}[htbp]
    \centering
    \begin{tabular}{cc}
    \includegraphics[width=0.45\textwidth]{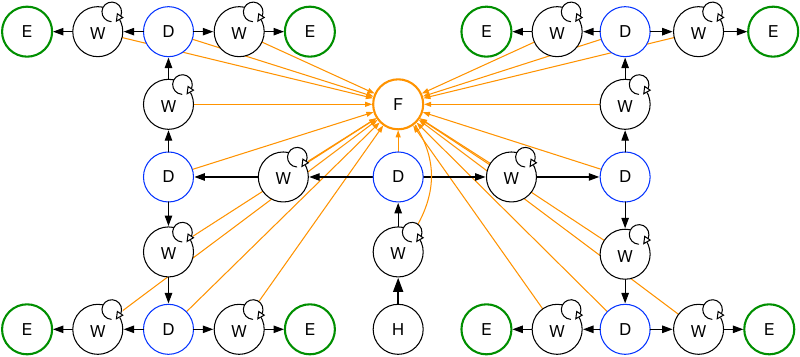}
         &     \includegraphics[width=0.21\textwidth]{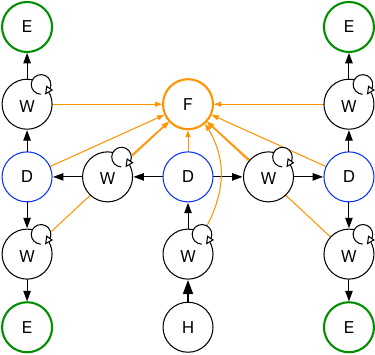}
 \\
    \end{tabular}
    \caption{CT-graph environments. States are: home (H), wait (W), decision (D), end (E), fail (F). Three actions at W and D nodes determine the next state. (Left) CT8: a depth-3 graph with three sequential decision states (D). Reward probability $1/3^7 = 1/2187$ reward/episodes. (Right) A depth-2 graph with 4 leaf states (CT4) that combined with CT8 results in the CT12 curriculum.}
    \label{fig:ct-graphs}
\end{figure}

For the \emph{CT8} curriculum, a graph instance with parameter $b=2$ and $d=3$ was employed --- $2^3$ end states. The 8 tasks comprise of each end state designated as the goal/reward location per task. For the \emph{CT12} curriculum, two graph instances with 4 ($b=2$ and $d=2$) and 8 ($b=2$ and $d=2$) different end/reward states were combined. Additionally, the $8$-task graph has a longer path to the reward that introduces variations in both the transition and reward functions. The \emph{CT12} curriculum was based on an interleave of the tasks from both graph instances (i.e., task 1 in 4-tasks, task 1 in 8-tasks, task 2 in 4-tasks, task 2 in 8-tasks, task 3 in 4-tasks, and so on). See Figure \ref{fig:ct-graphs} for a graphical representation of the 8-tasks and 4-tasks CT-graph. Lastly, the \emph{CT8 multi depth} curriculum was composed of the first two end/goal states in each of the following graph instance: (i) $b=2$ and $d=2$, (ii) $b=2$ and $d=3$, (iii) $b=2, d=4$, (iv) $b=2, d=5$.

With a branching factor (breadth) $b$ of 2 across all CT-graph curricula, the action space was defined as 3 (i.e., $b + 1$).

\subsection{Minigrid}
\label{apndx:env-minigrid}
\begin{figure}[htbp]
    \centering
    \begin{subfigure}{0.19\textwidth}
        \centering
        \includegraphics[width=\textwidth]{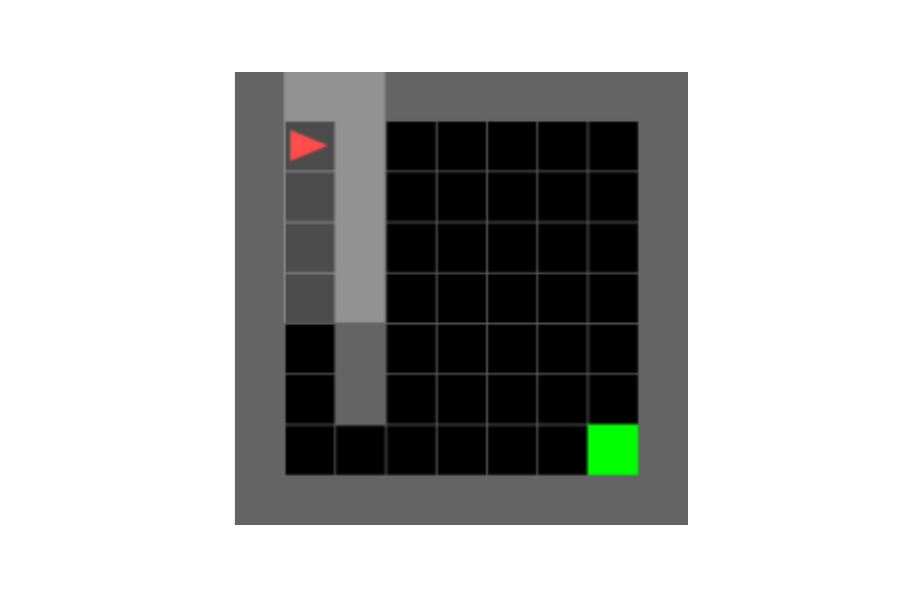}
    \end{subfigure}
    \begin{subfigure}{0.19\textwidth}
        \centering
        \includegraphics[width=\textwidth]{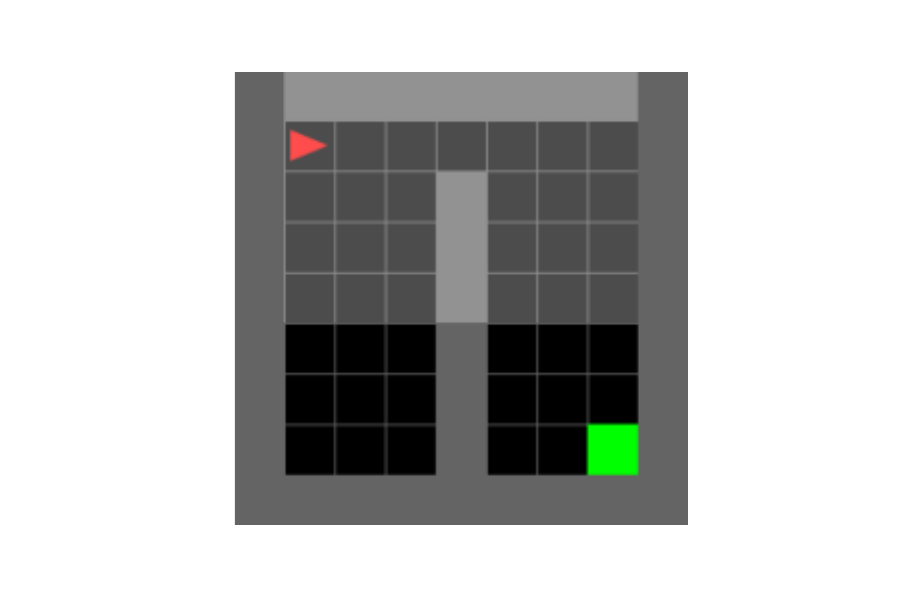}
    \end{subfigure}
    \begin{subfigure}{0.19\textwidth}
        \centering
        \includegraphics[width=\textwidth]{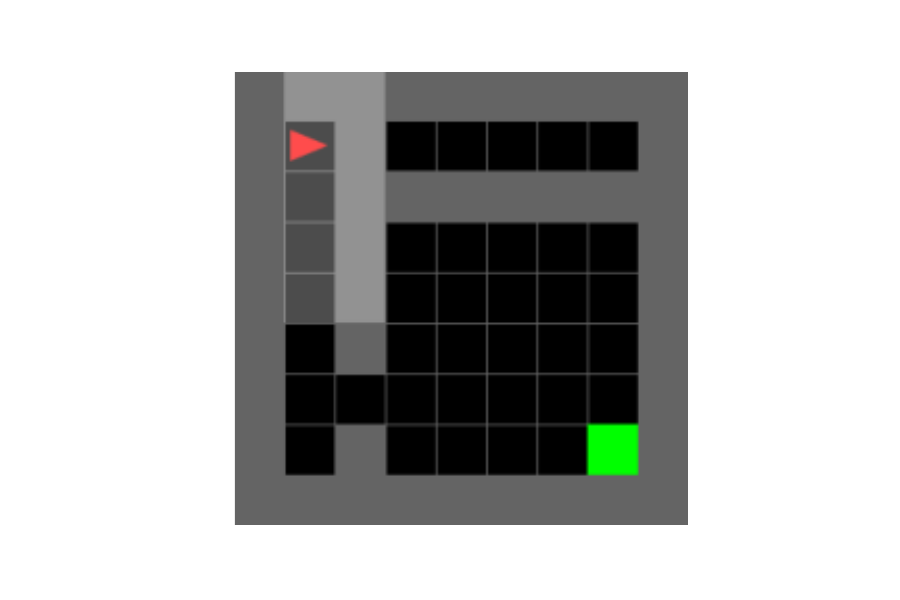}
    \end{subfigure}
    \begin{subfigure}{0.19\textwidth}
        \centering
        \includegraphics[width=\textwidth]{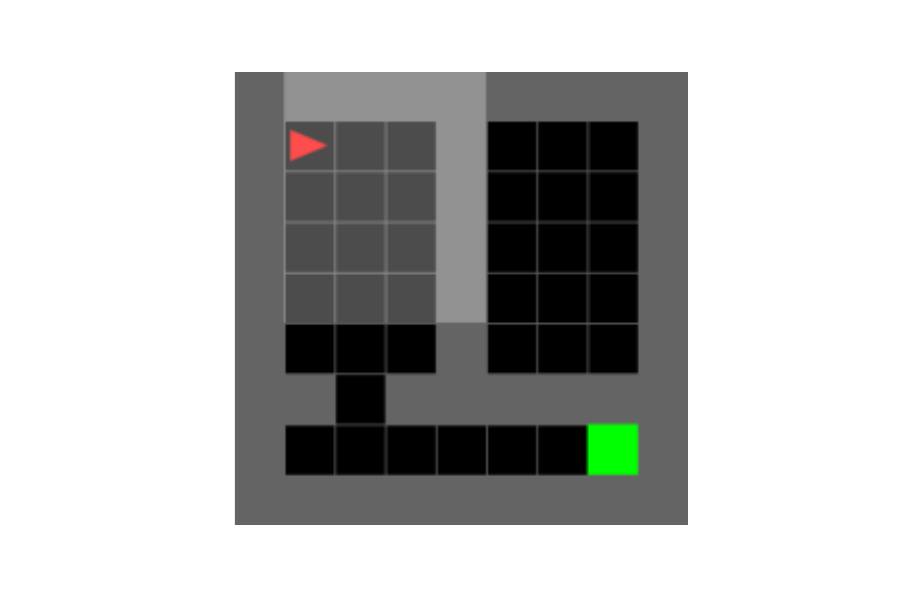}
    \end{subfigure}
    \begin{subfigure}{0.19\textwidth}
        \centering
        \includegraphics[width=\textwidth]{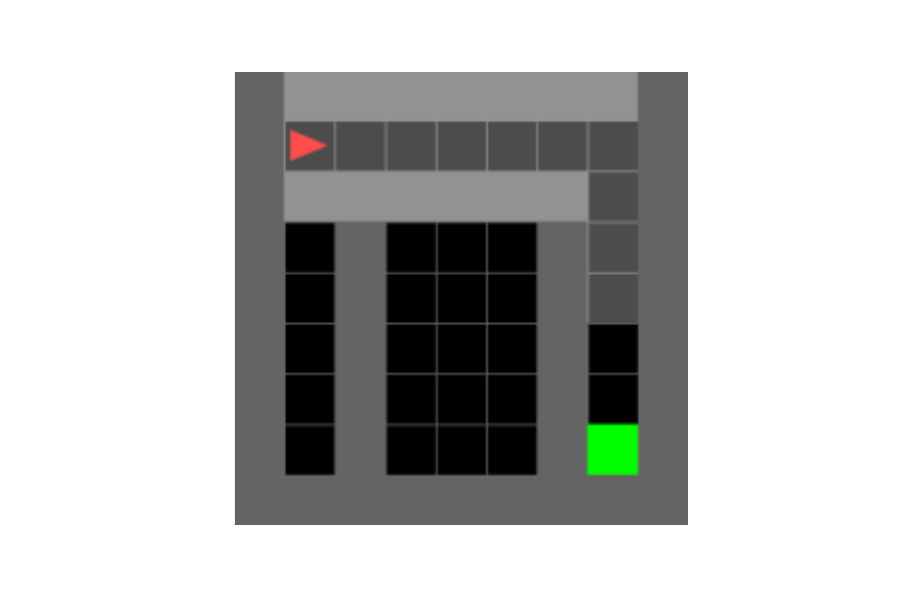}
    \end{subfigure}

    \begin{subfigure}{0.19\textwidth}
        \centering
        \includegraphics[width=\textwidth]{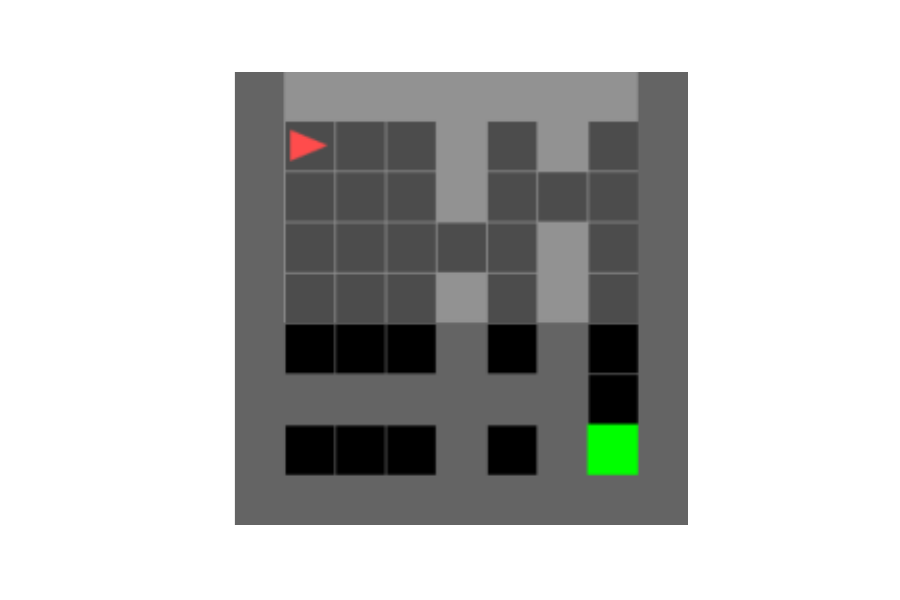}
    \end{subfigure}
    \begin{subfigure}{0.19\textwidth}
        \centering
        \includegraphics[width=\textwidth]{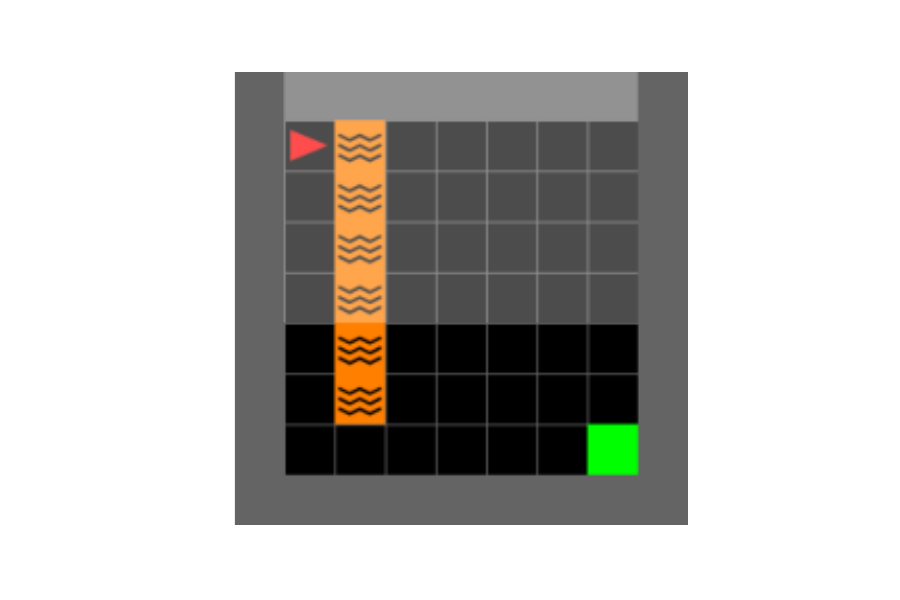}
    \end{subfigure}
    \begin{subfigure}{0.19\textwidth}
        \centering
        \includegraphics[width=\textwidth]{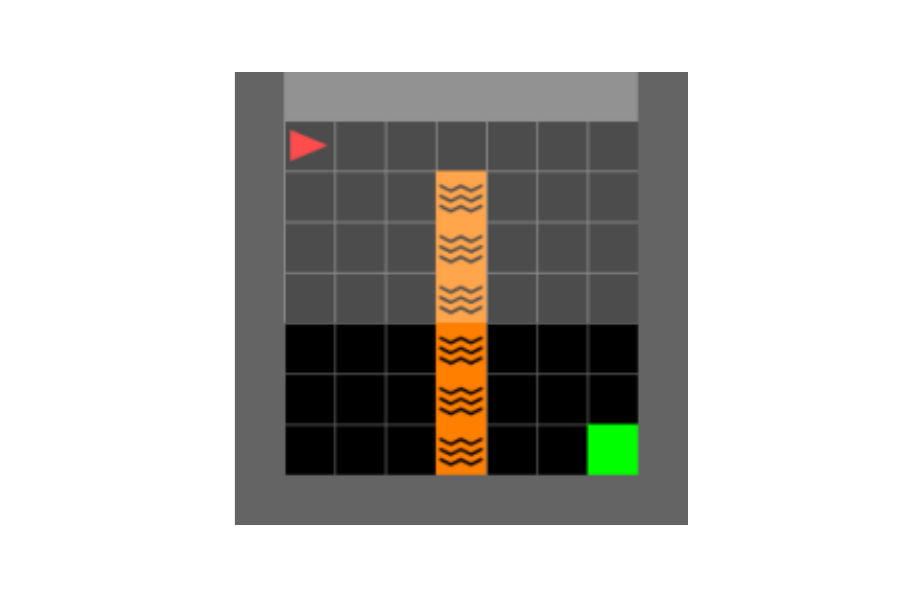}
    \end{subfigure}
    \begin{subfigure}{0.19\textwidth}
        \centering
        \includegraphics[width=\textwidth]{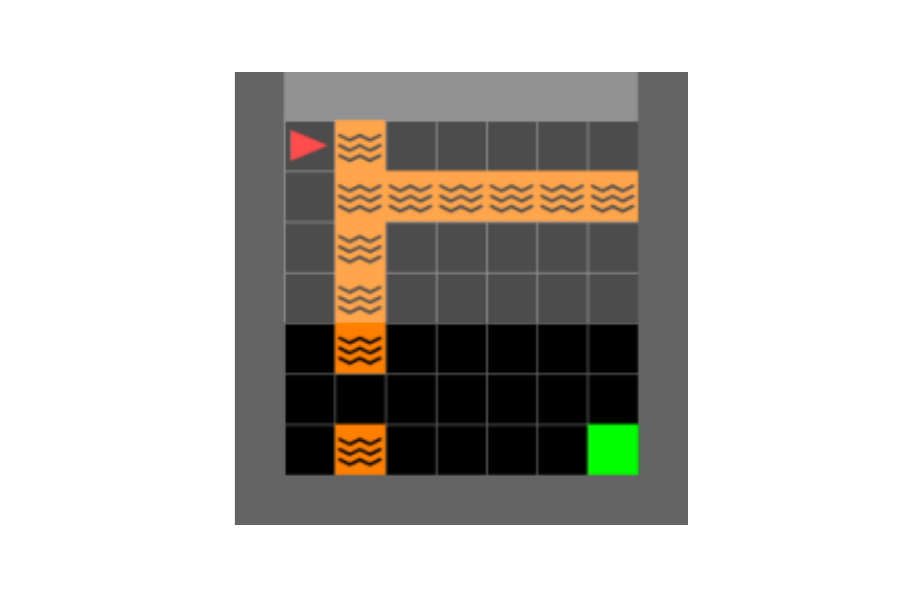}
    \end{subfigure}
    \begin{subfigure}{0.19\textwidth}
        \centering
        \includegraphics[width=\textwidth]{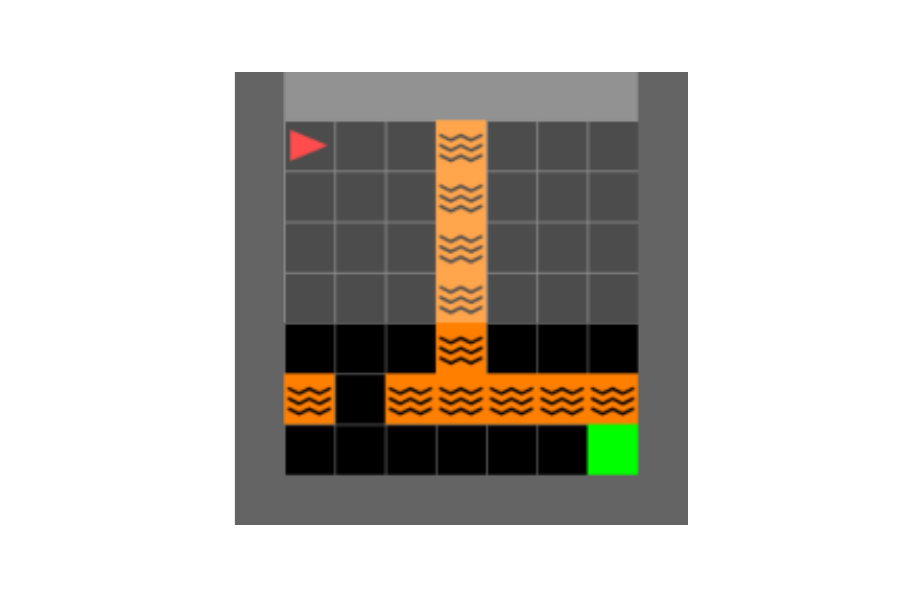}
    \end{subfigure}
 
    \caption{Visual representation of the 10 tasks in the \emph{MG10} curriculum. From left to right, two variants of each class: $\mathrm{SimpleCrossingS9N1}$, $\mathrm{SimpleCrossingS9N2}$, $\mathrm{SimpleCrossingS9N3}$, $\mathrm{LavaCrossingS9N1}$, $\mathrm{LavaCrossingS9N2}$.}
    \label{fig:env-mingrid}
\end{figure}

Similar to the CT-graph, the Minigrid \citep{gym_minigrid} is a sparse reward, discrete action navigation environment that consists of a number of predefined partially observable tasks with varying levels of complexity. The environment is setup as a grid world (with fast execution) where an agent is required to navigate to a goal location while avoiding obstacles such as walls, lava, moving balls, etc. It consist of a number of pre-defined grid worlds with several sub-variants defined by changing the random number generator seed. For all Minigrid experiments in this work, the default grid encoding was employed, with each state represented using a tensor of shape $7 \times 7 \times 3$. The agent only get a reward slightly under 1 (depending on the number of steps taken as defined in Equation \ref{eqn:minigrid-reward-fn}) when it arrives at the goal location, and a reward of 0 at every other state/time step:

\begin{equation}
    \label{eqn:minigrid-reward-fn}
    goal\_reward = 1 - 0.9 \times \frac{es}{ms}
\end{equation}

where $es$ defines the number of steps taken to navigate to the goal (a green color square), $ms$ is the maximum number of steps the agent is allowed to take in an episode. For \emph{MG10} curriculum, five pre-defined grid-worlds with two seed instances/variants (seed 860 and 861 was employed) per environment (hence 10 tasks) was employed. They are: $\mathrm{SimpleCrossingS9N1}$, $\mathrm{SimpleCrossingS9N2}$, $\mathrm{SimpleCrossingS9N3}$, $\mathrm{LavaCrossingS9N1}$, $\mathrm{LavaCrossingS9N2}$. Figure \ref{fig:env-mingrid} presents a visual illustration of the 5 grid worlds from which the tasks are derived. Note that when an agent steps a on lava (depicted in orange in the figure), the episode is terminated.

Although the default action space in Minigrid is 7, the action space was set to 3 (turn left, turn right, and move forward) in this work as only navigational capabilities were required by the agents across all tasks in the \emph{MG10} curriculum (i.e., other actions such as pick object, drop object, toggle and done actions were not necessary). Furthermore, the reduced action space eased the exploration demands across all methods when learning each task.

\subsection{Continual World}
\label{apndx:env-continual-world}
\begin{figure}
    \centering
    \includegraphics[width=0.7\textwidth]{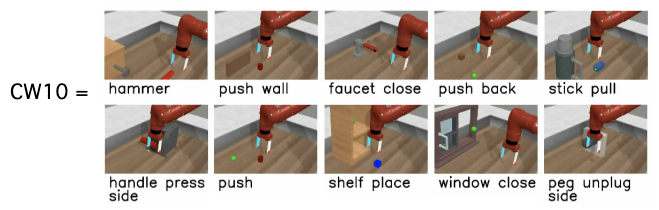}
    \caption{Visual representation of the 10 tasks \emph{CW10} \citep{wolczyk2021continualworld}}
    \label{fig:env_continual_world}
\end{figure}

The Continual World \citep{wolczyk2021continualworld} is a benchmark for lifelong/continual RL derived from the Meta-World environment \citep{yu2020meta} --- a benchmark consisting of 50 distinct simulated robotic tasks developed using the MuJoCo physics simulator \citep{todorov2012mujoco}. The \emph{CW10} curriculum in the benchmark comes from 10 tasks selected from the Meta World, with the goal of having a high variance in forward transfer across tasks. The 10 tasks (see Figure \ref{fig:env_continual_world}) are: hammer-v2, push-wall-v2, faucet-close-v2, push-back-v2, stick-pull-v2, handle-press-side-v2, push-v2, shelf-
place-v2, window-close-v2, peg-unplug-side-v2. The input/state space of each task is a 39 dimension vector representation (consisting of proprioceptor information of the robotic arm as well as position of the objects and goal location in the environment), with an action space of 4 that defines the movement of the robotic arm. The reward function is defined based on a multi-component structure where the agent is reward for achieve sub-goals (i.e., reaching objects, gripping objects, and placing objects or a subset of these) within each task. In addition to the reward, another metric called \emph{success metric} is used to measure performance --- where the agent gets a 1 if it solves the overall task or 0 otherwise.

Note that when the Continual World benchmark was released, the authors used what is now termed version 1 (v1) environments in the Meta-World. However, the Meta-World v1 environments contained some issues in the reward function \footnote{as discussed in \url{https://github.com/rlworkgroup/metaworld/issues/226} and \url{https://github.com/awarelab/continual_world/issues/2}} which was fixed in the updated v2 environments. Therefore, the experiments in the paper employed the use of the v2 environment for each task in the Continual World.

\subsection{ProcGen}
\label{apndx:env-procgen}
\begin{figure}[h]
    \centering
    \includegraphics[width=0.9\textwidth]{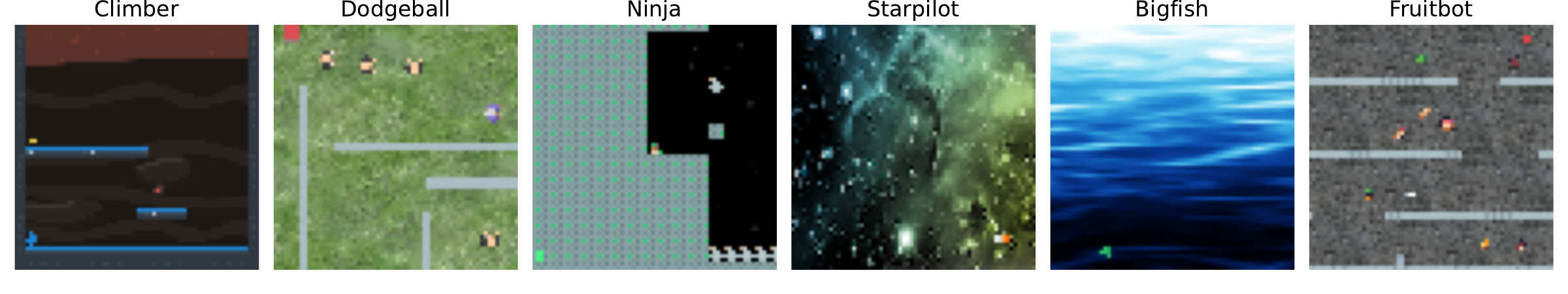}
    \caption{A snapshot of the tasks in the ProcGen curriculum. The texture, objects, RGB color, structure are procedurally generated.}
    \label{fig:env_procgen}
\end{figure}

The ProcGen \citep{cobbe2020procgen} is a discrete action benchmark that consist of 16 visual diverse video game tasks that are procedurally generated and computationally fast to run, with the aim of evaluating generalization ability of RL agents. It was proposed as a replacement of the Atari games benchmark, while being computationally faster to simulate than Atari. The benchmark was adapted for lifelong RL by \citep{powers2022cora} which introduced a lifelong RL curriculum based on a subset of the ProcGen games. The selected games are Climber, Dodgeball, Ninja, Starpilot, Bigfish, and Fruitbot as shown in Figure \ref{fig:env_procgen}. The input observation are RGB images of dimension $64 \times 64 \times 3$, along with 15 possible discrete actions. Also, the reward function and scales (range of values) are different for each task in the curriculum.

Due to the procedural nature of the environment, each game contains several levels and the properties of each game instance (such as objects, texture maps, layout, enemies etc) can be procedurally generated, thus ensuring high variance within each game. The procedural nature of the environment facilitates testing of lifelong RL agents in unseen environments, thus evaluating also generalization capabilities. Variation in tasks exists across the state and transition distributions.

\section{Additional Analysis}
\label{apndx:additional-analysis}
\subsection{Modulatory Mask Similarities}
\label{apndx:analyis-mask-similarities}
If similarities in tasks allow for our approach to exploit a linear combination of masks, it is reasonable to ask whether masks do reflect such similarity. We consider the two cases of: (1) $\mathrm{Mask_{RI}}$ where each mask is initialized randomly and (2) $\mathrm{Mask_{LC}}$ where each mask is a combination of a random mask and known masks. The analysis was conducted on masks learned in the \emph{CT8} curriculum. 

\begin{figure}[ht]
    \centering
    \includegraphics[width=0.8\textwidth]{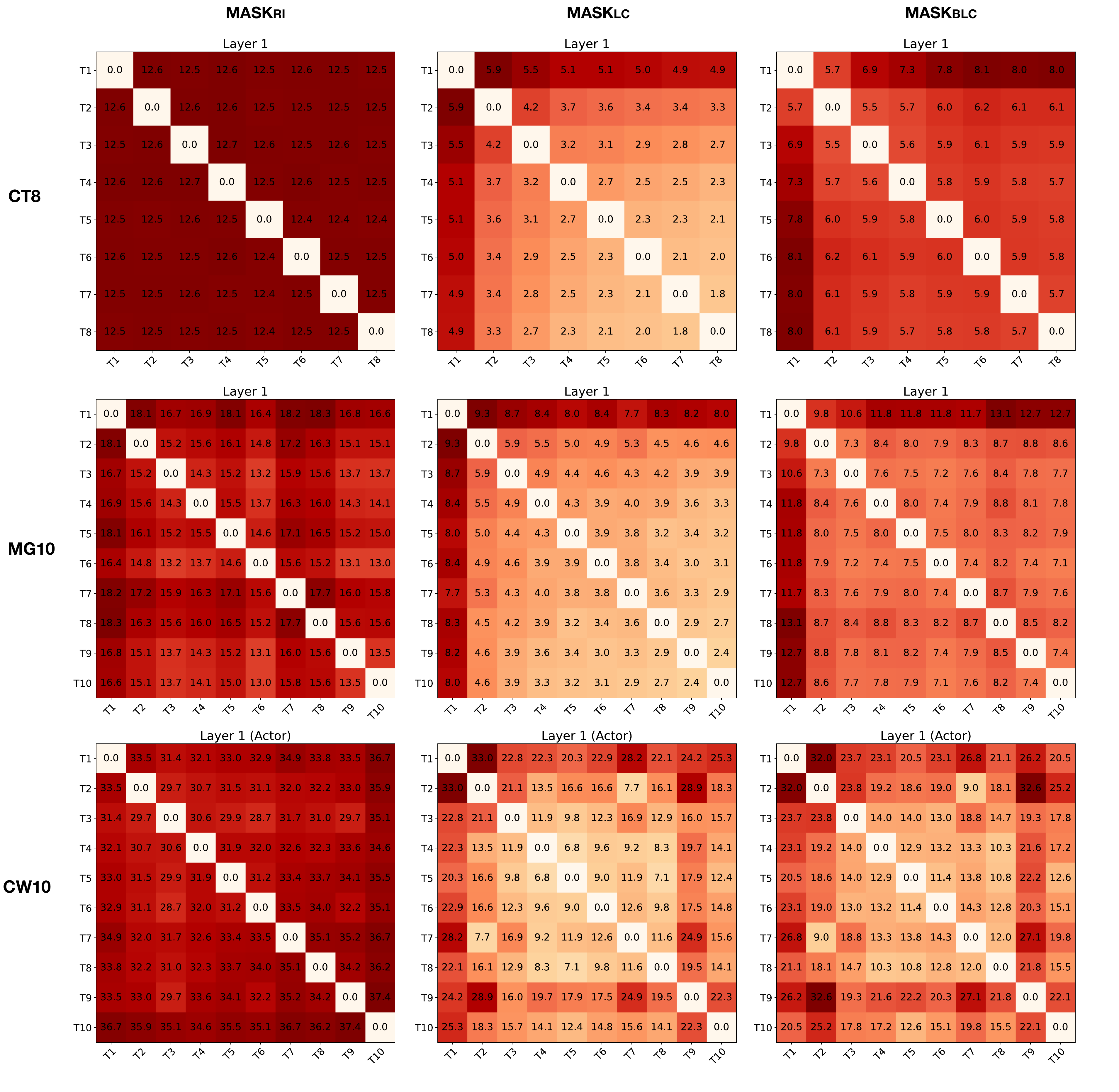}
    \caption{Pairwise mask distances between tasks (i.e., each plot computed as the L2 norm of the difference between task masks) in the first layer of the policy network for each modulatory masking method. Despite tasks having similarities in the CT-graph, Mingrid and Continual World curricula, in $\mathrm{MASK_{RI}}$ (Left), the learned masks across tasks show no correlation, $\mathrm{MASK_{LC}}$ (Middle) and $\mathrm{MASK_{LC}}$ (Right) show mask correlation across tasks (benefiting from knowledge re-use).}
    \label{fig:mask_sim}
\end{figure}

Figure \ref{fig:mask_sim} shows that, despite task similarities, random initialization of masks results in dissimilar masks. This result is expected as independent gradient optimizations will lead generally to different solutions. However, the linear combination of previously known masks is exploited in the tuning of new masks as we observed that the last two mask are significantly more similar to each other than the first two.  

\subsection{Linear Combination Coefficients}
\label{apndx:lcomb-coeffs}
\begin{figure}
    \centering
    \includegraphics[width=0.8\textwidth]{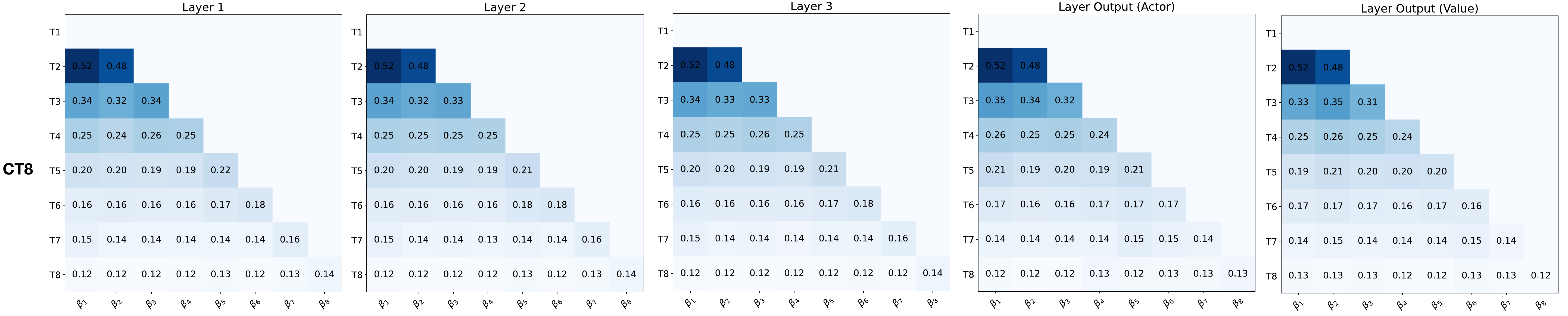}
    \includegraphics[width=0.8\textwidth]{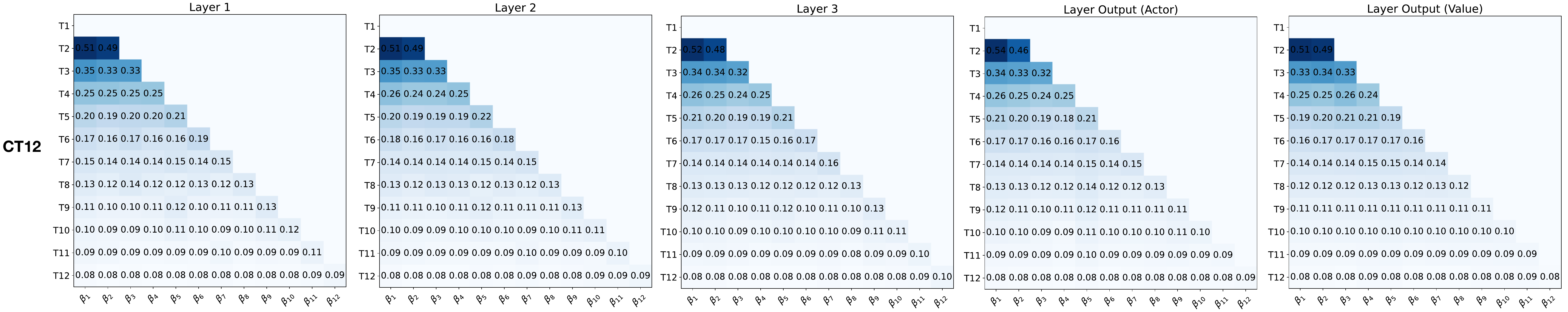}
    \includegraphics[width=0.8\textwidth]{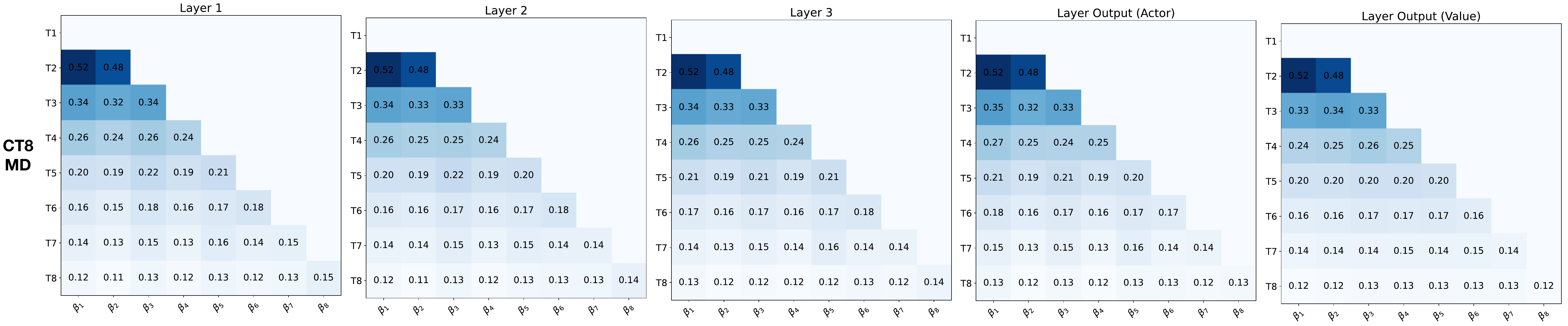}
    \includegraphics[width=0.8\textwidth]{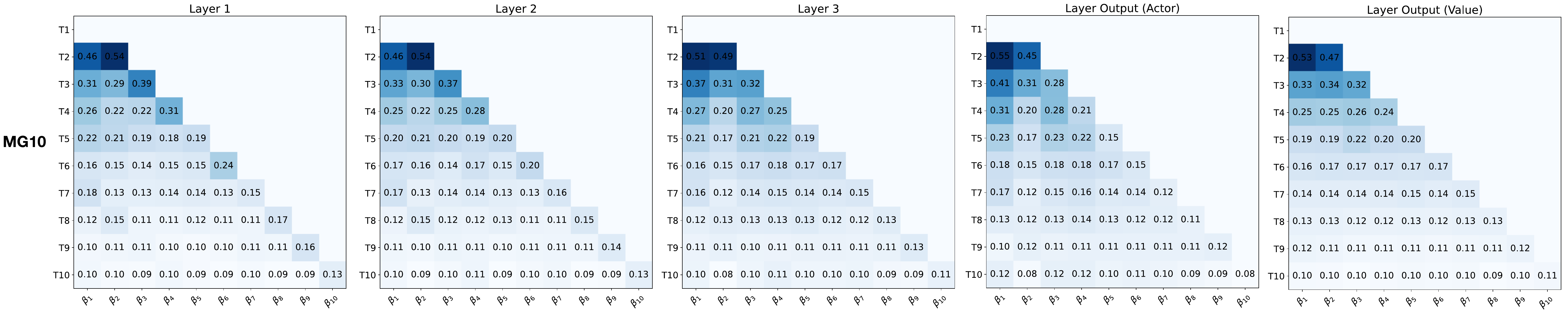}
    \caption{Per layer coefficients $\mathbf{\beta}$ in $\mathrm{Mask_{LC}}$ after training on the \emph{CT8}, \emph{CT12}, \emph{CT8 multi depth}, and \emph{MG10} curricula.}
    \label{fig:lcomb-coeff-ct8-ct12-ct8md-mg10}
\end{figure}

\begin{figure}
    \centering
    \includegraphics[width=0.8\textwidth]{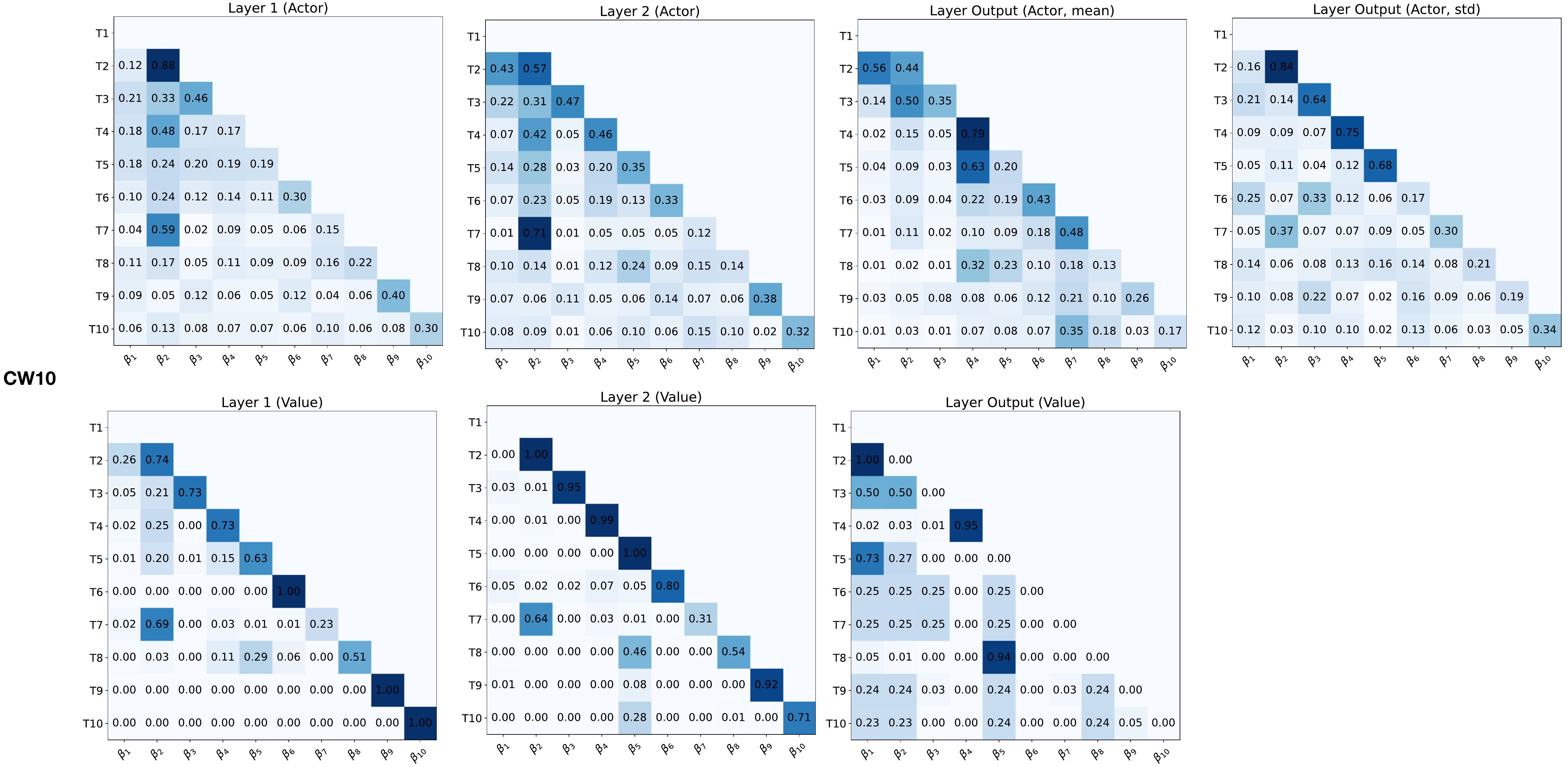}
    \caption{Per layer coefficients $\mathbf{\beta}$ in $\mathrm{Mask_{LC}}$ after training on the \emph{CW10} curriculum.}
    \label{fig:lcomb-coeff-cw10}
\end{figure}

In Section \ref{subsec:analyis-linear-comb-coeffs}, Figure \ref{fig:summary_mask_lc_lcomb_coeff} showed a summary of the linear combination co-efficients of the input and output layers of the $\mathrm{MASK_{LC}}$ network after training. For completeness, this section presents the co-efficients for all layers in the network, across the \emph{CT8}, \emph{CT12}, \emph{CT8 multi depth}, \emph{MG10}, \emph{CW10} curricula. The co-efficients are presented in Figures \ref{fig:lcomb-coeff-ct8-ct12-ct8md-mg10} and \ref{fig:lcomb-coeff-cw10}.

\section{Additional Results}
\label{apndx:additional-results}
\subsection{EWC Single versus Multi-Head Policy Network}
\label{apndx:ewc-sh-mh}
\begin{figure}[htbp]
    \centering
    \includegraphics[width=0.5\textwidth]{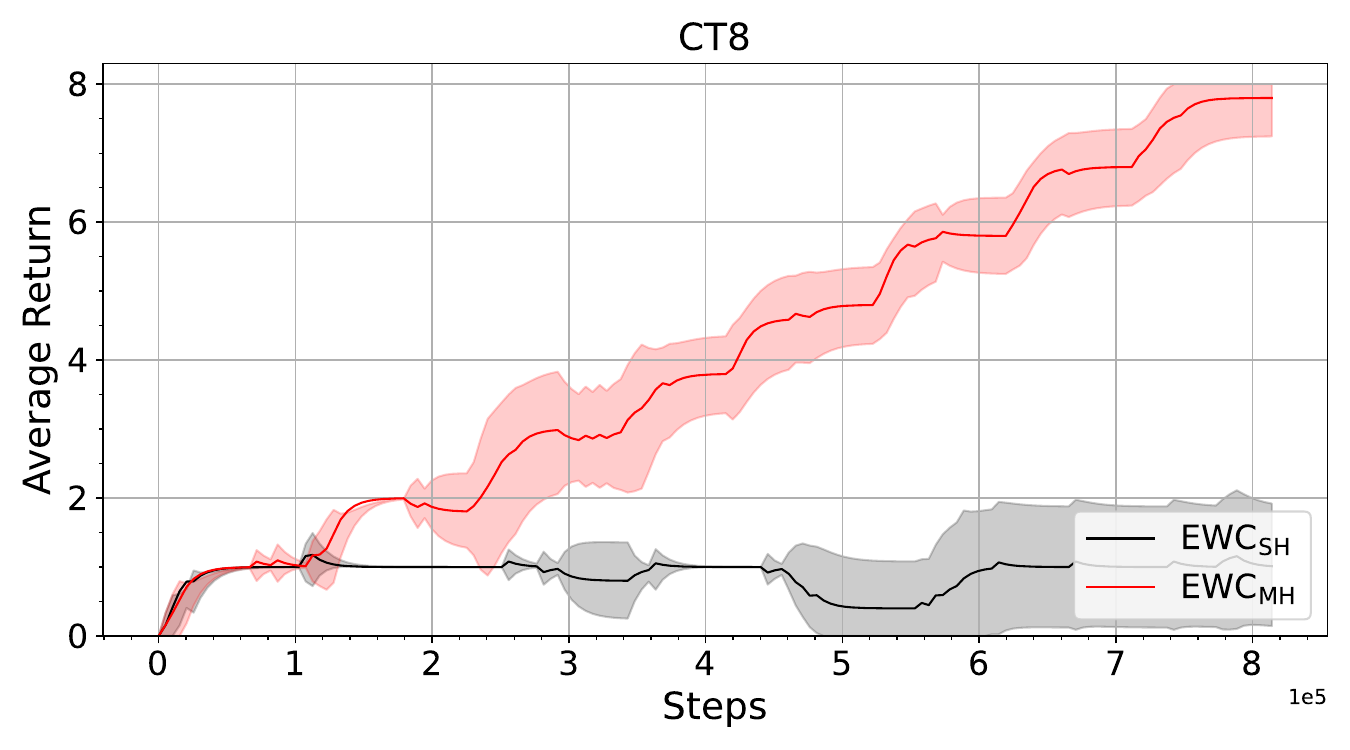}
    \caption{Continual evaluation comparison of EWC single head and multi-head policy networks in the \emph{CT8} curriculum.}
    \label{res:ct8_plot_eval_ewc_sh_mh}
\end{figure}

As highlighted in Section \ref{sec:experiments}, the results for the EWC lifelong RL agents presented were based on multi-head (multiple output layers) policy networks, while other methods employed a single head policy network. This is because the EWC single head network $\mathrm{EWC_{SH}}$ performed sub-optimally. In the CT-graph \emph{CT8} curriculum, Figure \ref{res:ct8_plot_eval_ewc_sh_mh} presents the continual evaluation comparison between the $\mathrm{EWC_{SH}}$ and $\mathrm{EWC_{MH}}$.

\subsection{Train plot for all methods}
\label{apndx:train-plots-all-methods}
\begin{figure}[h]
    \centering
    \begin{subfigure}{0.48\textwidth}
        \centering
        \includegraphics[width=\textwidth]{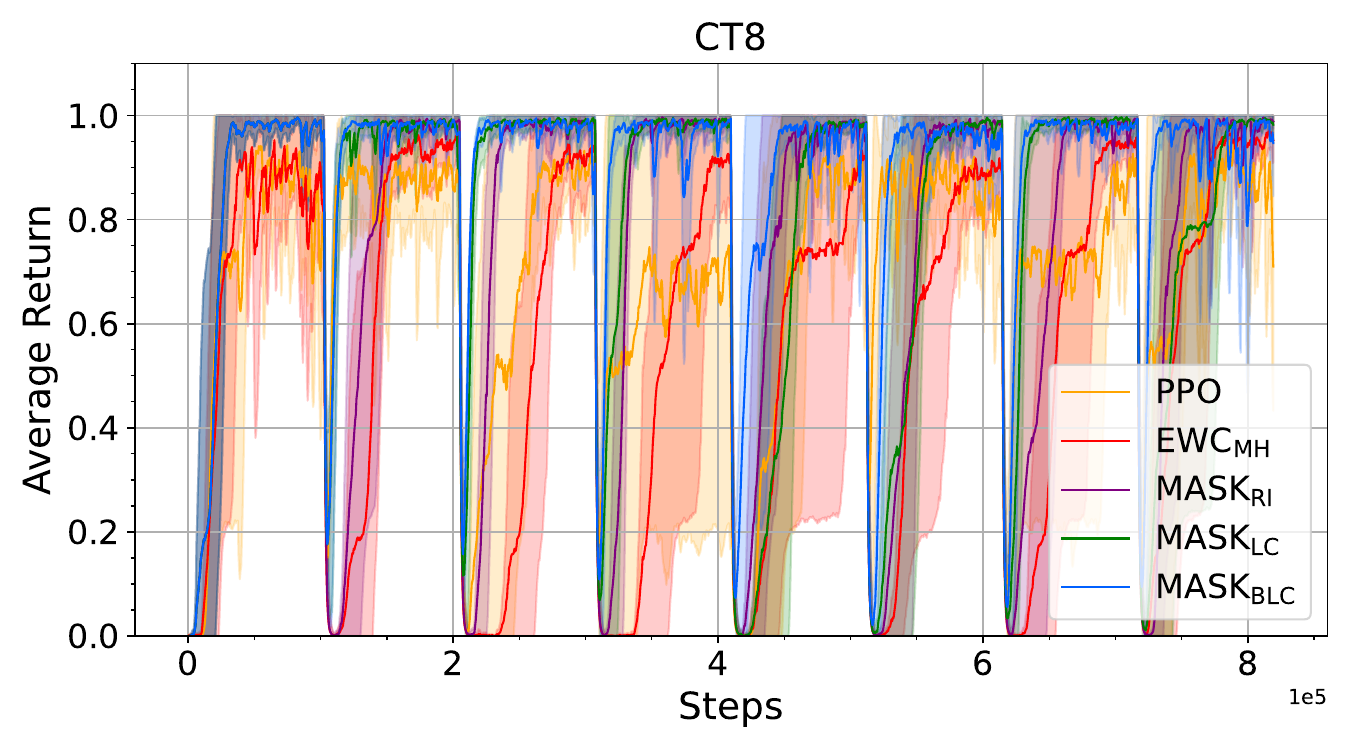}
        \caption{}
        \label{res:ct8_plot_train_all_methods}
    \end{subfigure}
    \begin{subfigure}{0.48\textwidth}
        \centering
        \includegraphics[width=\textwidth]{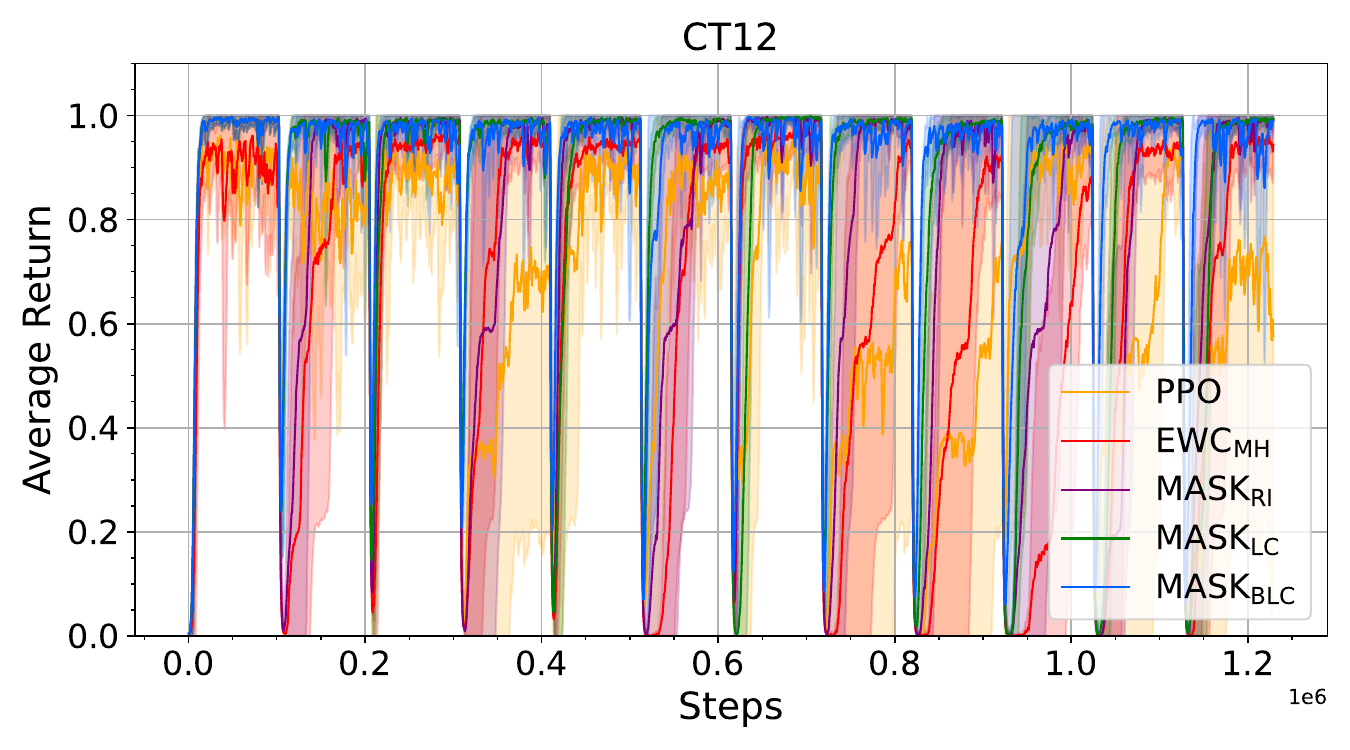}
        \caption{}
        \label{res:ct12_plot_train_all_methods}
    \end{subfigure}
    
    \begin{subfigure}{0.48\textwidth}
        \centering
        \includegraphics[width=\textwidth]{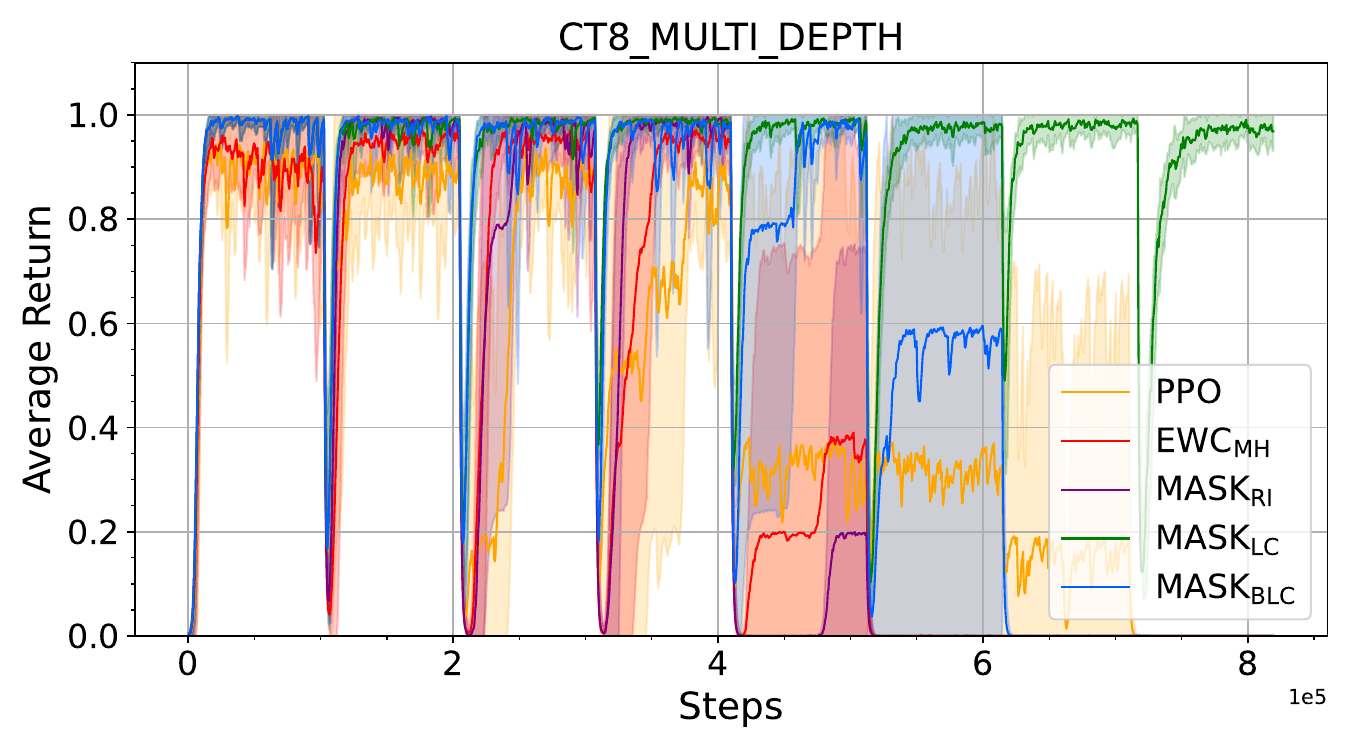}
        \caption{}
        \label{res:ct8_multi_depth_plot_train_all_methods}
    \end{subfigure}
    \begin{subfigure}{0.48\textwidth}
        \centering
        \includegraphics[width=\textwidth]{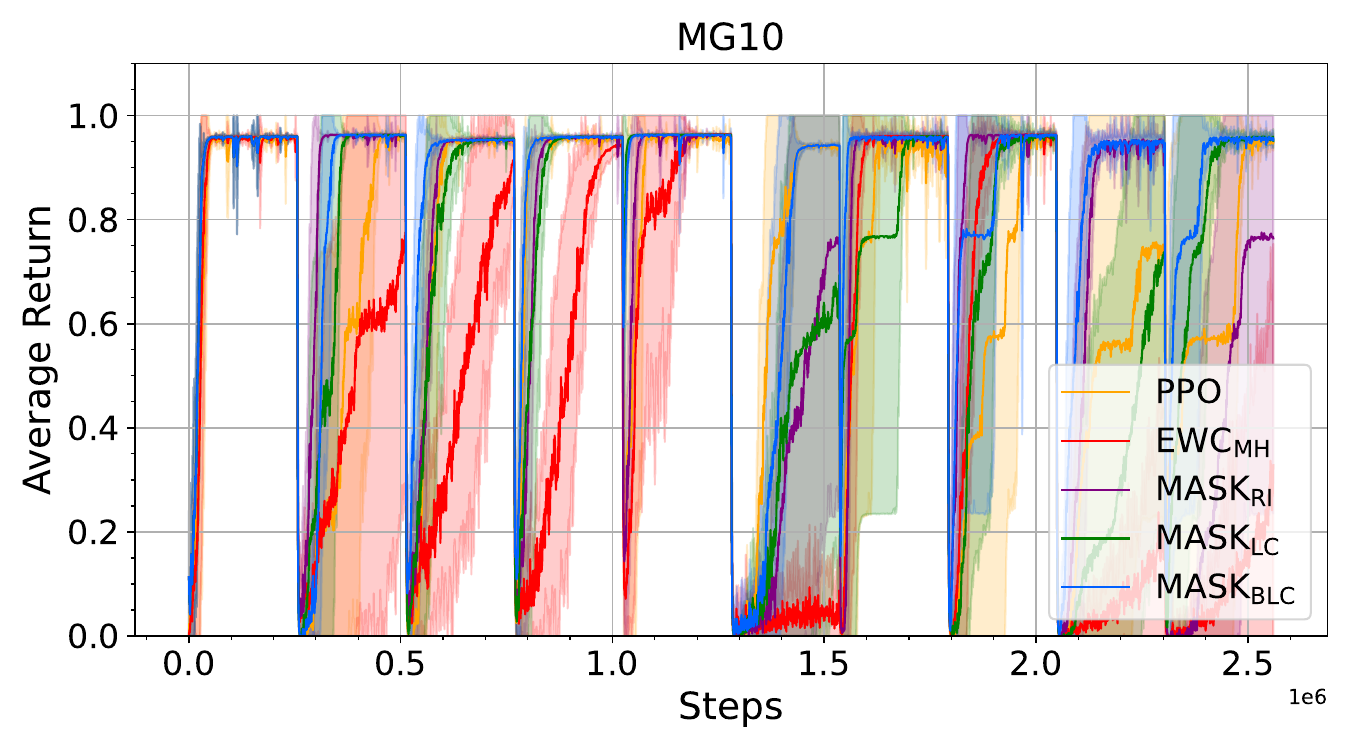}
        \caption{}
        \label{res:mg10_plot_train_all_methods}
    \end{subfigure}

    \begin{subfigure}{0.48\textwidth}
        \centering
        \includegraphics[width=\textwidth]{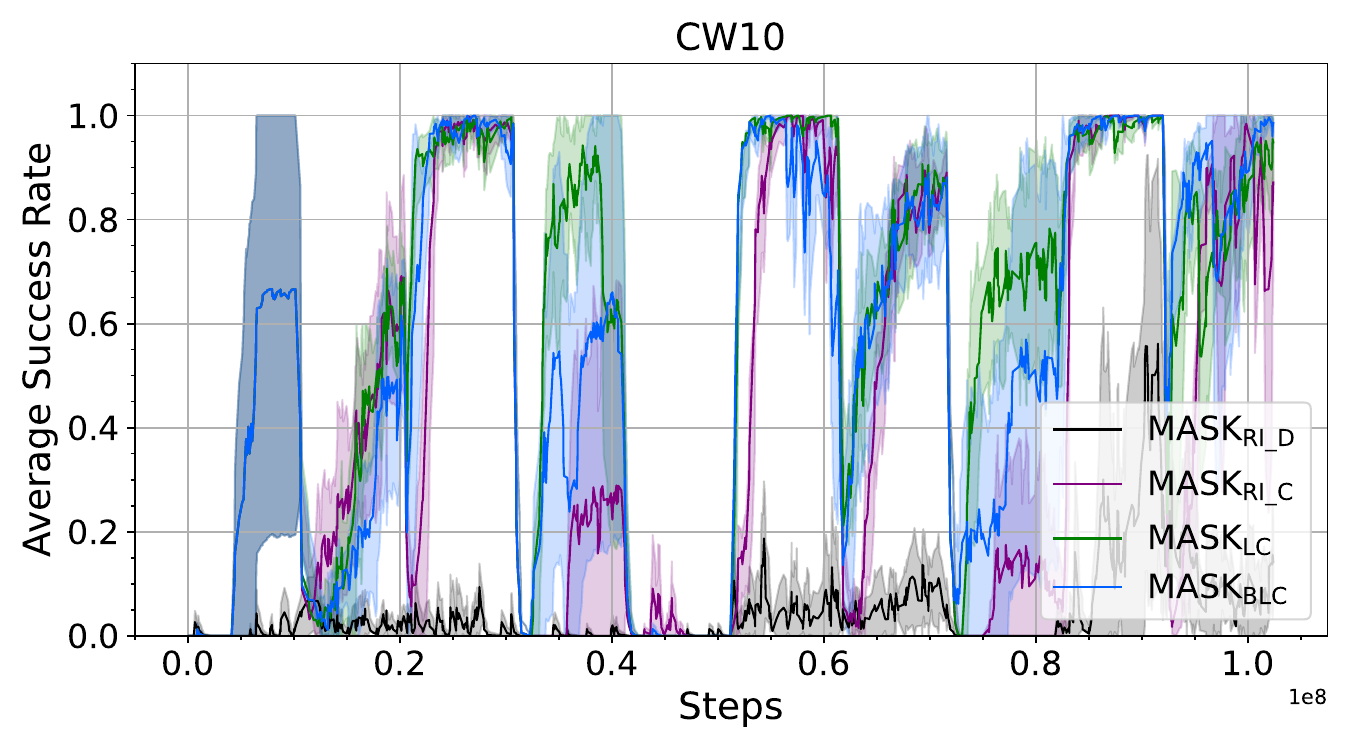}
        \caption{}
        \label{res:cw10_plot_train_all_methods}
    \end{subfigure}
    \caption{Lifelong training plots for all methods and baselines in CTgraph (CT), Minigrid (MG) and Continual World (CW) curricula: (a) \emph{CT8}, (b) \emph{CT12}, (c) \emph{CT8 multi depth}, (d) \emph{MG10}, and (e) \emph{CW10}.}
    \label{res:ct_mg_cw_plot_train_all_methods}
\end{figure}

In the lifelong training plots reported in Section \ref{sec:experiments}, only the masking methods were presented for the sake of clarity and readability. The plots in Figure \ref{res:ct_mg_cw_plot_train_all_methods} present the lifelong training plots containing all methods across the CT-graph, Minigrid and Continual World curricula.

\subsection{Per Task Forward Transfer}
\label{apndx:per-task-fwd-trsnf}
In the main text, the forward transfer metric was reported as the averaged across seed runs and tasks in the CT-graph, Minigrid and Continual World curricula. The reported information is expanded in this section to show the forward transfer metric per task (averaged across seed runs only), and reported in Tables \ref{tab:fwd-trnsf-per-task-ct8}, \ref{tab:fwd-trnsf-per-task-ct12}, \ref{tab:fwd-trnsf-per-task-ct8-md}, \ref{tab:fwd-trnsf-per-task-mg10}, and \ref{tab:fwd-trnsf-per-task-cw10}. The average across tasks is reported in the last column of each table. As noted in the main text, the tasks are learned independently of other tasks in $\mathrm{Mask_{RI}}$, thus they are omitted in the tables.

\begin{table}[h]
    \centering
    \begin{tabular}{lccccccccc}
        \toprule
        {} & \multicolumn{8}{c}{Tasks} & {} \\
        Method & 1 & 2 & 3 & 4 & 5 & 6 & 7 & 8 & Avg\\
        \midrule
        PPO & -0.03 & 0.62 & -0.28 & -0.19 & 0.06 & 0.58 & 0.17 & 0.23 & 0.15 \\
        $\mathrm{EWC_{MH}}$ & -0.04 & 0.11 & -0.91 & -0.67 & -0.06 & -0.05 & -0.22 & -0.04 & -0.23 \\
        $\mathrm{MASK_{LC}}$ & 0.34 & 0.81 & 0.73 & 0.62 & 0.13 & 0.28 & 0.65 & 0.15 & 0.46 \\
        $\mathrm{MASK_{BLC}}$ & 0.34 & 0.81 & 0.75 & 0.73 & 0.64 & 0.66 & 0.72 & 0.69 & 0.67 \\
        \bottomrule
    \end{tabular}
    \caption{Forward transfer per task in the \emph{CT8} curriculum, averaged across seed runs.}
    \label{tab:fwd-trnsf-per-task-ct8}
\end{table}

\begin{table}[h]
    \centering
    \begin{tabular}{lccccccccccccc}
        \toprule
        {} & \multicolumn{12}{c}{Tasks} & {} \\
        Method & 1 & 2 & 3 & 4 & 5 & 6 & 7 & 8 & 9 & 10 & 11 & 12 & Avg\\
        \midrule
        PPO & -0.02 & 0.30 & -0.04 & -0.46 & -0.33 & 0.12 & -0.46 & -0.49 & -0.38 & 0.48 & -0.17 & -0.36 & -0.15 \\
        $\mathrm{EWC_{MH}}$ & -0.10 & -0.22 & 0.05 & 0.21 & -0.02 & -0.49 & 0.30 & -0.44 & -0.12 & -0.89 & 0.13 & 0.24 & -0.11 \\
        $\mathrm{MASK_{LC}}$ & 0.50 & 0.77 & 0.47 & 0.83 & 0.18 & 0.75 & -0.05 & 0.74 & 0.66 & 0.34 & 0.48 & 0.36 & 0.50 \\
        $\mathrm{MASK_{BLC}}$ & 0.50 & 0.75 & 0.57 & 0.79 & 0.55 & 0.59 & 0.45 & 0.74 & 0.76 & 0.58 & 0.77 & 0.72 & 0.65 \\
        \bottomrule
    \end{tabular}
    \caption{Forward transfer per task in the \emph{CT12} curriculum, averaged across seed runs.}
    \label{tab:fwd-trnsf-per-task-ct12}
\end{table}

\begin{table}[h]
    \centering
    \begin{tabular}{lccccccccc}
        \toprule
        {} & \multicolumn{8}{c}{Tasks} & {} \\
        Method & 1 & 2 & 3 & 4 & 5 & 6 & 7 & 8 & Avg\\
        \midrule
        PPO & -0.10 & -0.14 & -0.23 & 0.10 & 0.31 & 0.31 & 0.14 & 0.00 & 0.05 \\
        $\mathrm{EWC_{MH}}$ & -0.05 & 0.06 & 0.30 & 0.28 & 0.23 & 0.00 & 0.00 & -0.00 & 0.10 \\
        $\mathrm{MASK_{LC}}$ & 0.45 & 0.54 & 0.76 & 0.87 & 0.94 & 0.89 & 0.95 & 0.87 & 0.79 \\
        $\mathrm{MASK_{BLC}}$ & 0.45 & 0.55 & 0.73 & 0.79 & 0.83 & 0.51 & 0.00 & 0.00 & 0.48 \\
        \bottomrule
    \end{tabular}
    \caption{Forward transfer per task in the \emph{CT8 multi depth} curriculum, averaged across seed runs.}
    \label{tab:fwd-trnsf-per-task-ct8-md}
\end{table}

\begin{table}[h]
    \centering
    \begin{tabular}{lccccccccccc}
        \toprule
        {} & \multicolumn{10}{c}{Tasks} & {} \\
        Method & 1 & 2 & 3 & 4 & 5 & 6 & 7 & 8 & 9 & 10 & Avg\\
        \midrule
        PPO & 0.00 & -1.34 & -0.17 & 0.40 & -0.20 & 0.41 & -0.55 & -1.83 & -0.96 & 0.25 & -0.40 \\
        $\mathrm{EWC_{MH}}$ & -0.01 & -2.05 & -1.60 & -1.47 & -0.65 & -0.58 & -0.10 & -0.62 & -2.27 & -0.99 & -1.04 \\
        $\mathrm{MASK_{LC}}$ & 0.17 & -0.48 & -0.17 & 0.10 & 0.60 & -0.14 & -0.71 & -1.01 & -1.43 & 0.56 & -0.25 \\
        $\mathrm{MASK_{BLC}}$ & 0.17 & -0.24 & 0.37 & 0.47 & 0.56 & 0.25 & 0.48 & -0.30 & 0.31 & 0.68 & 0.27 \\
        \bottomrule
    \end{tabular}
    \caption{Forward transfer per task in the \emph{MG10} curriculum, averaged across seed runs.}
    \label{tab:fwd-trnsf-per-task-mg10}
\end{table}

\begin{table}[h]
    \centering
    \begin{tabular}{lccccccccccc}
        \toprule
        {} & \multicolumn{10}{c}{Tasks} & {} \\
        Method & 1 & 2 & 3 & 4 & 5 & 6 & 7 & 8 & 9 & 10 & Avg\\
        \midrule
        PPO & -0.87 & -1.27 & -1.06 & 0.25 & -0.00 & -13.79 & -2.07 & -1.42 & -12.70 & -7.68 & -4.06 \\
        $\mathrm{EWC_{MH}}$ & -3.01 & -1.44 & -9.47 & 0.00 & -0.00 & -17.60 & -2.22 & -1.42 & -30.07 & -8.65 & -7.39 \\
        $\mathrm{MASK_{LC}}$ & -1.91 & -0.68 & 0.12 & 0.63 & -0.00 & 0.54 & -0.08 & -0.10 & -0.44 & -1.36 & -0.33 \\
        $\mathrm{MASK_{BLC}}$ & -1.91 & -0.89 & -0.23 & 0.40 & -0.00 & -1.84 & -0.15 & -0.63 & -0.34 & -0.51 & -0.61 \\
        \bottomrule
    \end{tabular}
    \caption{Forward transfer per task in the \emph{CW10} curriculum, averaged across seed runs.}
    \label{tab:fwd-trnsf-per-task-cw10}
\end{table}

\subsection{ProcGen: Per Task Forward Transfer Metric}
\label{apndx:procgen-per-task-forward-transfer}
The per task forward transfer metric for all methods except $\mathrm{MASK_{RI}}$ in the ProcGen curriculum is presented in Table \ref{tab:procgen_transfer_metrics}. Note that $\mathrm{MASK_{RI}}$ was omitted because the method does not inherently foster forward transfer as each task is learned independently of other tasks (i.e., for each task, a separate modulatory mask is independently initialized and optimized for the task).

\begin{table}[ht!]
\setlength\tabcolsep{2pt} 
\tiny
\input{procgen_metrics/procgen_transfer.tex}
\vspace{1em}
\caption{ProcGen transfer metrics.}
\label{tab:procgen_transfer_metrics}
\end{table}

\section{Learned Modulating Masks and Memory Requirements}
\label{appendx:learned-mask-and-memory}
In the current study, the binarization process, key to reduce memory use, was successful in the discrete benchmarks and only after performing the linear combination in $\mathrm{Mask_{LC}}$ and $\mathrm{Mask_{BLC}}$. Binarized masks resulted in poor performance in the continuous value Continual World benchmark, and if binarization was performed before the linear combination. Further studies could investigate this issue in more detail. It is possible that binarized masks could result in good performance if only the head layer was made continuous, thus ensuring smoothness in the output.  

A different approach to reduce memory consumption is to take advantage of the apparent equal representations of known masks (as shown in Figure \ref{fig:summary_mask_lc_lcomb_coeff}). If the advantage of previous knowledge can be represented as an average of previous masks, it is possible to modify the algorithm to maintain only a moving average of all previous masks. In such a case, the algorithm will combine a new mask with the average of all previous masks. This extreme version of the algorithm is memory efficient, but may under-perform in curricula where coefficients are tuned to be diverse as in the CW10 benchmark (Figure \ref{fig:summary_mask_lc_lcomb_coeff} right-most column). Building on this idea, a limited number of \emph{template} masks can be used instead of a single average mask. Each template can be a running average of a cluster of tasks, simply determined by L2 distances of masks, that will ensure good forward transfer while maintaining scalability.

One further approach to reduce memory is to experiment with high level of sparsity in the masks (Equations \ref{eqn:score-to-binary-mask} and \ref{eqn:score-to-continuous-mask}). Increasing the threshold (currently set to 0), or applying top k-winners as in \citep{wortsman2020supermasks} for supervised learning, may lead to a meta optimization process where significantly smaller masks maintain acceptable levels of performance. In summary, while more work is required to improve the memory efficiency of the proposed approaches, the success of the linear combination methods suggests venues of research to reduce memory consumption, while the performance advantages justify further research of masking methods in LRL.

\section{Time taken to reach X\% of Optimal (Target) Performance}
\label{apndx:target-performance}
From the training plots in Figure \ref{res:ct_mg_cw_plot_train_all_methods}, the time taken or training efficiency (i.e., number of training steps) per task to achieve a certain level of performance (i.e., X\% of the optimal performance) can be derived. Note that if an agent fails to achieve the specified performance level during a task training, then the agent is said to have failed that task. In the CT-graph and Minigrid curricula, a target of 75\% of the optimal performance was used to conduct the analysis, while 50\% was employed in the Continual World curriculum. The results of the analysis are reported in Tables \ref{tab:target-perf-each-task-ct8}, \ref{tab:target-perf-each-task-ct12}, \ref{tab:target-perf-each-task-ct8-md}, \ref{tab:target-perf-each-task-mg10}, and \ref{tab:target-perf-each-task-cw10}.
\begin{table}
    \centering
    \begin{tabular}{lccccccccc}
        \toprule
        {} & \multicolumn{8}{c}{Tasks ($step \times 10^3$)} & {} \\
        Method & 1 & 2 & 3 & 4 & 5 & 6 & 7 & 8 & Avg\\
        \midrule
        PPO & 23 & 6 & 30 & Fail & 35 & 5 & 23 & 16 & Fail \\
        $\mathrm{EWC_{MH}}$ & 23 & 36 & 54 & 54 & 46 & 40 & 44 & 30 & 41 \\
        $\mathrm{MASK_{RI}}$ & 20 & 28 & 21 & 20 & 25 & 30 & 28 & 21 & 24 \\
        $\mathrm{MASK_{LC}}$ & 20 & 7 & 8 & 13 & 39 & 30 & 13 & 28 & 20 \\
        $\mathrm{MASK_{BLC}}$ & 20 & 8 & 6 & 7 & 14 & 12 & 9 & 8 & 10 \\
        \bottomrule
    \end{tabular}
    \caption{Number of training steps taken to achieve 75\% optimal (target) performance per task in the \emph{CT8} curriculum. Steps rounded to the nearest thousand.}
    \label{tab:target-perf-each-task-ct8}
\end{table}
\begin{table}
    \centering
    \begin{tabular}{lccccccccccccc}
        \toprule
        {} & \multicolumn{12}{c}{Tasks ($steps \times 10^3$)} & {} \\
        Method & 1 & 2 & 3 & 4 & 5 & 6 & 7 & 8 & 9 & 10 & 11 & 12 & Avg \\
        \midrule
        PPO & 8 & 6 & 6 & Fail & 14 & 18 & 18 & Fail & Fail & 11 & 40 & Fail & Fail \\
        $\mathrm{EWC_{MH}}$ & 9 & 33 & 13 & 25 & 14 & 41 & 9 & 47 & 52 & 69 & 30 & 31 & 31 \\
        $\mathrm{MASK_{RI}}$ & 8 & 23 & 8 & 27 & 9 & 33 & 8 & 24 & 24 & 39 & 35 & 21 & 22 \\
        $\mathrm{MASK_{LC}}$ & 8 & 6 & 8 & 6 & 14 & 8 & 18 & 9 & 15 & 22 & 21 & 25 & 13 \\
        $\mathrm{MASK_{BLC}}$ & 8 & 6 & 5 & 6 & 6 & 11 & 8 & 7 & 9 & 14 & 8 & 9 & 8 \\
        \bottomrule
    \end{tabular}
    \caption{Number of training steps taken to achieve 75\% optimal (target) performance per task in the \emph{CT12} curriculum. Steps rounded to the nearest thousand.}
    \label{tab:target-perf-each-task-ct12}
\end{table}
\begin{table}
    \centering
    \begin{tabular}{lccccccccc}
        \toprule
        {} & \multicolumn{8}{c}{Tasks ($step \times 10^3$)} & {} \\
        Method & 1 & 2 & 3 & 4 & 5 & 6 & 7 & 8 & Avg\\
        \midrule
        PPO & 8 & 8 & 31 & 26 & Fail & Fail & Fail & Fail & Fail \\
        $\mathrm{EWC_{MH}}$ & 8 & 12 & 18 & 25 & Fail & Fail & Fail & Fail & Fail \\
        $\mathrm{MASK_{RI}}$ & 7 & 8 & 22 & 18 & Fail & Fail & Fail & Fail & Fail \\
        $\mathrm{MASK_{LC}}$ & 7 & 6 & 7 & 5 & 6 & 10 & 4 & 13 & 7 \\
        $\mathrm{MASK_{BLC}}$ & 7 & 6 & 7 & 6 & 16 & Fail & Fail & Fail & Fail \\
        \bottomrule
    \end{tabular}
    \caption{Number of training steps taken to achieve 75\% optimal (target) performance per task in the \emph{CT8 multi depth} curriculum. Steps rounded to the nearest thousand.}
    \label{tab:target-perf-each-task-ct8-md}
\end{table}
\begin{table}
    \centering
    \begin{tabular}{lccccccccccc}
        \toprule
        {} & \multicolumn{10}{c}{Tasks ($steps \times 10^3$)} & {} \\
        Method & 1 & 2 & 3 & 4 & 5 & 6 & 7 & 8 & 9 & 10 & Avg \\
        \midrule
        PPO & 29 & 118 & 54 & 19 & 23 & 91 & 36 & 92 & Fail & 79 & Fail \\
        $\mathrm{EWC_{MH}}$ & 28 & Fail & 142 & 127 & 46 & Fail & 31 & 47 & Fail & Fail & Fail \\
        $\mathrm{MASK_{RI}}$ & 22 & 38 & 49 & 37 & 18 & Fail & 23 & 23 & 63 & Fail & Fail \\
        $\mathrm{MASK_{LC}}$ & 22 & 71 & 55 & 37 & 2 & Fail & 41 & 64 & Fail & 47 & Fail \\
        $\mathrm{MASK_{BLC}}$ & 22 & 60 & 26 & 18 & 3 & 113 & 8 & 38 & 42 & 28 & 36 \\
        \bottomrule
    \end{tabular}
    \caption{Number of training steps taken to achieve 75\% optimal (target) performance per task in the \emph{MG10} curriculum. Steps rounded to the nearest thousand.}
    \label{tab:target-perf-each-task-mg10}
\end{table}
\begin{table}
    \centering
    \begin{tabular}{lccccccccccc}
        \toprule
        {} & \multicolumn{10}{c}{Tasks ($steps \times 10^6$)} & {} \\
        Method & 1 & 2 & 3 & 4 & 5 & 6 & 7 & 8 & 9 & 10 & Avg \\
        \midrule
        PPO & 5.4 & 0.1 & 2.5 & 0.1 & Fail & Fail & Fail & Fail & 6.5 & 0.1 & Fail \\
        $\mathrm{EWC_{MH}}$ & 0.7 & Fail & Fail & Fail & Fail & Fail & Fail & Fail & Fail & Fail & Fail \\
        $\mathrm{MASK_{RI\_D}}$ & Fail & Fail & Fail & Fail & Fail & Fail & Fail & Fail & Fail & 0.1 & Fail \\
        $\mathrm{MASK_{RI\_C}}$ & 6.2 & 0.1 & 0.1 & 0.1 & Fail & 2.8 & 0.1 & 0.1 & 2.8 & 0.1 & Fail \\
        $\mathrm{MASK_{LC}}$ & 6.2 & 0.1 & 0.1 & 0.1 & 0.1 & 1.9 & 0.1 & 0.1 & 0.1 & 0.1 & 0.9 \\
        $\mathrm{MASK_{BLC}}$ & 6.2 & 0.1 & 0.1 & 0.1 & 0.1 & 1.9 & 0.1 & 0.1 & 0.3 & 0.1 & 0.9 \\
        \bottomrule
    \end{tabular}
    \caption{Number of training steps taken to achieve 50\% optimal (target) performance per task in the \emph{CW10} curriculum. Steps rounded to the hundred thousand.}
    \label{tab:target-perf-each-task-cw10}
\end{table}

\end{document}

%% file: math_commands.tex
\usepackage{amsmath,amsfonts,bm}

\def\eqref#1{equation~\ref{#1}}

\def\1{\bm{1}}

\DeclareMathAlphabet{\mathsfit}{\encodingdefault}{\sfdefault}{m}{sl}
\SetMathAlphabet{\mathsfit}{bold}{\encodingdefault}{\sfdefault}{bx}{n}



%% file: procgen_metrics/procgen_transfer.tex
\subfloat[IMPALA]{ 
 \begin{tabular}{|l|llllll|l|}
\hline
 & 0-Climb.. & 1-Dodge.. & 2-Ninja & 3-Starp.. & 4-Bigfi.. & 5-Fruit.. & Avg \\
0-Climb.. & \cellcolor{green!0} -- & \cellcolor{green!0} -- & \cellcolor{green!0} -- & \cellcolor{green!0} -- & \cellcolor{green!0} -- & \cellcolor{green!0} -- & \cellcolor{green!0} -- \\
1-Dodge.. & \cellcolor{red!3} -0.9 & \cellcolor{green!0} -- & \cellcolor{green!0} -- & \cellcolor{green!0} -- & \cellcolor{green!0} -- & \cellcolor{green!0} -- & \cellcolor{red!3} -0.9 \\
2-Ninja & \cellcolor{green!10} 2.6 & \cellcolor{red!13} -3.3 & \cellcolor{green!0} -- & \cellcolor{green!0} -- & \cellcolor{green!0} -- & \cellcolor{green!0} -- & \cellcolor{red!1} -0.3 \\
3-Starp.. & \cellcolor{red!2} -0.7 & \cellcolor{green!6} 1.5 & \cellcolor{green!0} 0.1 & \cellcolor{green!0} -- & \cellcolor{green!0} -- & \cellcolor{green!0} -- & \cellcolor{green!1} 0.3 \\
4-Bigfi.. & \cellcolor{green!6} 1.7 & \cellcolor{red!14} -3.6 & \cellcolor{red!5} -1.3 & \cellcolor{red!1} -0.4 & \cellcolor{green!0} -- & \cellcolor{green!0} -- & \cellcolor{red!3} -0.9 \\
5-Fruit.. & \cellcolor{green!12} 3.1 & \cellcolor{red!3} -0.9 & \cellcolor{red!4} -1.0 & \cellcolor{green!2} 0.5 & \cellcolor{red!0} -0.2 & \cellcolor{green!0} -- & \cellcolor{green!1} 0.3 \\
Avg & \cellcolor{green!4} 1.2 & \cellcolor{red!6} -1.6 & \cellcolor{red!2} -0.7 & \cellcolor{red!0} 0.0 & \cellcolor{red!0} -0.2 & \cellcolor{green!0} -- & \cellcolor{red!0} -0.2 \\
\hline
\end{tabular}
}
\hspace{1em}
\subfloat[ONLINE EWC]{ 
 \begin{tabular}{|l|llllll|l|}
\hline
 & 0-Climb.. & 1-Dodge.. & 2-Ninja & 3-Starp.. & 4-Bigfi.. & 5-Fruit.. & Avg \\
0-Climb.. & \cellcolor{green!0} -- & \cellcolor{green!0} -- & \cellcolor{green!0} -- & \cellcolor{green!0} -- & \cellcolor{green!0} -- & \cellcolor{green!0} -- & \cellcolor{green!0} -- \\
1-Dodge.. & \cellcolor{red!0} 0.0 & \cellcolor{green!0} -- & \cellcolor{green!0} -- & \cellcolor{green!0} -- & \cellcolor{green!0} -- & \cellcolor{green!0} -- & \cellcolor{red!0} 0.0 \\
2-Ninja & \cellcolor{green!6} 1.6 & \cellcolor{red!8} -2.2 & \cellcolor{green!0} -- & \cellcolor{green!0} -- & \cellcolor{green!0} -- & \cellcolor{green!0} -- & \cellcolor{red!1} -0.3 \\
3-Starp.. & \cellcolor{red!3} -0.8 & \cellcolor{green!10} 2.7 & \cellcolor{red!0} -0.2 & \cellcolor{green!0} -- & \cellcolor{green!0} -- & \cellcolor{green!0} -- & \cellcolor{green!2} 0.6 \\
4-Bigfi.. & \cellcolor{green!6} 1.7 & \cellcolor{red!13} -3.4 & \cellcolor{green!0} 0.1 & \cellcolor{red!2} -0.6 & \cellcolor{green!0} -- & \cellcolor{green!0} -- & \cellcolor{red!2} -0.5 \\
5-Fruit.. & \cellcolor{green!2} 0.6 & \cellcolor{red!8} -2.2 & \cellcolor{red!0} 0.0 & \cellcolor{green!2} 0.5 & \cellcolor{green!2} 0.7 & \cellcolor{green!0} -- & \cellcolor{red!0} -0.1 \\
Avg & \cellcolor{green!2} 0.6 & \cellcolor{red!5} -1.3 & \cellcolor{red!0} -0.0 & \cellcolor{red!0} -0.0 & \cellcolor{green!2} 0.7 & \cellcolor{green!0} -- & \cellcolor{red!0} -0.1 \\
\hline
\end{tabular}
}

\subfloat[P\&C]{ 
 \begin{tabular}{|l|llllll|l|}
\hline
 & 0-Climb.. & 1-Dodge.. & 2-Ninja & 3-Starp.. & 4-Bigfi.. & 5-Fruit.. & Avg \\
0-Climb.. & \cellcolor{green!0} -- & \cellcolor{green!0} -- & \cellcolor{green!0} -- & \cellcolor{green!0} -- & \cellcolor{green!0} -- & \cellcolor{green!0} -- & \cellcolor{green!0} -- \\
1-Dodge.. & \cellcolor{red!0} 0.0 & \cellcolor{green!0} -- & \cellcolor{green!0} -- & \cellcolor{green!0} -- & \cellcolor{green!0} -- & \cellcolor{green!0} -- & \cellcolor{red!0} 0.0 \\
2-Ninja & \cellcolor{green!12} 3.2 & \cellcolor{green!0} 0.1 & \cellcolor{green!0} -- & \cellcolor{green!0} -- & \cellcolor{green!0} -- & \cellcolor{green!0} -- & \cellcolor{green!6} 1.6 \\
3-Starp.. & \cellcolor{red!16} -4.0 & \cellcolor{green!4} 1.1 & \cellcolor{green!9} 2.3 & \cellcolor{green!0} -- & \cellcolor{green!0} -- & \cellcolor{green!0} -- & \cellcolor{red!0} -0.2 \\
4-Bigfi.. & \cellcolor{green!6} 1.5 & \cellcolor{green!1} 0.4 & \cellcolor{red!0} -0.2 & \cellcolor{green!1} 0.3 & \cellcolor{green!0} -- & \cellcolor{green!0} -- & \cellcolor{green!2} 0.5 \\
5-Fruit.. & \cellcolor{red!3} -0.8 & \cellcolor{green!4} 1.2 & \cellcolor{red!7} -1.9 & \cellcolor{green!1} 0.3 & \cellcolor{green!3} 0.9 & \cellcolor{green!0} -- & \cellcolor{red!0} -0.0 \\
Avg & \cellcolor{red!0} -0.0 & \cellcolor{green!2} 0.7 & \cellcolor{green!0} 0.1 & \cellcolor{green!1} 0.3 & \cellcolor{green!3} 0.9 & \cellcolor{green!0} -- & \cellcolor{green!1} 0.3 \\
\hline
\end{tabular}
}
\hspace{1em}
\subfloat[CLEAR]{ 
 \begin{tabular}{|l|llllll|l|}
\hline
 & 0-Climb.. & 1-Dodge.. & 2-Ninja & 3-Starp.. & 4-Bigfi.. & 5-Fruit.. & Avg \\
0-Climb.. & \cellcolor{green!0} -- & \cellcolor{green!0} -- & \cellcolor{green!0} -- & \cellcolor{green!0} -- & \cellcolor{green!0} -- & \cellcolor{green!0} -- & \cellcolor{green!0} -- \\
1-Dodge.. & \cellcolor{red!0} 0.0 & \cellcolor{green!0} -- & \cellcolor{green!0} -- & \cellcolor{green!0} -- & \cellcolor{green!0} -- & \cellcolor{green!0} -- & \cellcolor{red!0} 0.0 \\
2-Ninja & \cellcolor{green!2} 0.5 & \cellcolor{red!19} -4.9 & \cellcolor{green!0} -- & \cellcolor{green!0} -- & \cellcolor{green!0} -- & \cellcolor{green!0} -- & \cellcolor{red!8} -2.2 \\
3-Starp.. & \cellcolor{red!0} 0.0 & \cellcolor{green!5} 1.4 & \cellcolor{red!2} -0.6 & \cellcolor{green!0} -- & \cellcolor{green!0} -- & \cellcolor{green!0} -- & \cellcolor{green!1} 0.3 \\
4-Bigfi.. & \cellcolor{red!3} -0.8 & \cellcolor{red!6} -1.5 & \cellcolor{green!0} 0.2 & \cellcolor{red!0} -0.2 & \cellcolor{green!0} -- & \cellcolor{green!0} -- & \cellcolor{red!2} -0.6 \\
5-Fruit.. & \cellcolor{green!4} 1.0 & \cellcolor{red!4} -1.1 & \cellcolor{red!3} -0.9 & \cellcolor{red!0} 0.0 & \cellcolor{red!1} -0.4 & \cellcolor{green!0} -- & \cellcolor{red!1} -0.3 \\
Avg & \cellcolor{green!0} 0.1 & \cellcolor{red!6} -1.5 & \cellcolor{red!1} -0.4 & \cellcolor{red!0} -0.1 & \cellcolor{red!1} -0.4 & \cellcolor{green!0} -- & \cellcolor{red!2} -0.5 \\
\hline
\end{tabular}
}

\subfloat[MASK LC]{ 
 \begin{tabular}{|l|llllll|l|}
\hline
 & 0-Climb.. & 1-Dodge.. & 2-Ninja & 3-Starp.. & 4-Bigfi.. & 5-Fruit.. & Avg \\
0-Climb.. & \cellcolor{green!0} -- & \cellcolor{green!0} -- & \cellcolor{green!0} -- & \cellcolor{green!0} -- & \cellcolor{green!0} -- & \cellcolor{green!0} -- & \cellcolor{green!0} -- \\
1-Dodge.. & \cellcolor{red!3} -0.8 & \cellcolor{green!0} -- & \cellcolor{green!0} -- & \cellcolor{green!0} -- & \cellcolor{green!0} -- & \cellcolor{green!0} -- & \cellcolor{red!3} -0.8 \\
2-Ninja & \cellcolor{green!8} 2.1 & \cellcolor{red!8} -2.0 & \cellcolor{green!0} -- & \cellcolor{green!0} -- & \cellcolor{green!0} -- & \cellcolor{green!0} -- & \cellcolor{green!0} 0.1 \\
3-Starp.. & \cellcolor{red!8} -2.2 & \cellcolor{green!12} 3.0 & \cellcolor{red!19} -4.8 & \cellcolor{green!0} -- & \cellcolor{green!0} -- & \cellcolor{green!0} -- & \cellcolor{red!5} -1.3 \\
4-Bigfi.. & \cellcolor{green!9} 2.3 & \cellcolor{red!8} -2.1 & \cellcolor{green!8} 2.0 & \cellcolor{red!9} -2.3 & \cellcolor{green!0} -- & \cellcolor{green!0} -- & \cellcolor{red!0} 0.0 \\
5-Fruit.. & \cellcolor{red!7} -1.9 & \cellcolor{red!3} -0.9 & \cellcolor{green!10} 2.6 & \cellcolor{red!7} -1.8 & \cellcolor{red!3} -0.8 & \cellcolor{green!0} -- & \cellcolor{red!2} -0.5 \\
Avg & \cellcolor{red!0} -0.1 & \cellcolor{red!2} -0.5 & \cellcolor{red!0} -0.0 & \cellcolor{red!8} -2.0 & \cellcolor{red!3} -0.8 & \cellcolor{green!0} -- & \cellcolor{red!2} -0.5 \\
\hline
\end{tabular}
}
\hspace{1em}
\subfloat[MASK BLC]{ 
 \begin{tabular}{|l|llllll|l|}
\hline
 & 0-Climb.. & 1-Dodge.. & 2-Ninja & 3-Starp.. & 4-Bigfi.. & 5-Fruit.. & Avg \\
0-Climb.. & \cellcolor{green!0} -- & \cellcolor{green!0} -- & \cellcolor{green!0} -- & \cellcolor{green!0} -- & \cellcolor{green!0} -- & \cellcolor{green!0} -- & \cellcolor{green!0} -- \\
1-Dodge.. & \cellcolor{red!1} -0.3 & \cellcolor{green!0} -- & \cellcolor{green!0} -- & \cellcolor{green!0} -- & \cellcolor{green!0} -- & \cellcolor{green!0} -- & \cellcolor{red!1} -0.3 \\
2-Ninja & \cellcolor{green!8} 2.2 & \cellcolor{red!8} -2.1 & \cellcolor{green!0} -- & \cellcolor{green!0} -- & \cellcolor{green!0} -- & \cellcolor{green!0} -- & \cellcolor{red!0} 0.0 \\
3-Starp.. & \cellcolor{red!5} -1.4 & \cellcolor{green!17} 4.4 & \cellcolor{red!20} -5.0 & \cellcolor{green!0} -- & \cellcolor{green!0} -- & \cellcolor{green!0} -- & \cellcolor{red!2} -0.6 \\
4-Bigfi.. & \cellcolor{green!0} 0.1 & \cellcolor{red!0} 0.0 & \cellcolor{green!10} 2.6 & \cellcolor{red!11} -2.8 & \cellcolor{green!0} -- & \cellcolor{green!0} -- & \cellcolor{red!0} -0.0 \\
5-Fruit.. & \cellcolor{green!4} 1.2 & \cellcolor{red!0} -0.0 & \cellcolor{green!6} 1.6 & \cellcolor{red!7} -1.9 & \cellcolor{green!1} 0.3 & \cellcolor{green!0} -- & \cellcolor{green!1} 0.3 \\
Avg & \cellcolor{green!1} 0.4 & \cellcolor{green!2} 0.6 & \cellcolor{red!0} -0.2 & \cellcolor{red!9} -2.3 & \cellcolor{green!1} 0.3 & \cellcolor{green!0} -- & \cellcolor{red!0} -0.1 \\
\hline
\end{tabular}
}

%% file: main.bbl
\begin{thebibliography}{83}
\providecommand{\natexlab}[1]{#1}
\providecommand{\url}[1]{\texttt{#1}}
\expandafter\ifx\csname urlstyle\endcsname\relax
  \providecommand{\doi}[1]{doi: #1}\else
  \providecommand{\doi}{doi: \begingroup \urlstyle{rm}\Url}\fi

\bibitem[Abbott \& Regehr(2004)Abbott and Regehr]{abbott2004synaptic}
LF~Abbott and Wade~G Regehr.
\newblock Synaptic computation.
\newblock \emph{Nature}, 431\penalty0 (7010):\penalty0 796--803, 2004.

\bibitem[Aljundi et~al.(2018)Aljundi, Babiloni, Elhoseiny, Rohrbach, and
  Tuytelaars]{aljundi2018mas}
Rahaf Aljundi, Francesca Babiloni, Mohamed Elhoseiny, Marcus Rohrbach, and
  Tinne Tuytelaars.
\newblock Memory aware synapses: Learning what (not) to forget.
\newblock In \emph{Proceedings of the European Conference on Computer Vision
  (ECCV)}, pp.\  139--154, 2018.

\bibitem[Alvarez-Melis \& Fusi(2020)Alvarez-Melis and Fusi]{alvarez2020otdd}
David Alvarez-Melis and Nicolo Fusi.
\newblock Geometric dataset distances via optimal transport.
\newblock \emph{Advances in Neural Information Processing Systems},
  33:\penalty0 21428--21439, 2020.

\bibitem[Arora et~al.(2018)Arora, Ge, Neyshabur, and Zhang]{arora2018stronger}
Sanjeev Arora, Rong Ge, Behnam Neyshabur, and Yi~Zhang.
\newblock Stronger generalization bounds for deep nets via a compression
  approach.
\newblock In \emph{International Conference on Machine Learning}, pp.\
  254--263. PMLR, 2018.

\bibitem[Avery \& Krichmar(2017)Avery and Krichmar]{avery2017neuromodulatory}
Michael~C Avery and Jeffrey~L Krichmar.
\newblock Neuromodulatory systems and their interactions: a review of models,
  theories, and experiments.
\newblock \emph{Frontiers in neural circuits}, pp.\  108, 2017.

\bibitem[Babaeizadeh et~al.(2017)Babaeizadeh, Frosio, Tyree, Clemons, and
  Kautz]{babaeizadeh2017ga3c}
Mohammad Babaeizadeh, Iuri Frosio, Stephen Tyree, Jason Clemons, and Jan Kautz.
\newblock Reinforcement learning through asynchronous advantage actor-critic on
  a gpu.
\newblock In \emph{International Conference on Learning Representations}, 2017.
\newblock URL \url{https://openreview.net/forum?id=r1VGvBcxl}.

\bibitem[Baker et~al.(2023)Baker, New, Aguilar-Simon, Al-Halah, Arnold,
  Ben-Iwhiwhu, Brna, Brooks, Brown, Daniels, Daram, Delattre, Dellana, Eaton,
  Fu, Grauman, Hostetler, Iqbal, Kent, Ketz, Kolouri, Konidaris, Kudithipudi,
  Learned-Miller, Lee, Littman, Madireddy, Mendez, Nguyen, Piatko, Pilly,
  Raghavan, Rahman, Ramakrishnan, Ratzlaff, Soltoggio, Stone, Sur, Tang,
  Tiwari, Vedder, Wang, Xu, Yanguas-Gil, Yedidsion, Yu, and
  Vallabha]{baker2023}
Megan~M. Baker, Alexander New, Mario Aguilar-Simon, Ziad Al-Halah,
  Sébastien~M.R. Arnold, Ese Ben-Iwhiwhu, Andrew~P. Brna, Ethan Brooks,
  Ryan~C. Brown, Zachary Daniels, Anurag Daram, Fabien Delattre, Ryan Dellana,
  Eric Eaton, Haotian Fu, Kristen Grauman, Jesse Hostetler, Shariq Iqbal, David
  Kent, Nicholas Ketz, Soheil Kolouri, George Konidaris, Dhireesha Kudithipudi,
  Erik Learned-Miller, Seungwon Lee, Michael~L. Littman, Sandeep Madireddy,
  Jorge~A. Mendez, Eric~Q. Nguyen, Christine Piatko, Praveen~K. Pilly, Aswin
  Raghavan, Abrar Rahman, Santhosh~Kumar Ramakrishnan, Neale Ratzlaff, Andrea
  Soltoggio, Peter Stone, Indranil Sur, Zhipeng Tang, Saket Tiwari, Kyle
  Vedder, Felix Wang, Zifan Xu, Angel Yanguas-Gil, Harel Yedidsion, Shangqun
  Yu, and Gautam~K. Vallabha.
\newblock A domain-agnostic approach for characterization of lifelong learning
  systems.
\newblock \emph{Neural Networks}, 2023.
\newblock ISSN 0893-6080.
\newblock \doi{https://doi.org/10.1016/j.neunet.2023.01.007}.
\newblock URL
  \url{https://www.sciencedirect.com/science/article/pii/S0893608023000072}.

\bibitem[Bear et~al.(2020)Bear, Connors, and Paradiso]{bear2020neuroscience}
Mark Bear, Barry Connors, and Michael~A Paradiso.
\newblock \emph{Neuroscience: Exploring the Brain, Enhanced Edition: Exploring
  the Brain}.
\newblock Jones \& Bartlett Learning, 2020.

\bibitem[Ben-Iwhiwhu et~al.(2022)Ben-Iwhiwhu, Dick, Ketz, Pilly, and
  Soltoggio]{ben2022context}
Eseoghene Ben-Iwhiwhu, Jeffery Dick, Nicholas~A Ketz, Praveen~K Pilly, and
  Andrea Soltoggio.
\newblock Context meta-reinforcement learning via neuromodulation.
\newblock \emph{Neural Networks}, 152:\penalty0 70--79, 2022.

\bibitem[Bengio et~al.(2013)Bengio, L{\'e}onard, and
  Courville]{bengio2013straight_through}
Yoshua Bengio, Nicholas L{\'e}onard, and Aaron Courville.
\newblock Estimating or propagating gradients through stochastic neurons for
  conditional computation.
\newblock \emph{arXiv preprint arXiv:1308.3432}, 2013.

\bibitem[Benna \& Fusi(2016)Benna and Fusi]{benna2016computational}
Marcus~K Benna and Stefano Fusi.
\newblock Computational principles of synaptic memory consolidation.
\newblock \emph{Nature neuroscience}, 19\penalty0 (12):\penalty0 1697--1706,
  2016.

\bibitem[Chaudhry et~al.(2018)Chaudhry, Dokania, Ajanthan, and
  Torr]{chaudhry2018riemannian}
Arslan Chaudhry, Puneet~K Dokania, Thalaiyasingam Ajanthan, and Philip~HS Torr.
\newblock Riemannian walk for incremental learning: Understanding forgetting
  and intransigence.
\newblock In \emph{Proceedings of the European Conference on Computer Vision
  (ECCV)}, pp.\  532--547, 2018.

\bibitem[Chaudhry et~al.(2019)Chaudhry, Ranzato, Rohrbach, and
  Elhoseiny]{chaudhry2018agem}
Arslan Chaudhry, Marc’Aurelio Ranzato, Marcus Rohrbach, and Mohamed
  Elhoseiny.
\newblock Efficient lifelong learning with a-{GEM}.
\newblock In \emph{International Conference on Learning Representations}, 2019.
\newblock URL \url{https://openreview.net/forum?id=Hkf2_sC5FX}.

\bibitem[Chevalier-Boisvert et~al.(2018)Chevalier-Boisvert, Willems, and
  Pal]{gym_minigrid}
Maxime Chevalier-Boisvert, Lucas Willems, and Suman Pal.
\newblock Minimalistic gridworld environment for openai gym.
\newblock \url{https://github.com/maximecb/gym-minigrid}, 2018.

\bibitem[Chevalier-Boisvert et~al.(2023)Chevalier-Boisvert, Dai, Towers,
  de~Lazcano, Willems, Lahlou, Pal, Castro, and Terry]{chevalier2023minigrid}
Maxime Chevalier-Boisvert, Bolun Dai, Mark Towers, Rodrigo de~Lazcano, Lucas
  Willems, Salem Lahlou, Suman Pal, Pablo~Samuel Castro, and Jordan Terry.
\newblock Minigrid \& miniworld: Modular \& customizable reinforcement learning
  environments for goal-oriented tasks.
\newblock \emph{arXiv preprint arXiv:2306.13831}, 2023.

\bibitem[Cobbe et~al.(2020)Cobbe, Hesse, Hilton, and
  Schulman]{cobbe2020procgen}
Karl Cobbe, Chris Hesse, Jacob Hilton, and John Schulman.
\newblock Leveraging procedural generation to benchmark reinforcement learning.
\newblock In \emph{International conference on machine learning}, pp.\
  2048--2056. PMLR, 2020.

\bibitem[Colas et~al.(2018)Colas, Sigaud, and Oudeyer]{colas2018many}
C{\'e}dric Colas, Olivier Sigaud, and Pierre-Yves Oudeyer.
\newblock How many random seeds? statistical power analysis in deep
  reinforcement learning experiments.
\newblock \emph{arXiv preprint arXiv:1806.08295}, 2018.

\bibitem[Courbariaux et~al.(2015)Courbariaux, Bengio, and
  David]{courbariaux2015binaryconnect}
Matthieu Courbariaux, Yoshua Bengio, and Jean-Pierre David.
\newblock Binaryconnect: Training deep neural networks with binary weights
  during propagations.
\newblock \emph{Advances in neural information processing systems}, 28, 2015.

\bibitem[De~Lange et~al.(2021)De~Lange, Aljundi, Masana, Parisot, Jia,
  Leonardis, Slabaugh, and Tuytelaars]{de2021continual}
Matthias De~Lange, Rahaf Aljundi, Marc Masana, Sarah Parisot, Xu~Jia,
  Ale{\v{s}} Leonardis, Gregory Slabaugh, and Tinne Tuytelaars.
\newblock A continual learning survey: Defying forgetting in classification
  tasks.
\newblock \emph{IEEE transactions on pattern analysis and machine
  intelligence}, 44\penalty0 (7):\penalty0 3366--3385, 2021.

\bibitem[Doya(2002)]{doya2002metalearn_neuromod}
Kenji Doya.
\newblock Metalearning and neuromodulation.
\newblock \emph{Neural networks}, 15\penalty0 (4-6):\penalty0 495--506, 2002.

\bibitem[Espeholt et~al.(2018)Espeholt, Soyer, Munos, Simonyan, Mnih, Ward,
  Doron, Firoiu, Harley, Dunning, et~al.]{espeholt2018impala}
Lasse Espeholt, Hubert Soyer, Remi Munos, Karen Simonyan, Vlad Mnih, Tom Ward,
  Yotam Doron, Vlad Firoiu, Tim Harley, Iain Dunning, et~al.
\newblock Impala: Scalable distributed deep-rl with importance weighted
  actor-learner architectures.
\newblock In \emph{International conference on machine learning}, pp.\
  1407--1416. PMLR, 2018.

\bibitem[Farajtabar et~al.(2020)Farajtabar, Azizan, Mott, and
  Li]{farajtabar2020ogd}
Mehrdad Farajtabar, Navid Azizan, Alex Mott, and Ang Li.
\newblock Orthogonal gradient descent for continual learning.
\newblock In \emph{International Conference on Artificial Intelligence and
  Statistics}, pp.\  3762--3773. PMLR, 2020.

\bibitem[Fellous \& Linster(1998)Fellous and Linster]{fellous1998computational}
Jean-Marc Fellous and Christiane Linster.
\newblock Computational models of neuromodulation.
\newblock \emph{Neural computation}, 10\penalty0 (4):\penalty0 771--805, 1998.

\bibitem[Frankle \& Carbin(2018)Frankle and Carbin]{frankle2018lottery}
Jonathan Frankle and Michael Carbin.
\newblock The lottery ticket hypothesis: Finding sparse, trainable neural
  networks.
\newblock \emph{arXiv preprint arXiv:1803.03635}, 2018.

\bibitem[Fujimoto et~al.(2018)Fujimoto, Hoof, and Meger]{fujimoto2018td3}
Scott Fujimoto, Herke Hoof, and David Meger.
\newblock Addressing function approximation error in actor-critic methods.
\newblock In \emph{International conference on machine learning}, pp.\
  1587--1596. PMLR, 2018.

\bibitem[Guo et~al.(2020)Guo, Liu, Yang, and Rosing]{guo2020gem_unified}
Yunhui Guo, Mingrui Liu, Tianbao Yang, and Tajana Rosing.
\newblock Improved schemes for episodic memory-based lifelong learning.
\newblock \emph{Advances in Neural Information Processing Systems},
  33:\penalty0 1023--1035, 2020.

\bibitem[Haarnoja et~al.(2018)Haarnoja, Zhou, Abbeel, and
  Levine]{haarnoja2018sac}
Tuomas Haarnoja, Aurick Zhou, Pieter Abbeel, and Sergey Levine.
\newblock Soft actor-critic: Off-policy maximum entropy deep reinforcement
  learning with a stochastic actor.
\newblock In \emph{International conference on machine learning}, pp.\
  1861--1870. PMLR, 2018.

\bibitem[Hadsell et~al.(2020)Hadsell, Rao, Rusu, and
  Pascanu]{hadsell2020embracing}
Raia Hadsell, Dushyant Rao, Andrei~A Rusu, and Razvan Pascanu.
\newblock Embracing change: Continual learning in deep neural networks.
\newblock \emph{Trends in cognitive sciences}, 24\penalty0 (12):\penalty0
  1028--1040, 2020.

\bibitem[Henderson et~al.(2018)Henderson, Islam, Bachman, Pineau, Precup, and
  Meger]{henderson2018deep}
Peter Henderson, Riashat Islam, Philip Bachman, Joelle Pineau, Doina Precup,
  and David Meger.
\newblock Deep reinforcement learning that matters.
\newblock In \emph{Proceedings of the AAAI conference on artificial
  intelligence}, volume~32, 2018.

\bibitem[Hoefler et~al.(2021)Hoefler, Alistarh, Ben-Nun, Dryden, and
  Peste]{hoefler2021sparsity}
Torsten Hoefler, Dan Alistarh, Tal Ben-Nun, Nikoli Dryden, and Alexandra Peste.
\newblock Sparsity in deep learning: Pruning and growth for efficient inference
  and training in neural networks.
\newblock \emph{The Journal of Machine Learning Research}, 22\penalty0
  (1):\penalty0 10882--11005, 2021.

\bibitem[Kaplanis et~al.(2018)Kaplanis, Shanahan, and
  Clopath]{kaplanis2018continual}
Christos Kaplanis, Murray Shanahan, and Claudia Clopath.
\newblock Continual reinforcement learning with complex synapses.
\newblock In \emph{International Conference on Machine Learning}, pp.\
  2497--2506. PMLR, 2018.

\bibitem[Kaplanis et~al.(2019)Kaplanis, Shanahan, and
  Clopath]{kaplanis2019policy}
Christos Kaplanis, Murray Shanahan, and Claudia Clopath.
\newblock Policy consolidation for continual reinforcement learning.
\newblock \emph{arXiv preprint arXiv:1902.00255}, 2019.

\bibitem[Kempka et~al.(2016)Kempka, Wydmuch, Runc, Toczek, and
  Ja{\'s}kowski]{kempka2016vizdoom}
Micha{\l} Kempka, Marek Wydmuch, Grzegorz Runc, Jakub Toczek, and Wojciech
  Ja{\'s}kowski.
\newblock Vizdoom: A doom-based ai research platform for visual reinforcement
  learning.
\newblock In \emph{2016 IEEE conference on computational intelligence and games
  (CIG)}, pp.\  1--8. IEEE, 2016.

\bibitem[Kessler et~al.(2020)Kessler, Parker-Holder, Ball, Zohren, and
  Roberts]{kessler2020unclear}
Samuel Kessler, Jack Parker-Holder, Philip Ball, Stefan Zohren, and Stephen~J
  Roberts.
\newblock Unclear: A straightforward method for continual reinforcement
  learning.
\newblock In \emph{Proceedings of the 37th International Conference on Machine
  Learning}, 2020.

\bibitem[Kessler et~al.(2022)Kessler, Parker-Holder, Ball, Zohren, and
  Roberts]{kessler2022owl}
Samuel Kessler, Jack Parker-Holder, Philip Ball, Stefan Zohren, and Stephen~J
  Roberts.
\newblock Same state, different task: Continual reinforcement learning without
  interference.
\newblock In \emph{Proceedings of the AAAI Conference on Artificial
  Intelligence}, volume~36, pp.\  7143--7151, 2022.

\bibitem[Khetarpal et~al.(2020)Khetarpal, Riemer, Rish, and
  Precup]{khetarpal2020towards}
Khimya Khetarpal, Matthew Riemer, Irina Rish, and Doina Precup.
\newblock Towards continual reinforcement learning: A review and perspectives.
\newblock \emph{arXiv preprint arXiv:2012.13490}, 2020.

\bibitem[Kirkpatrick et~al.(2017)Kirkpatrick, Pascanu, Rabinowitz, Veness,
  Desjardins, Rusu, Milan, Quan, Ramalho, Grabska-Barwinska,
  et~al.]{kirkpatrick2017ewc}
James Kirkpatrick, Razvan Pascanu, Neil Rabinowitz, Joel Veness, Guillaume
  Desjardins, Andrei~A Rusu, Kieran Milan, John Quan, Tiago Ramalho, Agnieszka
  Grabska-Barwinska, et~al.
\newblock Overcoming catastrophic forgetting in neural networks.
\newblock \emph{Proceedings of the national academy of sciences}, 114\penalty0
  (13):\penalty0 3521--3526, 2017.

\bibitem[Kolouri et~al.(2019)Kolouri, Ketz, Soltoggio, and
  Pilly]{kolouri2019scp}
Soheil Kolouri, Nicholas~A Ketz, Andrea Soltoggio, and Praveen~K Pilly.
\newblock Sliced cramer synaptic consolidation for preserving deeply learned
  representations.
\newblock In \emph{International Conference on Learning Representations}, 2019.

\bibitem[Koster et~al.(2022)Koster, Grothe, and
  Rettinger]{koster2022signing_sm}
Nils Koster, Oliver Grothe, and Achim Rettinger.
\newblock Signing the supermask: Keep, hide, invert.
\newblock \emph{arXiv preprint arXiv:2201.13361}, 2022.

\bibitem[Kudithipudi et~al.(2022)Kudithipudi, Aguilar-Simon, Babb, Bazhenov,
  Blackiston, Bongard, Brna, Chakravarthi~Raja, Cheney, Clune, Daram, Fusi,
  Helfer, Kay, Ketz, Kira, Kolouri, Krichmar, Kriegman, Levin, Madireddy,
  Manicka, Marjaninejad, McNaughton, Miikkulainen, Navratilova, Pandit, Parker,
  Pilly, Risi, Sejnowski, Soltoggio, Soures, Tolias, Urbina-Meléndez,
  Valero-Cuevas, Van~de Ven, Vogelstein, Wang, Weiss, Yanguas-Gil, Zou, and
  Siegelmann]{kudithipudi2022biological}
Dhireesha Kudithipudi, Mario Aguilar-Simon, Jonathan Babb, Maxim Bazhenov,
  Douglas Blackiston, Josh Bongard, Andrew~P Brna, Suraj Chakravarthi~Raja,
  Nick Cheney, Jeff Clune, Anurag Daram, Stefano Fusi, Peter Helfer, Leslie
  Kay, Nicholas Ketz, Zsolt Kira, Soheil Kolouri, Jeffrey~L. Krichmar, Sam
  Kriegman, Michael Levin, Sandeep Madireddy, Santosh Manicka, Ali
  Marjaninejad, Bruce McNaughton, Risto Miikkulainen, Zaneta Navratilova, Tej
  Pandit, Alice Parker, Praveen~K. Pilly, Sebastian Risi, Terrence~J.
  Sejnowski, Andrea Soltoggio, Nicholas Soures, Andreas~S. Tolias, Darío
  Urbina-Meléndez, Francisco~J. Valero-Cuevas, Gido~M. Van~de Ven, Joshua~T.
  Vogelstein, Felix Wang, Ron Weiss, Angel Yanguas-Gil, Xinyun Zou, and Hava
  Siegelmann.
\newblock Biological underpinnings for lifelong learning machines.
\newblock \emph{Nature Machine Intelligence}, 4\penalty0 (3):\penalty0
  196--210, 2022.

\bibitem[Lillicrap et~al.(2016)Lillicrap, Hunt, Pritzel, Heess, Erez, Tassa,
  Silver, and Wierstra]{lillicrap2016ddpg}
Timothy~P Lillicrap, Jonathan~J Hunt, Alexander Pritzel, Nicolas Heess, Tom
  Erez, Yuval Tassa, David Silver, and Daan Wierstra.
\newblock Continuous control with deep reinforcement learning.
\newblock In \emph{International Conference on Learning Representations}, 2016.

\bibitem[Lin et~al.(2022)Lin, Yang, Fan, and Zhang]{lin2022trgp}
Sen Lin, Li~Yang, Deliang Fan, and Junshan Zhang.
\newblock {TRGP}: Trust region gradient projection for continual learning.
\newblock In \emph{International Conference on Learning Representations}, 2022.
\newblock URL \url{https://openreview.net/forum?id=iEvAf8i6JjO}.

\bibitem[Liu et~al.(2022)Liu, Bai, Lu, Soltoggio, and
  Kolouri]{liu2022wasserstein}
Xinran Liu, Yikun Bai, Yuzhe Lu, Andrea Soltoggio, and Soheil Kolouri.
\newblock Wasserstein task embedding for measuring task similarities.
\newblock \emph{arXiv preprint arXiv:2208.11726}, 2022.

\bibitem[Lopez-Paz \& Ranzato(2017)Lopez-Paz and Ranzato]{lopez2017gem}
David Lopez-Paz and Marc'Aurelio Ranzato.
\newblock Gradient episodic memory for continual learning.
\newblock \emph{Advances in neural information processing systems}, 30, 2017.

\bibitem[Mallya \& Lazebnik(2018)Mallya and Lazebnik]{mallya2018packnet}
Arun Mallya and Svetlana Lazebnik.
\newblock Packnet: Adding multiple tasks to a single network by iterative
  pruning.
\newblock In \emph{Proceedings of the IEEE conference on Computer Vision and
  Pattern Recognition}, pp.\  7765--7773, 2018.

\bibitem[Mallya et~al.(2018)Mallya, Davis, and Lazebnik]{mallya2018piggyback}
Arun Mallya, Dillon Davis, and Svetlana Lazebnik.
\newblock Piggyback: Adapting a single network to multiple tasks by learning to
  mask weights.
\newblock In \emph{Proceedings of the European Conference on Computer Vision
  (ECCV)}, pp.\  67--82, 2018.

\bibitem[Martin \& Pilly(2019)Martin and Pilly]{martin2019ppn}
Charles~E Martin and Praveen~K Pilly.
\newblock Probabilistic program neurogenesis.
\newblock In \emph{ALIFE 2019: The 2019 Conference on Artificial Life}, pp.\
  440--447. MIT Press, 2019.

\bibitem[Mendez et~al.(2022)Mendez, van Seijen, and
  Eaton]{mendez2022neuralcomposition}
Jorge~A Mendez, Harm van Seijen, and Eric Eaton.
\newblock Modular lifelong reinforcement learning via neural composition.
\newblock In \emph{International Conference on Learning Representations}, 2022.
\newblock URL \url{https://openreview.net/forum?id=5XmLzdslFNN}.

\bibitem[Milan et~al.(2016)Milan, Veness, Kirkpatrick, Bowling, Koop, and
  Hassabis]{milan2016forgetmenot}
Kieran Milan, Joel Veness, James Kirkpatrick, Michael Bowling, Anna Koop, and
  Demis Hassabis.
\newblock The forget-me-not process.
\newblock \emph{Advances in Neural Information Processing Systems}, 29, 2016.

\bibitem[Mnih et~al.(2013)Mnih, Kavukcuoglu, Silver, Graves, Antonoglou,
  Wierstra, and Riedmiller]{mnih2013playing}
Volodymyr Mnih, Koray Kavukcuoglu, David Silver, Alex Graves, Ioannis
  Antonoglou, Daan Wierstra, and Martin Riedmiller.
\newblock Playing atari with deep reinforcement learning.
\newblock \emph{arXiv preprint arXiv:1312.5602}, 2013.

\bibitem[Mnih et~al.(2016)Mnih, Badia, Mirza, Graves, Lillicrap, Harley,
  Silver, and Kavukcuoglu]{mnih2016a3c}
Volodymyr Mnih, Adria~Puigdomenech Badia, Mehdi Mirza, Alex Graves, Timothy
  Lillicrap, Tim Harley, David Silver, and Koray Kavukcuoglu.
\newblock Asynchronous methods for deep reinforcement learning.
\newblock In \emph{International conference on machine learning}, pp.\
  1928--1937. PMLR, 2016.

\bibitem[New et~al.(2022)New, Baker, Nguyen, and Vallabha]{new2022lifelong}
Alexander New, Megan Baker, Eric Nguyen, and Gautam Vallabha.
\newblock Lifelong learning metrics.
\newblock \emph{arXiv preprint arXiv:2201.08278}, 2022.

\bibitem[Parisi et~al.(2019)Parisi, Kemker, Part, Kanan, and
  Wermter]{parisi2019continual}
German~I Parisi, Ronald Kemker, Jose~L Part, Christopher Kanan, and Stefan
  Wermter.
\newblock Continual lifelong learning with neural networks: A review.
\newblock \emph{Neural Networks}, 113:\penalty0 54--71, 2019.

\bibitem[Powers et~al.(2022)Powers, Xing, Kolve, Mottaghi, and
  Gupta]{powers2022cora}
Sam Powers, Eliot Xing, Eric Kolve, Roozbeh Mottaghi, and Abhinav Gupta.
\newblock Cora: Benchmarks, baselines, and metrics as a platform for continual
  reinforcement learning agents.
\newblock In \emph{Conference on Lifelong Learning Agents (CoLLAs)}, 2022.

\bibitem[Ramanujan et~al.(2020)Ramanujan, Wortsman, Kembhavi, Farhadi, and
  Rastegari]{ramanujan2020s}
Vivek Ramanujan, Mitchell Wortsman, Aniruddha Kembhavi, Ali Farhadi, and
  Mohammad Rastegari.
\newblock What's hidden in a randomly weighted neural network?
\newblock In \emph{Proceedings of the IEEE/CVF Conference on Computer Vision
  and Pattern Recognition}, pp.\  11893--11902, 2020.

\bibitem[Rolnick et~al.(2019)Rolnick, Ahuja, Schwarz, Lillicrap, and
  Wayne]{rolnick2019clear}
David Rolnick, Arun Ahuja, Jonathan Schwarz, Timothy Lillicrap, and Gregory
  Wayne.
\newblock Experience replay for continual learning.
\newblock \emph{Advances in Neural Information Processing Systems}, 32, 2019.

\bibitem[Rusu et~al.(2016)Rusu, Rabinowitz, Desjardins, Soyer, Kirkpatrick,
  Kavukcuoglu, Pascanu, and Hadsell]{rusu2016progressive}
Andrei~A Rusu, Neil~C Rabinowitz, Guillaume Desjardins, Hubert Soyer, James
  Kirkpatrick, Koray Kavukcuoglu, Razvan Pascanu, and Raia Hadsell.
\newblock Progressive neural networks.
\newblock \emph{arXiv preprint arXiv:1606.04671}, 2016.

\bibitem[Saha et~al.(2021)Saha, Garg, and Roy]{saha2021gpm}
Gobinda Saha, Isha Garg, and Kaushik Roy.
\newblock Gradient projection memory for continual learning.
\newblock In \emph{International Conference on Learning Representations}, 2021.
\newblock URL \url{https://openreview.net/forum?id=3AOj0RCNC2}.

\bibitem[Schulman et~al.(2015)Schulman, Levine, Abbeel, Jordan, and
  Moritz]{schulman2015trpo}
John Schulman, Sergey Levine, Pieter Abbeel, Michael Jordan, and Philipp
  Moritz.
\newblock Trust region policy optimization.
\newblock In \emph{International conference on machine learning}, pp.\
  1889--1897. PMLR, 2015.

\bibitem[Schulman et~al.(2017)Schulman, Wolski, Dhariwal, Radford, and
  Klimov]{schulman2017ppo}
John Schulman, Filip Wolski, Prafulla Dhariwal, Alec Radford, and Oleg Klimov.
\newblock Proximal policy optimization algorithms.
\newblock \emph{arXiv preprint arXiv:1707.06347}, 2017.

\bibitem[Schultz et~al.(1997)Schultz, Dayan, and Montague]{schultz1997neural}
Wolfram Schultz, Peter Dayan, and P~Read Montague.
\newblock A neural substrate of prediction and reward.
\newblock \emph{Science}, 275\penalty0 (5306):\penalty0 1593--1599, 1997.

\bibitem[Schwarz et~al.(2018)Schwarz, Czarnecki, Luketina, Grabska-Barwinska,
  Teh, Pascanu, and Hadsell]{schwarz2018pnc}
Jonathan Schwarz, Wojciech Czarnecki, Jelena Luketina, Agnieszka
  Grabska-Barwinska, Yee~Whye Teh, Razvan Pascanu, and Raia Hadsell.
\newblock Progress \& compress: A scalable framework for continual learning.
\newblock In \emph{International Conference on Machine Learning}, pp.\
  4528--4537. PMLR, 2018.

\bibitem[Schwarz et~al.(2021)Schwarz, Jayakumar, Pascanu, Latham, and
  Teh]{schwarz2021powerpropagation}
Jonathan Schwarz, Siddhant Jayakumar, Razvan Pascanu, Peter~E Latham, and Yee
  Teh.
\newblock Powerpropagation: A sparsity inducing weight reparameterisation.
\newblock \emph{Advances in neural information processing systems},
  34:\penalty0 28889--28903, 2021.

\bibitem[Serra et~al.(2018)Serra, Suris, Miron, and Karatzoglou]{serra2018hat}
Joan Serra, Didac Suris, Marius Miron, and Alexandros Karatzoglou.
\newblock Overcoming catastrophic forgetting with hard attention to the task.
\newblock In \emph{International Conference on Machine Learning}, pp.\
  4548--4557. PMLR, 2018.

\bibitem[Shao et~al.(2018)Shao, Zhao, Li, and Zhu]{shao2018learning}
Kun Shao, Dongbin Zhao, Nannan Li, and Yuanheng Zhu.
\newblock Learning battles in vizdoom via deep reinforcement learning.
\newblock In \emph{2018 IEEE Conference on Computational Intelligence and Games
  (CIG)}, pp.\  1--4. IEEE, 2018.

\bibitem[Sokar et~al.(2021)Sokar, Mocanu, and Pechenizkiy]{sokar2021spacenet}
Ghada Sokar, Decebal~Constantin Mocanu, and Mykola Pechenizkiy.
\newblock Spacenet: Make free space for continual learning.
\newblock \emph{Neurocomputing}, 439:\penalty0 1--11, 2021.

\bibitem[Soltoggio et~al.(2008)Soltoggio, Bullinaria, Mattiussi, D{\"u}rr, and
  Floreano]{soltoggio2008evolutionary}
Andrea Soltoggio, John~A Bullinaria, Claudio Mattiussi, Peter D{\"u}rr, and
  Dario Floreano.
\newblock Evolutionary advantages of neuromodulated plasticity in dynamic,
  reward-based scenarios.
\newblock In \emph{Proceedings of the 11th international conference on
  artificial life (Alife XI)}, number CONF, pp.\  569--576. MIT Press, 2008.

\bibitem[Soltoggio et~al.(2018)Soltoggio, Stanley, and Risi]{soltoggio2018born}
Andrea Soltoggio, Kenneth~O Stanley, and Sebastian Risi.
\newblock Born to learn: the inspiration, progress, and future of evolved
  plastic artificial neural networks.
\newblock \emph{Neural Networks}, 108:\penalty0 48--67, 2018.

\bibitem[Soltoggio et~al.(2019)Soltoggio, Ladosz, Ben-Iwhiwhu, and
  Dick]{soltoggio2019ctgraph}
Andrea Soltoggio, Pawel Ladosz, Eseoghene Ben-Iwhiwhu, and Jeff Dick.
\newblock The {CT}-graph environments, 2019.
\newblock URL \url{https://github.com/soltoggio/ct-graph}.

\bibitem[Soltoggio et~al.(2023)Soltoggio, Ben-Iwhiwhu, Peridis, Ladosz, Dick,
  Pilly, and Kolouri]{soltoggio2023ctgraph}
Andrea Soltoggio, Eseoghene Ben-Iwhiwhu, Christos Peridis, Pawel Ladosz,
  Jeffery Dick, Praveen~K Pilly, and Soheil Kolouri.
\newblock The configurable tree graph (ct-graph): measurable problems in
  partially observable and distal reward environments for lifelong
  reinforcement learning.
\newblock \emph{arXiv preprint arXiv:2302.10887}, 2023.

\bibitem[Todorov et~al.(2012)Todorov, Erez, and Tassa]{todorov2012mujoco}
Emanuel Todorov, Tom Erez, and Yuval Tassa.
\newblock Mujoco: A physics engine for model-based control.
\newblock In \emph{2012 IEEE/RSJ international conference on intelligent robots
  and systems}, pp.\  5026--5033. IEEE, 2012.

\bibitem[Van~de Ven \& Tolias(2019)Van~de Ven and Tolias]{van2019three}
Gido~M Van~de Ven and Andreas~S Tolias.
\newblock Three scenarios for continual learning.
\newblock \emph{arXiv preprint arXiv:1904.07734}, 2019.

\bibitem[Van~Hasselt et~al.(2016)Van~Hasselt, Guez, and
  Silver]{van2016doubledqn}
Hado Van~Hasselt, Arthur Guez, and David Silver.
\newblock Deep reinforcement learning with double q-learning.
\newblock In \emph{Proceedings of the AAAI conference on artificial
  intelligence}, volume~30, 2016.

\bibitem[von Oswald et~al.(2020)von Oswald, Henning, Grewe, and
  Sacramento]{oswald2020continual_hypernet}
Johannes von Oswald, Christian Henning, Benjamin~F. Grewe, and João
  Sacramento.
\newblock Continual learning with hypernetworks.
\newblock In \emph{International Conference on Learning Representations}, 2020.
\newblock URL \url{https://openreview.net/forum?id=SJgwNerKvB}.

\bibitem[Von~Oswald et~al.(2021)Von~Oswald, Zhao, Kobayashi, Schug, Caccia,
  Zucchet, and Sacramento]{von2021learning}
Johannes Von~Oswald, Dominic Zhao, Seijin Kobayashi, Simon Schug, Massimo
  Caccia, Nicolas Zucchet, and Jo{\~a}o Sacramento.
\newblock Learning where to learn: Gradient sparsity in meta and continual
  learning.
\newblock \emph{Advances in Neural Information Processing Systems},
  34:\penalty0 5250--5263, 2021.

\bibitem[Wo{\l}czyk et~al.(2021)Wo{\l}czyk, Zaj{\k{a}}c, Pascanu, Kuci{\'n}ski,
  and Mi{\l}o{\'s}]{wolczyk2021continualworld}
Maciej Wo{\l}czyk, Micha{\l} Zaj{\k{a}}c, Razvan Pascanu, {\L}ukasz
  Kuci{\'n}ski, and Piotr Mi{\l}o{\'s}.
\newblock Continual world: A robotic benchmark for continual reinforcement
  learning.
\newblock \emph{Advances in Neural Information Processing Systems},
  34:\penalty0 28496--28510, 2021.

\bibitem[Wortsman et~al.(2020)Wortsman, Ramanujan, Liu, Kembhavi, Rastegari,
  Yosinski, and Farhadi]{wortsman2020supermasks}
Mitchell Wortsman, Vivek Ramanujan, Rosanne Liu, Aniruddha Kembhavi, Mohammad
  Rastegari, Jason Yosinski, and Ali Farhadi.
\newblock Supermasks in superposition.
\newblock \emph{Advances in Neural Information Processing Systems},
  33:\penalty0 15173--15184, 2020.

\bibitem[Yoon et~al.(2018)Yoon, Yang, Lee, and Hwang]{yoon2018den}
Jaehong Yoon, Eunho Yang, Jeongtae Lee, and Sung~Ju Hwang.
\newblock Lifelong learning with dynamically expandable networks.
\newblock In \emph{International Conference on Learning Representations}, 2018.
\newblock URL \url{https://openreview.net/forum?id=Sk7KsfW0-}.

\bibitem[Yu et~al.(2020)Yu, Quillen, He, Julian, Hausman, Finn, and
  Levine]{yu2020meta}
Tianhe Yu, Deirdre Quillen, Zhanpeng He, Ryan Julian, Karol Hausman, Chelsea
  Finn, and Sergey Levine.
\newblock Meta-world: A benchmark and evaluation for multi-task and meta
  reinforcement learning.
\newblock In \emph{Conference on robot learning}, pp.\  1094--1100. PMLR, 2020.

\bibitem[Zeng et~al.(2019)Zeng, Chen, Cui, and Yu]{zeng2019owm}
Guanxiong Zeng, Yang Chen, Bo~Cui, and Shan Yu.
\newblock Continual learning of context-dependent processing in neural
  networks.
\newblock \emph{Nature Machine Intelligence}, 1\penalty0 (8):\penalty0
  364--372, 2019.

\bibitem[Zenke et~al.(2017)Zenke, Poole, and Ganguli]{zenke2017si}
Friedemann Zenke, Ben Poole, and Surya Ganguli.
\newblock Continual learning through synaptic intelligence.
\newblock In \emph{International Conference on Machine Learning}, pp.\
  3987--3995. PMLR, 2017.

\bibitem[Zhou et~al.(2019)Zhou, Lan, Liu, and Yosinski]{zhou2019deconstructing}
Hattie Zhou, Janice Lan, Rosanne Liu, and Jason Yosinski.
\newblock Deconstructing lottery tickets: Zeros, signs, and the supermask.
\newblock \emph{Advances in neural information processing systems}, 32, 2019.

\bibitem[Zhou et~al.(2018)Zhou, Veitch, Austern, Adams, and
  Orbanz]{zhou2018non}
Wenda Zhou, Victor Veitch, Morgane Austern, Ryan~P Adams, and Peter Orbanz.
\newblock Non-vacuous generalization bounds at the imagenet scale: a
  pac-bayesian compression approach.
\newblock \emph{arXiv preprint arXiv:1804.05862}, 2018.

\end{thebibliography}
